\PassOptionsToPackage{numbers, compress}{natbib}
\documentclass{article}

\usepackage[main, final]{neurips_2026}   
\providecommand{\linenumbers}{}\providecommand{\nolinenumbers}{}

\usepackage[utf8]{inputenc}
\usepackage[T1]{fontenc}
\usepackage[hidelinks]{hyperref}
\usepackage{url}
\usepackage{booktabs}
\usepackage{amsfonts}
\usepackage{nicefrac}
\usepackage{microtype}

\usepackage{graphicx}
\usepackage{multirow}
\usepackage{amsmath, amssymb, amsthm}
\usepackage{mathtools}
\usepackage[table]{xcolor}     
\usepackage{colortbl}
\usepackage{enumitem}
\usepackage[capitalize,noabbrev]{cleveref}

\usepackage{float}
\usepackage{algorithm}
\usepackage{algorithmic}

\usepackage{rotating}

\usepackage{caption}
\usepackage{subcaption}
\makeatletter
\@ifundefined{table*}{%
  \newenvironment{table*}{\begin{table}}{\end{table}}%
}{}
\@ifundefined{figure*}{%
  \newenvironment{figure*}{\begin{figure}}{\end{figure}}%
}{}
\makeatother

\theoremstyle{plain}
\newtheorem{theorem}{Theorem}[section]
\newtheorem{proposition}[theorem]{Proposition}
\newtheorem{lemma}[theorem]{Lemma}

\theoremstyle{definition}

\newtheorem{assumption}[theorem]{Assumption}
\theoremstyle{remark}

\newcommand{\std}[1]{\ensuremath{\,\pm{\scriptstyle #1}}}

\newcommand{\mstd}[2]{\ensuremath{#1 \pm {\scriptstyle #2}}}
\newcommand{\bst}[2]{\ensuremath{\mathbf{#1 \pm {\scriptstyle #2}}}}

\newif\ifshowrev 
\showrevfalse
\definecolor{revblue}{RGB}{0,70,200}
\definecolor{revviolet}{RGB}{140,0,170}
\newcommand{\rev}[1]{\ifshowrev{\color{revblue}#1}\else#1\fi}
\newenvironment{revblock}{\ifshowrev\color{revblue}\fi}{}
\newcommand{\revfloat}{\ifshowrev\color{revblue}\fi}  

\newcommand{\discuss}[1]{\ifshowrev{\color{revviolet}[\textbf{DISCUSS:} #1]}\fi}

\title{Black-Box Uncertainty Quantification for Large Language Models via Ensemble-of-Ensembles}

\author{%
  \mdseries\rev{Wang Ma$^{1,*}$\quad Debarun Bhattacharjya$^{2,*}$\quad Junkyu Lee$^{3,*}$} \\
  \rev{Nhan H.\ Pham$^{4,*}$\quad Harsha Kokel$^{3,*}$\quad Qiang Ji$^{1}$} \\
  \rev{$^{1}$Rensselaer Polytechnic Institute\qquad $^{2}$Lirio\qquad $^{3}$IBM Software Innovation Lab} \\
  \rev{$^{4}$Independent Researcher} \\
  \rev{\texttt{\{maw6, jiq\}@rpi.edu, dbhattacharjya@lirio.com, Junkyu.Lee@ibm.com,}} \\
  \rev{\texttt{nhan.ph0407@gmail.com, harsha.kokel@ibm.com}}
}

\begin{document}

\maketitle
{\renewcommand{\thefootnote}{\fnsymbol{footnote}}%
 \footnotetext[1]{\rev{This work was done while Wang Ma was an intern, and Debarun Bhattacharjya, Junkyu Lee, Nhan H. Pham, and Harsha Kokel were researchers, at IBM Research.}}}
{\renewcommand{\thefootnote}{}\footnotetext{\rev{Code: \url{https://github.com/wma17/Black-box_UQ_for_LLMs_Ensemble_of_Ensemble}}}}

\begin{abstract}
Reliable use of large language models (LLMs) requires uncertainty estimates that separate ambiguity intrinsic to a prompt from genuine knowledge gaps. Bayesian and deep-ensemble methods deliver this aleatoric/epistemic decomposition in principle, but are computationally prohibitive at LLM scale and require white-box access. Existing black-box methods, in contrast, summarize to a single scalar that cannot make the distinction.
We introduce a \textbf{two-level ensemble} that bridges this gap. Stochastic decoding of a fixed prompt forms the inner ensemble; meaning-preserving semantic perturbations of the prompt form the outer ensemble. Variability is measured in a continuous embedding space, and the law of total covariance gives an exact \rev{within-/between-perturbation decomposition to estimate aleatoric and epistemic uncertainty (AU/EU)}; it requires only black-box sampling access.
On top of this estimator, we contribute three theoretical results that ground the framework: (i) a finite-sample bias identity for nested Monte Carlo with a closed-form bias-corrected epistemic estimator; (ii) a \emph{Bayesian bridge} bounding the gap between perturbation-population AU/EU and Bayesian targets; and (iii) a \emph{separation result} identifying a class of failure modes that any consistency-only black-box method provably misses while perturbation-based EU detects.
Across five short QA benchmarks and four instruction-tuned LLMs spanning two families and two scales, our estimator \rev{is competitive with} strong white-box baselines\rev{, and at a matched generation budget it matches leading black-box methods on error ranking while decomposing AU and EU}; controlled paired interventions \rev{show that AU and EU respond selectively to} distinct, actionable sources of failure: \rev{prompt ambiguity and model knowledge gaps}; and a long-form pilot on the NQ long-answer split shows the framework extends to claim-level factuality without retraining.
\end{abstract}

\section{Introduction \& Related Work}
\label{sec:intro}

\begin{revblock}
Large language models (LLMs) are widely deployed in search, code assistance, and scientific workflows, where uncertainty quantification (UQ) helps detect hallucinations and guide abstention. Crucially, when an LLM is uncertain, the appropriate action depends on \emph{why}: an ambiguous query calls for clarification, whereas a knowledge gap calls for retrieval or deferral. Existing black-box UQ methods, however, collapse output variability into a single consistency score that flags unreliability without diagnosing its source. We disentangle these two sources from black-box sampled generations via a two-level ensemble with exact finite-sample bias correction.
\end{revblock}

\textbf{From Bayesian UQ to black-box LLMs.}
Bayesian neural networks~\citep{gal2016dropout} and deep ensembles~\citep{lakshminarayanan2017ensemble} give a principled aleatoric/epistemic split via the law of total variance, but exact or approximate inference over billions of parameters is intractable for modern LLMs, where deployment often offers only \emph{black-box} sampling access.

\textbf{Related work.}
LLM-UQ divides into two strands: \emph{white-box} methods that exploit logits, hidden
states, or trained heads to score semantic uncertainty or detect hallucinations~\citep{kadavath2022l,Fadeeva2023LMPolygraph,kuhn2023se,farquhar202dse},
and \emph{black-box} methods that infer uncertainty from behavioral variability across
samples without internal access~\citep{manakul2023selfcheckgpt,Gao2024SPUQ,Xiao2025ConsistencyHypothesis,Yaldiz2025DoNotDesign,bhattacharjya2025simba};
surveys offer broader context~\citep{xia2025survey,liu2025uncertainty}. A growing line of
\emph{semantic-aware} black-box scorers sharpens semantic entropy via kernel-, graph-, or
volume-based objects on response embeddings~\citep{nikitin2024kle,nguyen2025snne,zhao2025sese,li2026semanticvolume,phillips2025geometric}.
Concurrent AU/EU-split attempts for LLMs use trained clarification heads, spectral
decompositions, variational ICL, paraphrase prompting, or conformal coverage~\citep{hou2023decomposing,walha2026spectral,jayasekera2025vud,cox2025mapping,noorani2025cpq}.
These works share our goal but emit a single scalar or are structurally \emph{one-level}
(perturbations \emph{or} decoding, not both), so they cannot expose the AU/EU split our two-level construction targets. The closest
perturbation-based black-box method is SPUQ~\citep{Gao2024SPUQ}, which perturbs prompts
but is one-level and reports a scalar consistency score. Outside the LLM setting,
\citet{wang2024epistemic} argued from a first-order optimization view that local input
perturbations approximate parameter perturbations for pre-trained vision models; we adapt
that intuition to LLMs and turn it from heuristic motivation into an explicit AU/EU
approximation bound for a nested estimator (Thm.~\ref{thm:end2end}\rev{; App.~\ref{app:wangji}}). Our two-level
construction is what enables both the AU/EU split (Sec.~\ref{sec:two_level_mc}) and the
separation from consistency-only UQ that \rev{fixed-prompt} methods provably cannot exhibit
(Sec.~\ref{sec:consistency_failure}).

\textbf{Contributions.}
We move black-box LLM-UQ beyond a scalar consistency score: our \textbf{two-level ensemble} (Fig.~\ref{fig:method_overview}) admits an exact \rev{within-/between-perturbation} decomposition of embedded-generation covariance \rev{to estimate AU and EU}, and we equip the decomposition with theory that makes it both estimable and falsifiable. Concretely:
\begin{enumerate}[leftmargin=*, noitemsep, topsep=0pt]
    \item \textbf{Exact \rev{within-/between-perturbation} decomposition with finite-$M$ bias correction \rev{for AU/EU estimation}.} An exact \rev{within-/between-perturbation} split of embedded-generation covariance, plus a closed-form bias correction that makes the epistemic estimator unbiased at any inner sample size $M\!\geq\!2$ (Sec.~\ref{sec:two_level_mc}).
    \item \textbf{Bayesian bridge \rev{to parameter-posterior AU/EU} with diagnostic conditions.} Defect bounds connecting perturbation-population AU/EU to Bayesian AU/EU through three local conditions that we estimate as diagnostics from the same outputs the estimator already collects (Sec.~\ref{sec:bayes_bridge_main}).
    \item \textbf{Provable separation from consistency-only UQ.} A class of failures any consistency-only method provably misses while perturbation-based EU detects\rev{; in paired interventions, prompt clarification significantly reduces EU while leaving fixed-prompt baselines unchanged} (Sec.~\ref{sec:consistency_failure},~\ref{subsec:au_eu_validation}).
    \item \textbf{Empirical results.} On five short-form QA benchmarks and four instruction-tuned LLMs spanning two families and two scales, the estimator \rev{is competitive with} strong white-box baselines (Sec.~\ref{subsec:main_results}); \rev{at a matched generation budget, it matches leading one-level black-box methods on error ranking while decomposing AU and EU (Sec.~\ref{subsec:matched_compute});} paired interventions on AmbigQA and Natural Questions \rev{show that AU and EU respond selectively to} distinct, actionable sources of failure (Sec.~\ref{subsec:au_eu_validation}); the predicted finite-$M$ bias correction is verified empirically (Sec.~\ref{subsec:bias_correction}); a long-form pilot on the NQ long-answer split shows the framework extends to claim-level factuality without retraining (Sec.~\ref{subsec:longform}); and ablations confirm stability under embedding choice and sampling budget (Sec.~\ref{sec:ablation_main}).
\end{enumerate}

\section{Background}

\noindent\textbf{Bayesian uncertainty decomposition.}
Bayesian neural networks~\citep{goan2020bnn,gal2016dropout} treat parameters $\theta$ as
random variables with posterior $p(\theta\mid D)$, yielding a predictive
$p(y\mid x,D)=\mathbb E_{p(\theta\mid D)}[p(y\mid x,\theta)]$ that decomposes total
predictive uncertainty into \emph{aleatoric} (irreducible data/decoding noise) and
\emph{epistemic} (reducible model knowledge) components~\citep{kendall2017uncertainties,hullermeier2021hullermeier2021auandeu,depeweg2018decomposition}.
For continuous outputs parameterized by $(\mu_\theta(x),\sigma_\theta^2(x))$~\citep{amini2020regression},
{\small
\[
\mathrm{AU}=\mathbb E_{p(\theta\mid D)}[\sigma_\theta^2(x)],\quad
\mathrm{EU}=\mathrm{Var}_{p(\theta\mid D)}[\mu_\theta(x)],\quad
\mathrm{TU}=\mathrm{AU}+\mathrm{EU}.
\]
}
This regression form is the conceptual target we adapt: we replace scalar mean/variance
with embedding-space mean/covariance so the same identity applies to free-form generation,
and we replace the parameter posterior (intractable for billion-parameter LLMs) with a
perturbation distribution we can sample from.

\noindent\textbf{Input perturbation as a black-box surrogate.}
Direct sampling from $p(\theta\mid D)$ is infeasible for pre-trained black-box LLMs. As
a methodological motivation, the local equivalence of \citet{wang2024epistemic}
(Prop.\ 3.3) approximates small parameter variations by structured input-embedding
perturbations: $f(x{+}\Delta x,\theta)\approx f(x,\theta{+}\Delta\theta)$, so input
perturbations induce output changes similar to those a small parameter
perturbation would induce. We treat this as a \emph{proxy} for $p(y\mid x,D)$ at fixed
parameters: input perturbations serve as a tractable surrogate for an intractable parameter
posterior. The exact relationship is deferred to Thm.~\ref{thm:end2end}
(Sec.~\ref{sec:bayes_bridge_main}), which formalizes the local conditions and bounds the
gap between perturbation-population AU/EU and Bayesian AU/EU. Throughout, we operate on
three distinct objects: a Bayesian target, an operational perturbation-population
quantity, and a finite-sample estimator; the estimator we build in
Sec.~\ref{sec:method} is defined in perturbation-population terms, and the Bayesian
interpretation \rev{is established under the local bridge conditions of Sec.~\ref{sec:bayes_bridge_main}}.

\section{Proposed Method: Uncertainty Quantification via Ensemble-of-Ensembles}
\label{sec:method}

\begin{figure}[t]
\centering
\includegraphics[width=\linewidth]{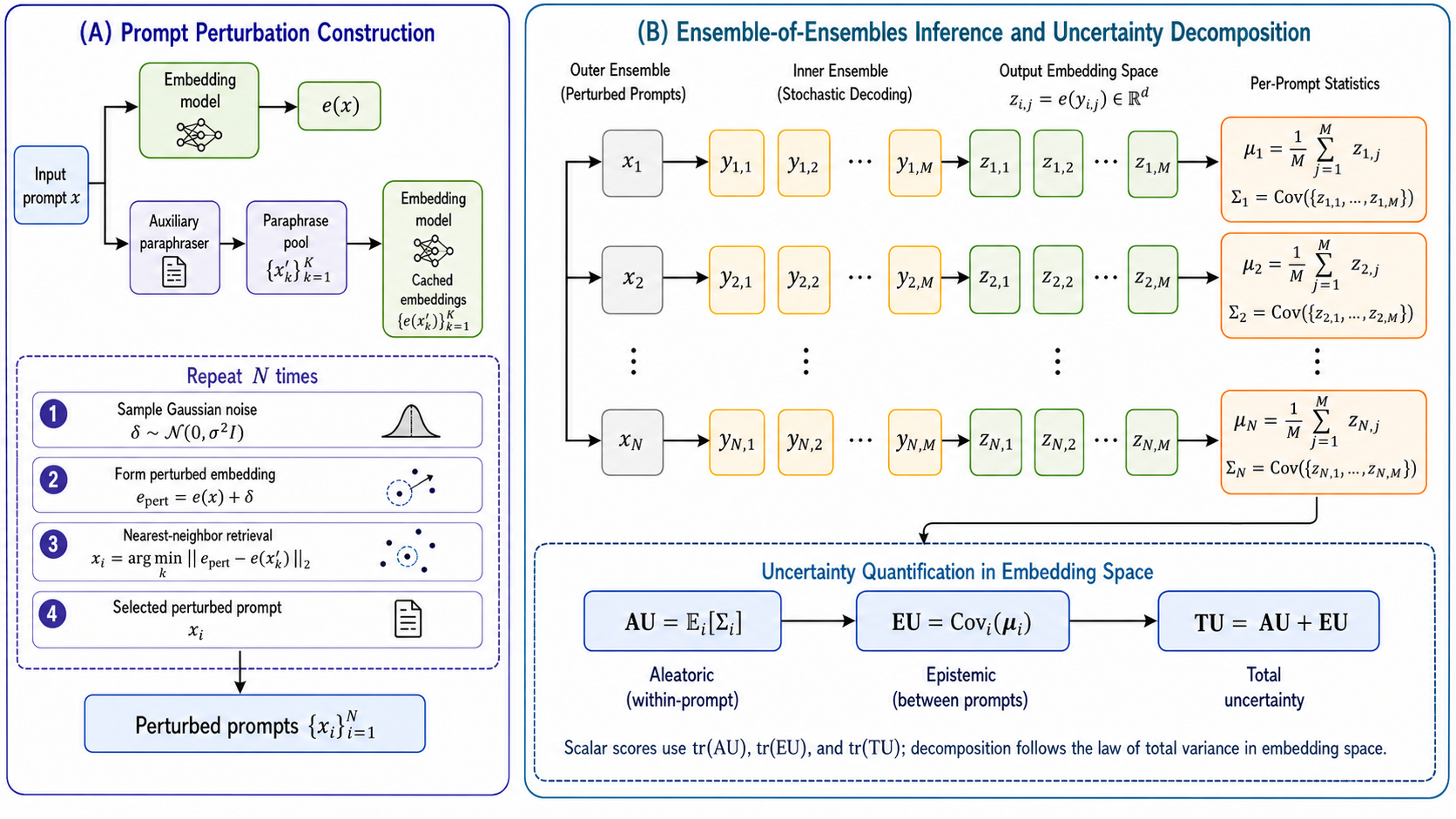}
\caption{\textbf{Two-level ensemble pipeline.}
\textbf{(A)} Localized semantic perturbation: snap noisy query embeddings to nearest of
$K_{\mathrm{pool}}$ cached paraphrases ($N$ times $\to$ outer ensemble).
\textbf{(B)} Per-prompt $M$ stochastic decodes; embedding-space statistics
$(\hat\mu_i,\hat\Sigma_i)$ feed the law-of-total-covariance AU/EU/TU split.}
\label{fig:method_overview}
\end{figure}

We propose a two-level black-box UQ framework for LLMs that requires neither retraining
nor access to token-level logits. An \emph{outer} ensemble draws meaning-preserving
semantic perturbations $\Delta\!\sim\!p_\Delta$ of the input $x$, yielding
$x_\Delta\!:=\!T(x,\Delta)$; an \emph{inner} ensemble draws stochastic decodes
$Y\!\sim\!p(\cdot\!\mid\!x_\Delta)$ at fixed deployed parameters $\theta^\star$. We measure
variability in a sentence-embedding space $Z=e(Y)\in\mathbb{R}^d$ (assume
$\mathbb{E}\|Z\|^2<\infty$), where the law of total covariance gives
{\small
\begin{equation}
\underbrace{\operatorname{Cov}(Z)}_{U_T(x)}
=\underbrace{\mathbb{E}_{\Delta}[\operatorname{Cov}(Z\!\mid\!\Delta)]}_{U_A(x)\;\text{(within-perturbation)}}
+\underbrace{\operatorname{Cov}_{\Delta}(\mathbb{E}[Z\!\mid\!\Delta])}_{U_E(x)\;\text{(between-perturbation)}}.
\label{eq:total_covariance}
\end{equation}
}
We adopt Eq.~\eqref{eq:total_covariance} as the definition of the
\emph{perturbation-population} uncertainties $(U_A,U_E,U_T)\equiv(\mathcal U_A^\Delta,\mathcal U_E^\Delta,\mathcal U_T^\Delta)$.
The interpretation mirrors the Bayesian split: $U_A$ aggregates within-perturbation
variability (decoding stochasticity at a fixed prompt and ambiguity in the prompt
itself), while $U_E$ measures between-perturbation disagreement of conditional means,
i.e.\ how much the model's expected behavior changes under semantically equivalent
phrasings. The corresponding Bayesian targets and finite-sample estimators are formally
distinguished in Sec.~\ref{sec:bayes_bridge_main} (full definitions in
App.~\ref{app:def_three_flavors}\rev{; summary in App.~Tab.~\ref{tab:notation_flavors}}). Fig.~\ref{fig:method_overview} summarizes the full
pipeline; pseudocode is in App.~Alg.~\ref{alg:two_level_uq}.

\noindent\textbf{Outer ensemble: meaning-preserving perturbations.}\label{sec:perturbation}
We obtain $N$ semantically equivalent variants $\{x_1,\dots,x_N\}$ of $x$ by a two-stage
localized procedure adapted from~\citet{hu2024localized}: its embedding-space
nearest-neighbor step provides the locality our Bayesian bridge
(Sec.~\ref{sec:bayes_bridge_main}) requires.

\noindent\textit{Stage 1 (offline, once per dataset).} An auxiliary paraphraser produces
a pool $\mathcal P_x = \{x'_k\}_{k=1}^{K_{\mathrm{pool}}}$ for each input $x$; we cache its embeddings
$\{e(x'_k)\}_{k=1}^{K_{\mathrm{pool}}}$.

\noindent\textit{Stage 2 (online, per query).} For $i=1,\dots,N$ we sample isotropic
Gaussian noise in embedding space and snap to the nearest cached paraphrase:
\begin{equation}
\delta_i \sim \mathcal N(0,\varepsilon^2 I_d),\qquad
x_i \;=\; \arg\min_{x'_k\in\mathcal P_x}\; D\!\big(e(x)+\delta_i,\;e(x'_k)\big),
\label{eq:perturb_step}
\end{equation}
where $D(\cdot,\cdot)$ is cosine distance. Since $N\!\ll\!K_{\mathrm{pool}}$, this draws a localized
neighborhood rather than enumerating the pool. The hyperparameters $(K_{\mathrm{pool}},\varepsilon)$
trade off locality (staying inside a region where the bridge linearization is plausible)
and coverage (paraphrase diversity); both are ablated in App.~\ref{sec:ablation_app}.
The AU/EU split is robust across perturbation families: random paraphrasing and SPUQ-style
noise both yield meaningful AU/EU under the two-level pipeline~(\rev{App.~}Tab.~\ref{tab:pert_methods}).

\noindent\textbf{Inner ensemble: stochastic decoding.}
For each $x_i$ we draw $M$ stochastic decodes $\{y_{i,j}\}_{j=1}^M\!\sim\!p(\cdot\!\mid\!x_i)$
under fixed sampling temperature, and embed them as $z_{i,j} = e(y_{i,j})$. The per-prompt
sample mean and covariance are
\begin{equation}
\hat\mu_i \;=\; \tfrac{1}{M}\sum_{j=1}^M z_{i,j},\qquad
\hat\Sigma_i \;=\; \tfrac{1}{M-1}\sum_{j=1}^M (z_{i,j}-\hat\mu_i)(z_{i,j}-\hat\mu_i)^\top.
\label{eq:inner_stats}
\end{equation}

\noindent\textbf{Two-level estimators and bias correction.}
Aggregating across the outer ensemble gives the empirical AU and (raw) EU estimators
{\small
\begin{equation}
\widehat{\mathcal U}_A(x) \;=\; \tfrac{1}{N}\sum_{i=1}^N \hat\Sigma_i,\qquad
\widehat{\mathcal U}_E^{\mathrm{raw}}(x) \;=\; \tfrac{1}{N-1}\sum_{i=1}^N (\hat\mu_i-\bar\mu)(\hat\mu_i-\bar\mu)^\top,\quad
\bar\mu = \tfrac{1}{N}\sum_i \hat\mu_i.
\label{eq:est_raw}
\end{equation}
}
Because $\hat\mu_i$ is itself an $M$-sample average,
$\widehat{\mathcal U}_E^{\mathrm{raw}}$ absorbs an $\tfrac{1}{M}$ leakage from
$\widehat{\mathcal U}_A$ (Prop.~\ref{prop:finiteM_bias}); we therefore use the
\emph{bias-corrected} estimator throughout,
{\small
\begin{equation}
\widehat{\mathcal U}_E(x) \;=\; \widehat{\mathcal U}_E^{\mathrm{raw}}(x) \;-\; \tfrac{1}{M}\,\widehat{\mathcal U}_A(x),\qquad
\widehat{\mathcal U}_T(x) \;=\; \widehat{\mathcal U}_A(x) + \widehat{\mathcal U}_E(x).
\label{eq:est_corrected}
\end{equation}
}
For correctness ranking we reduce each $d{\times}d$ matrix to a scalar by the trace,
$u_\bullet(x) = \operatorname{tr}\!\big(\widehat{\mathcal U}_\bullet(x)\big)$; \texttt{mean}, \texttt{max},
and \texttt{sum} reductions are ablated in App.~\ref{sec:ablation_app}. Per-query cost is
$NM$ LLM decodes plus a one-time paraphrase pool of size $K_{\mathrm{pool}}$ amortized across the dataset\rev{, and can be reduced via per-query adaptive stopping (App.~\ref{app:adaptive})}.
Throughout the paper, $K_{\mathrm{pool}}$ refers exclusively to this offline pool; it is distinct from the perturbation count of one-level baselines.

\section{Theoretical Results}
\label{sec:theory}


\begin{revblock}
We establish three theoretical results: (i) an exact finite-$M$ bias identity and concentration bound for the perturbation-population estimators (Pillar~1, Sec.~\ref{sec:two_level_mc}); (ii) a Bayesian bridge bounding the gap to parameter-posterior AU/EU via three data-estimable local diagnostics (Pillar~2, Sec.~\ref{sec:bayes_bridge_main}); and (iii) a separation theorem characterizing failure modes that fixed-prompt consistency methods provably miss (Pillar~3, Sec.~\ref{sec:consistency_failure}); proofs in App.~\ref{app:theory}.
\end{revblock}

\subsection{Pillar 1: bias-corrected finite-$M$ estimator}
\label{sec:two_level_mc}

Estimating $U_A,U_E$ from $N$ outer perturbations and $M$ inner decodes per perturbation
introduces a structural bias: the naive epistemic estimator absorbs $\tfrac{1}{M}$ of the
aleatoric covariance.

\begin{proposition}[Finite-$M$ bias of the epistemic estimator; informal]
\label{prop:finiteM_bias}
The naive estimator $\widehat U_E^{\mathrm{raw}}$ obtained as the sample covariance of
inner means is biased upward by exactly $\tfrac{1}{M}\,\mathcal U_A^\Delta(x)$ relative to
the perturbation-population EU, while $\widehat U_A^{(M)}$ is unbiased. The corrected
estimator $\widehat U_E^{(M)}:=\widehat U_E^{\mathrm{raw}}-\tfrac{1}{M}\widehat U_A^{(M)}$
is unbiased and concentrates at $O(\sqrt{\log(1/\delta)/N})$ plus lower-order $1/M$ terms.
Formal statement, constants, and proof in App.~\ref{app:mc_bounds}
(Prop.~\ref{prop:finiteM_bias_app}, Thm.~\ref{thm:two_level_mc_app}).
\end{proposition}

\noindent The leakage is empirically large at small $M$: $\rho(\widehat U_A,\widehat U_E^{\mathrm{raw}}){=}0.65$ at $M{=}3$ on TriviaQA, and the closed-form correction—a single matrix subtraction—drives it to $0.12$ at the same budget (Sec.~\ref{subsec:bias_correction}, Tab.~\ref{tab:bias_correction}).

\subsection{Pillar 2: Bayesian bridge with diagnostic conditions}
\label{sec:bayes_bridge_main}

The targets here are the \emph{Bayesian} AU/EU $\mathcal U_A^{\mathrm{Bayes}}(x){:=}\mathbb E_\theta[\Sigma_\theta(x)]$, $\mathcal U_E^{\mathrm{Bayes}}(x){:=}\mathrm{Cov}_\theta(m_\theta(x))$ defined over a parameter posterior $p(\theta\mid\mathcal D)$: the ground truth Bayesian UQ ultimately seeks.
Pillar~1 bounds the (estimator)$\to$(perturbation-population) gap from
Eq.~\eqref{eq:total_covariance}; the bridge below bounds the
(perturbation-population)$\to$(Bayesian) gap; composing the two yields an end-to-end bound
(formal definitions of the three flavors collected in App.~\ref{app:def_three_flavors};
proofs in App.~\ref{app:bridge_conditions}).

\begin{theorem}[Bayesian bridge for perturbation AU/EU; informal]
\label{thm:end2end}
Under three local conditions on the model around $x$ (first-order linearity of the
conditional mean, locally Gaussian posterior, and bounded semantic coverage defect),
the bias-corrected estimators of Pillar~1 satisfy
{\small
\[
\big\|\widehat U_A^{(M)}(x)-\mathcal U_A^{\mathrm{Bayes}}(x)\big\|_F
\;\le\;\underbrace{\text{MC error}}_{O(\sqrt{\log(1/\delta)/N})}
\;+\;\underbrace{\varepsilon_A(x)}_{\text{bridge defect}},
\]
}
and analogously for $\widehat U_E^{(M)}$. The defects $\varepsilon_A,\varepsilon_E$
shrink with the three conditions; precise statement, assumptions, and proof are in
App.~\ref{app:bridge_conditions} (Thms.~\ref{thm:EU_bridge}, \ref{thm:AU_bridge}).
\end{theorem}

The two model-side conditions align with existing LLM theory: local first-order linearity
matches the regime of linear-probing-then-fine-tuning~\citep{ren2023linearprobing}, and the
locally-Gaussian posterior is the Laplace approximation standard in the Bayesian-LoRA
line~\citep{yang2023bayesianlora}. We further \emph{measure} each defect from outputs the
estimator already collects: D1, an OLS residual of $\mu_\Delta(x)$ on $\Delta$ (local linearity); D2, a Henze--Zirkler statistic on the embedding clouds (local Gaussianity); D3, an MMD between the perturbation cloud and a held-out semantic reference (distributional coverage). Per-dataset values are in App.~\ref{sec:bridge_instructions} (Tab.~\ref{tab:bridge_diagnostic}). \rev{Consistent with first-order scaling, sweeping $\varepsilon\in[0.01,0.2]$ induces a nearly linear rise in EU ($R^2{=}0.978$) while AU shifts by $2.3\%$ (App.~\ref{app:strength}); instruction perturbations similarly act as implicit parameter updates under the dual gradient view of ICL~\citep{dai2023gptlearn} (App.~\ref{app:spuq_difference}).}

\subsection{Pillar 3: provable separation from consistency-only UQ}
\label{sec:consistency_failure}

\begin{theorem}[Perturbation--consistency separation; informal]
\label{thm:spurious_determinism}
Let $\mathcal{F}_{\mathrm{cons}}$ be \emph{any} uncertainty estimator that depends only
on repeated decoding at a fixed prompt (e.g.,
\citealp{wang2022self,manakul2023selfcheckgpt,kuhn2023se}; SPUQ~\citealp{Gao2024SPUQ} is
not in this class because it perturbs the prompt before decoding). On the class of
inputs where decoding is effectively deterministic but $\mu_\Delta(x)$ varies
non-trivially across semantic-preserving perturbations,
$\mathcal{F}_{\mathrm{cons}}(x)\!=\!0$ while perturbation EU is bounded \emph{away} from
zero. The class has positive measure whenever the model has prompt-format-driven
outputs. Formal statement and construction in App.~\ref{app:consistency_failure}.
\end{theorem}

\noindent Confident hallucinations, memorized templates, and prompt-format artifacts produce stable outputs at a fixed prompt, so any consistency-only score is exactly zero on them while perturbation EU \rev{remains} positive. \rev{In Sec.~\ref{subsec:au_eu_validation}, disambiguating a prompt significantly reduces EU while leaving all fixed-prompt baselines unchanged (Tab.~\ref{tab:intervention_baselines}).}

\section{Experiments}
\label{sec:experiments}

\subsection{Experimental setup}
\label{subsec:exp_setup}

\textbf{Experiment roadmap.}
We organize the experiments around \rev{six} questions.
\textbf{(Q1)} Under the standard UQ protocol, do our uncertainty scores rank incorrect
answers above correct ones?~(\S\ref{subsec:main_results}, Tab.~\ref{tab:main_full})
\rev{\textbf{(Q2)} At a matched budget, how does two-level sampling compare with one-level designs?~(\S\ref{subsec:matched_compute}, Tab.~\ref{tab:matched_compute})}
\textbf{(Q\rev{3})} Beyond correctness prediction, do AU and EU respond to \emph{different}
actionable sources of failure?~(\S\ref{subsec:au_eu_validation},
Tabs.~\ref{tab:au_eu_paired}\rev{,~\ref{tab:intervention_baselines}})
\textbf{(Q\rev{4})} Does the finite-$M$ bias correction predicted by Prop.~\ref{prop:finiteM_bias}
behave as the theory predicts?~(\S\ref{subsec:bias_correction},
Tab.~\ref{tab:bias_correction})
\textbf{(Q\rev{5})} Does the framework extend beyond short-form answers to claim-level long-form
factuality?~(\S\ref{subsec:longform}, Tab.~\ref{tab:longform_claim})
\textbf{(Q\rev{6})} Are the conclusions stable under key design choices such as the embedding
backbone and the two-level sampling budget?~(\S\ref{sec:ablation_main},
Tab.~\ref{tab:ablations_main})

\textbf{Tasks and datasets.}
We evaluate black-box uncertainty estimation on five short-form QA benchmarks:
\textbf{TriviaQA}~\citep{joshi2017triviaqa} (no-context), \textbf{TruthfulQA}~\citep{lin2021truthfulqa},
\textbf{GPQA}~\citep{rein2024gpqa}, \textbf{SciQA}~\citep{auer2023sciqa}, and
\textbf{CoQA}~\citep{reddy2019coqa}; we also include \textbf{AmbigQA}~\citep{min2020ambigqa} and
\textbf{Natural Questions}~\citep{kwiatkowski2019nq} for the paired AU/EU interventions in
\S\ref{subsec:au_eu_validation}, and the \textbf{NQ long-answer} split for the long-form
pilot in \S\ref{subsec:longform}. The goal is to test whether uncertainty reliably predicts
answer correctness.

\textbf{Models.}
We use four instruction-tuned LLMs spanning two families and two scales:
\textbf{LLaMA-3-70B}, \textbf{LLaMA-3-8B}, \textbf{Qwen2.5-72B}, and
\textbf{Qwen2.5-7B}~\citep{grattafiori2024llama,bai2025qwen2}. The main paper presents the
smaller two (LLaMA-3-8B, Qwen2.5-7B) in Tab.~\ref{tab:main_full}; 70B/72B results are in
App.~Tab.~\ref{tab:big_models}.

\textbf{Correctness labeling.}
Two complementary automatic criteria:
(1) \textbf{Cosine similarity} between sentence embeddings of generated vs.\ reference
answers (threshold $>0.8$),
and (2) \textbf{ROUGE-L} overlap (threshold $>0.5$)~\citep{reimers2019sentencebert,lin2004rouge}.
Both criteria are reported to reduce metric-specific artifacts. \rev{In \S\ref{subsec:matched_compute} and App.~\ref{app:audit}, we additionally evaluate under \textbf{Exact Match (EM)} on TriviaQA as an encoder-independent correctness criterion.} 

\textbf{Uncertainty measures.}
We report AUROC for correctness prediction using \textbf{aleatoric} (AU), \textbf{epistemic}
(EU), and \textbf{total} uncertainty (TU) as defined in Eq.~\eqref{eq:total_covariance};
following the UQ convention, the positive class is \emph{incorrect} and the scoring signal
is the (uncorrected sign) uncertainty value, so AUROC measures how well higher uncertainty
ranks errors above correct answers.

\textbf{Baselines.}
Black-box: D-SE~\citep{kuhn2023se}, Self-Consistency~\citep{wang2022self},
SIMBA~\citep{bhattacharjya2025simba}, SelfCheckGPT~\citep{manakul2023selfcheckgpt},
SPUQ~\citep{Gao2024SPUQ}, and a P(True) probe~\citep{kadavath2022l}. White-box:
G-NLL~\citep{aichberger2024gnll}, PE / LN-PE~\citep{malinin2020pe}, and
SE / LN-SE~\citep{farquhar202dse,kuhn2023se}. The long-form pilot
(\S\ref{subsec:longform}) additionally compares against five long-form-specific scorers
from~\citet{bouchard2026finegrained}: avgEntail, NonContradiction, UQLM cosine,
plus SelfCheckGPT and ROUGE-L (cons.) at long-form. Full baseline configurations are in
App.~\ref{app:exp_settings}; SPUQ uses its three perturbation modes with $K_{\mathrm{SPUQ}}{=}10$
perturbations (SPUQ's perturbation count, distinct from our offline pool size $K_{\mathrm{pool}}$;
$K_{\mathrm{SPUQ}}$ plays the role of our outer count $N$) and pairwise ROUGE-L
aggregation~\citep{Gao2024SPUQ} (see App.~\ref{app:spuq_difference}).

\noindent\textbf{Sampling budget.}
For the headline AUROC table (Tab.~\ref{tab:main_full}) we use $N{=}M{=}20$ \rev{(budget ablation in App.~\ref{sec:budget_ablation_appendix})}; the AU/EU
validation experiments (\S\ref{subsec:au_eu_validation}) and all other ablations use the
compute-efficient default $N{=}M{=}10$, and the long-form pilot (\S\ref{subsec:longform})
uses the further-reduced $N{=}M{=}5$ for efficiency over a separately-extracted claim set
(see App.~\ref{app:longform}). When comparing to SPUQ \rev{in App.~\ref{app:spuq_difference}}, we match the \emph{outer perturbation
count} ($K_{\mathrm{SPUQ}}{=}N{=}10$); our method then incurs the additional cost of an inner ensemble
($M$ stochastic decodes per perturbation) which is intrinsic to the two-level decomposition
and not present in any one-level baseline. Per-method per-question call counts and wall-clock
times are tabulated in App.~Tab.~\ref{tab: time}.

\rev{\textbf{Other settings.}}
Decoding temperature $T$ is treated as an
internal property of the deployed LLM, not an estimator knob: a targeted T-sweep
(App.~\ref{app:temperature_stability}) shows EU AUROC stable across $T\!\in\![0.3,1.5]$. The
perturbation noise scale $\varepsilon{=}0.1$ comes from the noise-level ablation
(App.~\ref{sec:noise_level}); the paraphrase pool size is $K_{\mathrm{pool}}{=}500$
\rev{for short-form QA} (the ablation in App.~Tab.~\ref{tab:Np} shows $K_{\mathrm{pool}}{\in}[20,100]$ is
also competitive).

\begin{table*}[!t]
\centering
\caption{\textbf{Headline correctness-prediction AUROC} across five short-form benchmarks.
Cos: cosine $>0.8$; R-L: ROUGE-L $>0.5$. $N{=}M{=}20$. \textbf{Bold}: overall best;
\underline{underlined}: best black-box; \colorbox{gray!40}{\strut gray}: best white-box.
Std.\ devs.\ and 70B/72B results in
App.~Tabs.~\ref{tab:main},\ref{gpqa_sciqa_coqa},\ref{tab:big_models}.}
\label{tab:main_full}
\resizebox{\textwidth}{!}{
\setlength{\tabcolsep}{3.5pt} 
\renewcommand{\arraystretch}{1.1}
\begin{tabular}{ll | cc | cc | cc | cc | cc || cc | cc | cc | cc | cc}
\toprule
& \multirow{3}{*}{\textbf{Method}} & \multicolumn{10}{c||}{\textbf{LLaMA-3-8B}} & \multicolumn{10}{c}{\textbf{Qwen2.5-7B}} \\
\cmidrule(lr){3-12} \cmidrule(lr){13-22}
& & \multicolumn{2}{c|}{TriviaQA} & \multicolumn{2}{c|}{TruthfulQA} & \multicolumn{2}{c|}{SciQA} & \multicolumn{2}{c|}{GPQA} & \multicolumn{2}{c||}{CoQA} & \multicolumn{2}{c|}{TriviaQA} & \multicolumn{2}{c|}{TruthfulQA} & \multicolumn{2}{c|}{SciQA} & \multicolumn{2}{c|}{GPQA} & \multicolumn{2}{c}{CoQA} \\
& & Cos & R-L & Cos & R-L & Cos & R-L & Cos & R-L & Cos & R-L & Cos & R-L & Cos & R-L & Cos & R-L & Cos & R-L & Cos & R-L \\
\midrule

\multirow{3}{*}{\rotatebox{90}{\textit{Ours}}}   & AU & \underline{\textbf{0.799}} & \underline{\textbf{0.845}} & 0.782 & 0.694 & \underline{0.735} & \underline{\textbf{0.757}} & \underline{\textbf{0.770}} & \underline{\textbf{0.839}} & 0.738 & 0.735 & \underline{\textbf{0.773}} & \underline{\textbf{0.815}} & \underline{\textbf{0.812}} & \underline{\textbf{0.731}} & 0.516 & 0.508 & 0.506 & 0.758 & 0.585 & 0.641 \\
& EU & 0.585 & 0.588 & 0.745 & 0.647 & 0.534 & 0.548 & 0.746 & 0.590 & 0.630 & 0.631 & 0.570 & 0.547 & 0.684 & 0.583 & 0.502 & 0.501 & 0.612 & \underline{\textbf{0.949}} & 0.545 & 0.516 \\
& TU & 0.784 & 0.826 & \underline{\textbf{0.799}} & \underline{\textbf{0.715}} & 0.680 & 0.703 & 0.671 & 0.769 & 0.723 & 0.724 & 0.751 & 0.782 & 0.809 & 0.719 & 0.515 & 0.512 & 0.536 & 0.914 & 0.548 & 0.625 \\
\midrule
\multirow{6}{*}{\textit{\rotatebox{90}{Black-box}}}  & D-SE & 0.742 & 0.784 & 0.511 & 0.550 & 0.698 & 0.709 & 0.641 & 0.667 & 0.648 & 0.659 & 0.647 & 0.667 & 0.530 & 0.540 & 0.503 & 0.503 & \underline{0.757} & 0.672 & 0.615 & 0.531 \\
& S-Consis. & 0.740 & 0.782 & 0.524 & 0.555 & 0.712 & 0.706 & 0.602 & 0.621 & 0.641 & 0.602 & 0.647 & 0.667 & 0.536 & 0.542 & 0.658 & 0.660 & 0.555 & 0.544 & 0.529 & 0.515 \\
& SIMBA & 0.749 & 0.785 & 0.506 & 0.520 & 0.732 & 0.722 & 0.635 & 0.697 & 0.583 & 0.537 & 0.632 & 0.663 & 0.598 & 0.643 & \underline{\textbf{0.714}} & \underline{0.695} & 0.546 & 0.513 & 0.566 & 0.610 \\
& SelfCheckGPT & 0.754 & 0.789 & 0.517 & 0.527 & 0.723 & 0.713 & 0.639 & 0.690 & 0.566 & 0.523 & 0.641 & 0.650 & 0.592 & 0.651 & 0.693 & 0.676 & 0.548 & 0.503 & \underline{\textbf{0.651}} & \underline{0.685} \\
& SPUQ & 0.615 & 0.641 & 0.602 & 0.583 & 0.659 & 0.651 & 0.554 & 0.510 & 0.505 & 0.505 & 0.616 & 0.640 & 0.544 & 0.590 & 0.646 & 0.634 & 0.592 & 0.512 & 0.506 & 0.513 \\
& P(True) & 0.632 & 0.652 & 0.600 & 0.639 & 0.566 & 0.568 & 0.524 & 0.623 & \underline{\textbf{0.788}} & \underline{\textbf{0.774}} & 0.603 & 0.638 & 0.763 & 0.717 & 0.680 & 0.681 & 0.561 & 0.519 & 0.520 & 0.526 \\
\midrule
\multirow{5}{*}{\rotatebox{90}{ \textit{White-box}}} & G-NLL & \cellcolor{gray!40}0.786 & \cellcolor{gray!40}0.798 & 0.546 & 0.549 & \cellcolor{gray!40}\textbf{0.770} & \cellcolor{gray!40}0.756 & 0.506 & 0.543 & \cellcolor{gray!40}0.707 & 0.664 & 0.646 & 0.662 & \cellcolor{gray!40}0.611 & \cellcolor{gray!40}0.700 & 0.524 & 0.626 & \cellcolor{gray!40}\textbf{0.850} & 0.748 & 0.501 & \cellcolor{gray!40}\textbf{0.827} \\
& PE & 0.710 & 0.732 & \cellcolor{gray!40}0.578 & \cellcolor{gray!40}0.576 & 0.725 & 0.719 & \cellcolor{gray!40}0.680 & 0.542 & 0.667 & 0.636 & 0.644 & 0.657 & 0.562 & 0.622 & 0.552 & 0.675 & 0.610 & \cellcolor{gray!40}0.869 & 0.511 & 0.711 \\
& LN-PE & 0.761 & 0.796 & 0.496 & 0.496 & 0.737 & 0.725 & 0.639 & 0.549 & 0.658 & 0.654 & 0.637 & 0.658 & 0.570 & 0.623 & \cellcolor{gray!40}0.568 & 0.650 & 0.625 & 0.787 & 0.526 & 0.808 \\
& SE & 0.739 & 0.774 & 0.560 & 0.525 & 0.742 & 0.742 & 0.601 & 0.511 & 0.690 & \cellcolor{gray!40}0.675 & \cellcolor{gray!40}0.657 & \cellcolor{gray!40}0.674 & 0.597 & 0.625 & 0.508 & \cellcolor{gray!40}\textbf{0.723} & 0.742 & 0.608 & \cellcolor{gray!40}0.589 & 0.667 \\
& LN-SE & 0.742 & 0.777 & 0.543 & 0.492 & 0.735 & 0.737 & 0.623 & \cellcolor{gray!40}0.608 & 0.663 & 0.664 & 0.653 & 0.670 & 0.573 & 0.608 & 0.534 & 0.692 & 0.826 & 0.637 & 0.566 & 0.543 \\
\bottomrule
\end{tabular}
}
\end{table*}

\begin{table}[t]
\revfloat
\centering
\caption{\textbf{Matched generation budget ($B$) comparison} on TriviaQA (EM label; AUROC with incorrect as positive; $400$ questions per model). \textbf{Bold}: best per column; \underline{underlined}: best black-box. $^\dagger$Requires token log-probabilities (white-box). Bootstrap $95\%$ CIs and additional baselines: App.~Tabs.~\ref{tab:matched_compute_ci_llama8b},~\ref{tab:matched_compute_ci_qwen7b}.}
\label{tab:matched_compute}
\scriptsize
\setlength{\tabcolsep}{4pt}
\renewcommand{\arraystretch}{1.05}
\begin{tabular}{l|ccc|ccc}
\toprule
& \multicolumn{3}{c|}{\textbf{LLaMA-3-8B}} & \multicolumn{3}{c}{\textbf{Qwen2.5-7B}} \\
\textbf{Method} & $B{=}10$ & $25$ & $100$ & $10$ & $25$ & $100$ \\
\midrule
SPUQ (one-level, perturbed prompts) & \underline{\textbf{0.756}} & \underline{\textbf{0.769}} & 0.778 & 0.759 & 0.764 & 0.782 \\
Embedding variance (fixed prompt) & 0.754 & 0.762 & \underline{\textbf{0.782}} & \underline{\textbf{0.774}} & \underline{0.801} & \underline{0.816} \\
Perturbation-only ($N{=}B$, $M{=}1$) & 0.754 & 0.667 & 0.657 & 0.743 & 0.553 & 0.549 \\
Decoupled (separate components, summed) & \underline{\textbf{0.756}} & 0.738 & 0.750 & 0.765 & 0.712 & 0.731 \\
Semantic Density, uniform & 0.600 & 0.594 & 0.586 & 0.721 & 0.776 & 0.765 \\
Semantic Density, likelihood-weighted$^\dagger$ & 0.663 & 0.651 & 0.644 & \textbf{0.774} & \textbf{0.821} & \textbf{0.821} \\
\midrule
Ours: AU (within) & 0.744 & 0.752 & 0.766 & 0.772 & 0.789 & 0.788 \\
Ours: TU (total) & 0.744 & 0.737 & 0.755 & 0.736 & 0.734 & 0.751 \\
\bottomrule
\end{tabular}
\end{table}

\subsection{Headline AUROC: short-form correctness prediction}
\label{subsec:main_results}

Tab.~\ref{tab:main_full} reports correctness-prediction AUROC against strong white-box and
black-box baselines under the standard protocol; the behavioral semantics of the AU/EU
split are tested in \S\ref{subsec:au_eu_validation}.

\textbf{(i) AU and TU are the strongest black-box signals overall}\rev{, competitive with}
white-box methods that require token logits: AU reaches $0.845$/$0.815$ under ROUGE-L on
TriviaQA (LLaMA-3-8B / Qwen2.5-7B), the best black-box score in both columns, and AU \rev{or} TU
\rev{achieves the best black-box score in $12$ and the best overall score in $11$} of $20$ benchmark$\times$correctness$\times$model cells.

\textbf{(ii) AU is robust where prior black-box baselines collapse:} on TruthfulQA, SPUQ
($\le 0.602$) and SIMBA\rev{\ on LLaMA-3-8B} ($\le 0.520$) drift toward chance while AU remains informative.

\textbf{(iii) EU contributes a complementary signal.} EU targets knowledge-gap detection
rather than generic correctness ranking, \rev{as validated} by paired intervention in
Tab.~\ref{tab:au_eu_paired}; on Qwen2.5-7B / GPQA / ROUGE-L it is the single best score
across \emph{all} methods ($0.949$), and on the \rev{three} hard benchmarks (GPQA, SciQA,
CoQA) AU and EU together hold the top black-box position in \rev{$5/12$} Cos/R-L columns. The
70B/72B counterpart \rev{on TriviaQA} (App.~Tab.~\ref{tab:big_models}) extends the same pattern.

\begin{table}[t]
\centering
\caption{\textbf{$2{\times}2$ paired interventions} (LLaMA-3-8B, $N{=}M{=}10$, $1{,}000$
examples). All four contrasts significant at $p{<}0.01$. \rev{$\Delta$: change in AUROC. Protocol} and Qwen
replication: App.~\ref{app:au_eu_validation}.}
\label{tab:au_eu_paired}
\scriptsize
\setlength{\tabcolsep}{4pt}
\renewcommand{\arraystretch}{1.05}
\begin{tabular}{ll|cc|c}
\toprule
\textbf{Test} & \textbf{Metric} & $\Delta\mathrm{AU}$ & $\Delta\mathrm{EU}$ & \textbf{Contrast ($z$, $p$)} \\
\midrule
\multirow{2}{*}{\textbf{AmbigQA} (clarify $\Rightarrow$ moves AU)}
& Cosine  & $\mathbf{+0.058}\!\pm\!0.005$ & $+0.021\!\pm\!0.007$ & $\Delta\mathrm{AU}\!-\!\Delta\mathrm{EU}{=}\!+0.037$ ($z{=}2.74$, $p{=}0.006$) \\
& ROUGE-L & $\mathbf{+0.044}\!\pm\!0.005$ & $+0.012\!\pm\!0.007$ & $\Delta\mathrm{AU}\!-\!\Delta\mathrm{EU}{=}\!+0.032$ ($z{=}4.18$, $p{<}0.001$) \\
\midrule
\multirow{2}{*}{\textbf{NQ} (evidence $\Rightarrow$ moves EU)}
& Cosine  & $+0.039\!\pm\!0.006$ & $\mathbf{+0.069}\!\pm\!0.007$ & $\Delta\mathrm{EU}\!-\!\Delta\mathrm{AU}{=}\!+0.030$ ($z{=}4.28$, $p{<}0.001$) \\
& ROUGE-L & $+0.031\!\pm\!0.006$ & $\mathbf{+0.078}\!\pm\!0.007$ & $\Delta\mathrm{EU}\!-\!\Delta\mathrm{AU}{=}\!+0.047$ ($z{=}6.56$, $p{<}0.001$) \\
\bottomrule
\end{tabular}
\end{table}

\begin{revblock}
\subsection{Matched-budget comparison against one-level designs}
\label{subsec:matched_compute}

Tab.~\ref{tab:matched_compute} fixes the per-question generation budget $B\!\in\!\{10,25,100\}$ across $400$ TriviaQA questions per model (EM labels) and compares alternative allocations: all $B$ decodes at the original prompt (\emph{fixed prompt}); one decode per paraphrase across $B$ paraphrases (\emph{perturbation-only}, $N{=}B$, $M{=}1$); an equal split between the two with summed variances (\emph{decoupled}); SPUQ; Semantic Density~\citep{qiu2024semantic}; and our nested estimator with $N{\times}M{=}5{\times}2$, $5{\times}5$, and $10{\times}10$.

\textbf{(i) Competitive error ranking with an explicit AU/EU split.} At every budget $B$, SPUQ, fixed-prompt embedding variance, and our AU lie within $0.04$ AUROC of one another with overlapping $95\%$ CIs (our TU is within $0.01$--$0.03$ on LLaMA-3-8B and $0.04$--$0.07$ on Qwen2.5-7B of the top black-box score), matching one-level error ranking while decomposing AU and EU.

\textbf{(ii) Stability across budgets.} Our AU and TU vary by less than $0.03$ across all budgets $B$. By contrast, perturbation-only ($N{=}B, M{=}1$) drops from $0.754$ to $0.667$ on LLaMA-3-8B and from $0.743$ to $0.553$ on Qwen2.5-7B between $B{=}10$ and $25$, as $M{=}1$ lacks inner averaging and draws from an extended paraphrase pool at $B{\ge}25$ (App.~\ref{app:matched_compute}).

\textbf{(iii) Semantic Density.} Our AU outperforms uniform Semantic Density in every column, whereas the likelihood-weighted variant leads on Qwen2.5-7B at $B{\ge}25$ by utilizing white-box token probabilities.
\end{revblock}

\begin{table}[t]
\begin{minipage}[t]{0.53\linewidth}
\revfloat
\centering
\caption{\textbf{Standardized score shifts under paired interventions} (LLaMA-3-8B; $1{,}000$ pairs per intervention): mean of $(\text{after}-\text{before})$ divided by the SD of per-question changes. \textbf{Bold}: $95\%$ bootstrap CI excludes zero.}
\label{tab:intervention_baselines}
\scriptsize
\setlength{\tabcolsep}{3pt}
\renewcommand{\arraystretch}{1.05}
\resizebox{\linewidth}{!}{%
\begin{tabular}{l|cc}
\toprule
\textbf{Score} & \textbf{Clarify (AmbigQA)} & \textbf{Evidence (NQ)} \\
\midrule
Ours: AU (within) & $-0.013$ & $\mathbf{-0.610}$ \\
Ours: EU (between) & $\mathbf{-0.218}$ & $\mathbf{-0.604}$ \\
Ours: TU (total) & $\mathbf{-0.156}$ & $\mathbf{-0.725}$ \\
\midrule
Embedding variance (fixed prompt) & $-0.038$ & $\mathbf{-0.546}$ \\
Self-Consistency & $-0.022$ & $\mathbf{-0.518}$ \\
D-SE & $-0.031$ & $\mathbf{-0.576}$ \\
Semantic Density, uniform & $+0.018$ & $\mathbf{+0.079}$ \\
SPUQ (perturbed prompts) & $\mathbf{-0.106}$ & $\mathbf{-0.836}$ \\
\bottomrule
\end{tabular}}
\end{minipage}\hfill
\begin{minipage}[t]{0.44\linewidth}
\centering
\caption{\textbf{Bias correction validated.} Spearman $\rho$ between AU and EU on TriviaQA
(LLaMA-3-8B, $N{=}10$, Cosine).}
\label{tab:bias_correction}
\scriptsize
\setlength{\tabcolsep}{3pt}
\renewcommand{\arraystretch}{1.05}
\resizebox{\linewidth}{!}{%
\begin{tabular}{c|ccccc}
\toprule
$M$ & 3 & 5 & 10 & 20 & 30 \\
\midrule
$\rho(\widehat U_A,\widehat U_E^{\mathrm{raw}})$
   & $0.647$ & $0.521$ & $0.393$ & $0.312$ & $0.283$ \\
$\rho(\widehat U_A,\widehat U_E^{\mathrm{corr}})$
   & $0.115$ & $0.169$ & $0.199$ & $0.211$ & $0.213$ \\
Gap & $0.532$ & $0.352$ & $0.194$ & $0.101$ & $0.070$ \\
\bottomrule
\end{tabular}}
\end{minipage}
\end{table}

\subsection{Behavioral validation of the AU/EU split}
\label{subsec:au_eu_validation}

We test the AU/EU split behaviorally with a $2{\times}2$ paired-intervention protocol that
targets one source of uncertainty at a time, directly probing the actionability of the
decomposition—a property correctness-prediction AUROC alone cannot reveal. All experiments use LLaMA-3-8B with
$N{=}M{=}10$ on $1{,}000$ examples (Qwen2.5-7B replication in App.~\ref{app:au_eu_validation}). The paired statistic differences the
per-example AUROC contribution between treatment and control, yielding a Wald $z$-test for
the contrast $\Delta\mathrm{AU}-\Delta\mathrm{EU}$ (or its twin).

\noindent\textbf{Paired interventions ($2{\times}2$).}
On AmbigQA~\citep{min2020ambigqa}, adding a gold clarification removes ambiguity; on
Natural Questions~\citep{kwiatkowski2019nq}, adding the gold evidence passage closes the
knowledge gap (perturbations applied only to the question, not the passage). The framework
predicts clarification moves \emph{AU} \rev{more than} EU, and evidence moves \emph{EU} \rev{more than}
AU. Tab.~\ref{tab:au_eu_paired} confirms both: the contrast is significant in every cell
($p{<}0.01$), and the absolute change on the predicted-target head is \rev{about $2$--$4{\times}$}
the change on the off-target head in every row. 

\begin{revblock}
\noindent\textbf{Score-level shifts and fixed-prompt baselines.}
Whereas Tab.~\ref{tab:au_eu_paired} evaluates discriminative error-ranking gains ($\Delta\text{AUROC}$), Tab.~\ref{tab:intervention_baselines} measures standardized shifts in raw uncertainty magnitude under both interventions across our three components, four fixed-prompt baselines, and SPUQ (LLaMA-3-8B, $10$ decodes per baseline; App.~\ref{app:intervention_baselines}).

Supplying evidence on NQ significantly reduces uncertainty across all methods except uniform Semantic Density. By contrast, clarifying the prompt on AmbigQA significantly reduces \emph{only} the perturbation-sensitive scores—EU ($-0.218$), TU ($-0.156$), and SPUQ ($-0.106$)—while all fixed-prompt baselines remain statistically unchanged, corroborating Thm.~\ref{thm:spurious_determinism}.

Because SPUQ emits a single scalar that drops under both interventions, it cannot distinguish prompt ambiguity from missing knowledge. Our decomposition separates the two: clarification resolves cross-paraphrase divergence over competing interpretations (lowering EU magnitude in Tab.~\ref{tab:intervention_baselines}) while sharpening within-prompt variance to better rank residual errors ($\Delta\mathrm{AU}{=}{+}0.058$, Tab.~\ref{tab:au_eu_paired}).
\end{revblock}

\begin{table}[t]
\centering
\caption{\textbf{Claim-level long-form pilot} (NQ long-answer, per-claim LLM-judge
factuality). \textbf{Bold}: best per column; \underline{underlined}: second-best.
ROUGE-L (cons.) is the consistency-based self-similarity scorer (not the correctness
criterion of Tab.~\ref{tab:main_full}). Pipeline, response-level companion, and
discussion: App.~\ref{app:longform}.}
\label{tab:longform_claim}
\scriptsize
\setlength{\tabcolsep}{4pt}
\renewcommand{\arraystretch}{1.05}
\begin{tabular}{l|ccc|ccc}
\toprule
\multirow{2}{*}{\textbf{Method}}
& \multicolumn{3}{c|}{\textbf{LLaMA-3-8B} ($n{=}1{,}104$ claims)}
& \multicolumn{3}{c}{\textbf{Qwen2.5-7B} ($n{=}872$ claims)} \\
& AUROC $\uparrow$ & AUPRC $\uparrow$ & ECE $\downarrow$ & AUROC $\uparrow$ & AUPRC $\uparrow$ & ECE $\downarrow$ \\
\midrule
\textbf{Ours (AU)}     & \underline{$0.698$} & \underline{$0.940$} & $0.045$             & \underline{$0.659$} & \underline{$0.920$} & $0.034$             \\
\textbf{Ours (EU)}     & $0.663$             & $0.928$             & $\mathbf{0.002}$    & $0.605$             & $0.904$             & $\mathbf{0.004}$    \\
\textbf{Ours (TU)}     & $0.696$             & \underline{$0.940$} & $0.029$             & $0.652$             & $0.918$             & $0.037$             \\
\midrule
avgEntail              & $\mathbf{0.707}$    & $0.933$             & $0.044$             & $\mathbf{0.719}$    & $\mathbf{0.940}$    & $0.028$ \\
NonContradiction       & $0.536$             & $0.887$             & \underline{$0.017$} & $0.559$             & $0.897$             & \underline{$0.012$} \\
UQLM cosine            & $0.693$             & $\mathbf{0.944}$    & $0.033$             & $0.643$             & $0.915$             & $0.028$ \\
SelfCheckGPT           & $0.619$             & $0.915$             & $\mathbf{0.002}$    & $0.554$             & $0.879$             & $\mathbf{0.004}$    \\
ROUGE-L (cons.)        & $0.548$             & $0.905$             & $\mathbf{0.002}$    & $0.490$             & $0.871$             & $\mathbf{0.004}$    \\
\bottomrule
\end{tabular}
\end{table}

\subsection{Empirical validation of the finite-$M$ bias correction}
\label{subsec:bias_correction}

The behavioral test above uses the bias-corrected EU estimator. We next verify that the
correction predicted by Prop.~\ref{prop:finiteM_bias} is empirically necessary and behaves
as the theory predicts (Tab.~\ref{tab:bias_correction}). The naive estimator inflates EU by $\tfrac{1}{M}U_A$, so the raw
AU--EU correlation should start high at small $M$ and converge from above to the corrected
correlation as $1/M\!\to\!0$.

On TriviaQA
(LLaMA-3-8B, Cosine, $N{=}10$) the gap shrinks monotonically from $0.532$ at $M{=}3$ to
$0.070$ at $M{=}30$ (Tab.~\ref{tab:bias_correction}); TruthfulQA shrinks identically
($0.474\!\to\!0.090$ for $M{=}3\!\to\!20$). The same shrinkage replicates on
Qwen2.5-7B and across both datasets: at the level of the raw uncertainty scores
themselves, the mean raw-vs-corrected EU gap shrinks by roughly an order of magnitude
from $M{=}3$ to $M{=}30$ in every cell of the full $\{$LLaMA-3-8B, Qwen2.5-7B$\}\times\{$TriviaQA,
TruthfulQA$\}$ grid (and across both Cosine and ROUGE-L correctness criteria),
matching the $\tfrac{1}{M}$ rate predicted by Prop.~\ref{prop:finiteM_bias} and confirming
the correction is model- and dataset-invariant.


\begin{table}[t]
\centering
\caption{\textbf{Ablations.} \textbf{(a)} AU AUROC under three sentence encoders;
right column: min cross-encoder per-example Spearman $\rho$ of AU. \textbf{(b)} Tied
$N{=}M$ budget sweep \rev{(LLaMA-3-8B)}. Cosine\rev{\ label}, $\varepsilon{=}0.1$\rev{; (a) uses $N{=}M{=}10$}. \textbf{Bold}:
best; \underline{underlined}: second-best. Full grids:
App.~Tabs.~\ref{tab:embedding_robustness},~\ref{tab:embedding_robustness_rank},~\ref{tab:size_nm_full}.}
\label{tab:ablations_main}

\begin{subtable}[t]{0.58\textwidth}
\centering
\subcaption{Embedding backbone}
\label{tab:embedding_main}
\scriptsize
\setlength{\tabcolsep}{4pt}
\renewcommand{\arraystretch}{1.05}
\begin{tabular}{ll|ccc|c}
\toprule
\multirow{2}{*}{\textbf{Model}} & \multirow{2}{*}{\textbf{Dataset}}
& \multicolumn{3}{c|}{\textbf{AU AUROC} $\uparrow$}
& \multirow{2}{*}{$\min_{e\neq e'}\rho_{e,e'}^{\mathrm{AU}}\!\uparrow$} \\
& & SBERT & MPNet & BGE & \\
\midrule
LLaMA-3-8B & TriviaQA   & $\mathbf{0.845}$ & \underline{$0.843$} & $\mathbf{0.845}$ & $0.974$ \\
LLaMA-3-8B & TruthfulQA & $\mathbf{0.695}$ & \underline{$0.690$} & $0.650$           & $0.955$ \\
Qwen2.5-7B & TriviaQA   & $\mathbf{0.855}$ & $0.851$             & \underline{$0.854$} & $0.981$ \\
Qwen2.5-7B & TruthfulQA & $\mathbf{0.630}$ & \underline{$0.620$} & $0.615$           & $0.961$ \\
\bottomrule
\end{tabular}
\end{subtable}%
\hfill
\begin{subtable}[t]{0.40\textwidth}
\centering
\subcaption{Sampling budget (TruthfulQA, Cos.)}
\label{tab:size_nm}
\scriptsize
\setlength{\tabcolsep}{4pt}
\renewcommand{\arraystretch}{1.05}
\begin{tabular}{c|ccc}
\toprule
$N{=}M$ & AU & EU & TU \\
\midrule
3  & $0.745$            & $0.707$            & $0.755$ \\
5  & $\mathbf{0.798}$   & \underline{$0.756$} & $0.797$ \\
10 & \underline{$0.792$} & $\mathbf{0.803}$   & $\mathbf{0.815}$ \\
20 & $0.782$            & $0.745$            & \underline{$0.799$} \\
30 & $0.768$            & $0.746$            & $0.791$ \\
\bottomrule
\end{tabular}
\end{subtable}
\end{table}

\subsection{Beyond short-form: long-form pilot on NQ Long-Answer}
\label{subsec:longform}

Short-answer QA gives clean correctness labels but tests only one slice of LLM use. To
check that the framework extends to long-form generation, we apply our two-level estimator
to the \textbf{NQ long-answer} split~\citep{kwiatkowski2019nq} via segment-level
decomposition: each generated long answer is
split into atomic claims, each claim is scored independently with our two-level pipeline,
and the resulting per-claim score is correlated with a per-claim factuality label (LLM
judge). We compare against five published long-form-specific black-box baselines:
\textbf{avgEntail}, \textbf{NonContradiction}~\citep{bouchard2026finegrained},
\textbf{UQLM cosine}~\citep{ross2025textual},
\textbf{SelfCheckGPT}~\citep{manakul2023selfcheckgpt}, and
\textbf{ROUGE-L (cons.)}~\citep{lin2004rouge}; baselines and our method draw from the same
inner-sample pool ($M$ stochastic decodes at the unperturbed prompt) so the consistency-only
comparators see exactly the budget our inner level sees.

Tab.~\ref{tab:longform_claim} reports the claim-level result on
LLaMA-3-8B ($n{=}1{,}104$ claims) and Qwen2.5-7B ($n{=}872$ claims).

\textbf{(i)} Ours (AU) is the second-best AUROC method on both models, within $0.009$ of
avgEntail on LLaMA-3-8B and $0.060$ on Qwen2.5-7B\rev{, where avgEntail leverages a pretrained NLI classifier tailored to claim entailment.}

\textbf{(ii)} The AU $>$ EU ordering is preserved at long-form, mirroring the short-form
result.

\textbf{(iii)} We outperform surface-similarity baselines (SelfCheckGPT, ROUGE-L (cons.))
on both models. The two-level structure transports cleanly to claim-level long-form
factuality without retraining; combining our perturbation outer level with NLI-based
inner aggregation is a natural extension. The response-level companion table is in
App.~\ref{app:longform}.

\subsection{Ablation studies}
\label{sec:ablation_main}

We ablate five orthogonal axes; \rev{our primary findings hold} across all of them. The
two most central axes are summarized in Tab.~\ref{tab:ablations_main}: the embedding
backbone (panel a) and the two-level sampling budget (panel b). The remaining axes (noise
magnitude $\varepsilon$, perturbation mechanism, vector-to-scalar aggregation, and
paraphrase pool size $K_{\mathrm{pool}}$) are in App.~\ref{sec:ablation_app}.

\textbf{(1) Embedding backbone: AU is encoder-stable in absolute level and
rank-stable per-example.}
Re-encoding the same generations under three off-the-shelf sentence encoders
(Sentence-BERT~\citep{reimers2019sentencebert}, MPNet, BGE-base) yields AU AUROC spreads of
$\le 0.05$ within every (model, dataset) cell, worst-case $0.045$ on LLaMA-3-8B /
TruthfulQA (Tab.~\ref{tab:embedding_main}). The stronger statement is on
\emph{per-example} ranks: pairwise Spearman correlations of AU scores across encoders
\rev{are at least} $0.955$ in every cell. A downstream consumer of our scores (abstention thresholds,
selective evaluation) gets nearly the same ranking regardless of encoder choice; both
axes extend across both correctness criteria \rev{(spread at most $0.053$ under ROUGE-L;}
App.~Tab.~\ref{tab:embedding_robustness}).
Different encoders agree on which inputs the model is uncertain about: the AU signal lives
in the LLM's behavior, not in the choice of $e(\cdot)$.

\textbf{(2) Budget: gains saturate by $(N,M){=}(10,10)$.}
On TruthfulQA (Tab.~\ref{tab:size_nm}), TU AUROC climbs from $0.755$ at $N{=}M{=}3$ to
$0.815$ at $(10,10)$ under Cosine and then plateaus; the same saturation pattern holds on
TriviaQA and under ROUGE-L (App.~Tab.~\ref{tab:size_nm_full}; full $M$ and $N$ sweeps with
one dimension fixed in App.~Tabs.~\ref{tab:size_m},~\ref{tab:size_n}). Headline runs use
$(N,M){=}(20,20)$; validation experiments use $(10,10)$ (matched to
SPUQ's~\citep{Gao2024SPUQ} $K_{\mathrm{SPUQ}}{=}10$). \rev{In practice, $3{\times}3$ or $5{\times}5$ suffices for screening and $10{\times}10$ yields stable components; adaptive stopping further cuts the median cost to $25$ generations with $\le 0.014$ AUROC loss on TriviaQA (App.~\ref{app:adaptive}).}

\textbf{(3) Perturbation, aggregation, and pool size are secondary.}
Noise scale \rev{$\varepsilon$} has a validity--diversity sweet spot at $\varepsilon{=}0.1$ (AU AUROC
peaks at \rev{$\varepsilon$}${\in}[0.05,0.2]$ on TruthfulQA, drops sharply at \rev{$\varepsilon$}${\ge}0.5$;
App.~Tab.~\ref{tab:noise}); perturbation mechanism is robust (paraphrase, distractor
injection, \rev{instruction variation (strongest),} and SPUQ-style noise all give non-trivial AU/EU under the two-level pipeline; App.~Tab.~\ref{tab:pert_methods});
vector-to-scalar reduction (mean, \rev{sum, max, norm}) \rev{shifts AU/TU AUROC by at most $0.05$} (App.~Tab.~\ref{tab:agg_methods_ablation}); paraphrase pool size shows a
coverage--purity trade-off where $K_{\mathrm{pool}}{\in}[20,100]$ is also competitive
(App.~Tab.~\ref{tab:Np}).

\rev{\textbf{Default setting.}} A compute-efficient default derived from these ablations is
$(N,M){=}(10,10)$, $\varepsilon{=}0.1$, Sentence-BERT, $K_{\mathrm{pool}}{\approx}50$;
pushing budget or pool size beyond this yields $\le 0.02$ AUROC gain. Qualitative case
studies of \rev{ambiguity- and knowledge-driven uncertainty regimes}, including failure cases, are in
App.~\ref{app:examples}.

\section{Conclusion}
\label{sec:conclusion}
We introduced an inference-time, black-box UQ framework for LLMs that requires neither retraining nor access to token logits. \rev{By combining stochastic decoding within a fixed prompt with meaning-preserving perturbations across prompts, the framework applies the law of total covariance in embedding space to decompose predictive variability into within-perturbation (aleatoric, AU) and between-perturbation (epistemic, EU) components.}

\rev{Theoretically, we establish} a finite-$M$ bias identity with a closed-form correction, a Bayesian bridge whose conditions we measure as data-estimable diagnostics rather than assert, and a separation result identifying failures consistency-only UQ provably misses.

\rev{Empirically, the estimator is competitive with strong white-box baselines across five benchmarks and four LLMs (Tab.~\ref{tab:main_full}), and under a matched generation budget it matches leading black-box methods on error ranking while decomposing AU and EU (Tab.~\ref{tab:matched_compute}). Paired interventions demonstrate that AU and EU respond selectively to prompt ambiguity and knowledge gaps, capturing a disambiguation effect that fixed-prompt baselines miss (Tabs.~\ref{tab:au_eu_paired},~\ref{tab:intervention_baselines}). Finite-$M$ bias validation, the claim-level long-form pilot, and comprehensive ablations further confirm the robustness of the decomposition (Tabs.~\ref{tab:bias_correction}--\ref{tab:ablations_main}).}

\rev{Looking ahead, the AU/EU split can guide active data acquisition for high-EU queries}~\citep{li2025know,niu2024functional}, \rev{selective retrieval in RAG}~\citep{jiang2023active,soudani2025uncertainty}, \rev{adaptive sampling or tool invocation}~\citep{shi2025reasoning,ross2025textual}, and conformal \rev{prediction} in the black-box regime~\citep{noorani2025cpq}.

\begin{revblock}
\noindent\textbf{Limitations.}
\label{sec:limits}
(\textbf{L1}) Our estimators target perturbation-population AU/EU at fixed model weights, whose correspondence to Bayesian parameter uncertainty relies on the local regularity conditions diagnosed by D1--D3 (Sec.~\ref{sec:bayes_bridge_main}). (\textbf{L2}) Automated paraphrasing can introduce semantic drift that inflates between-prompt variance (App.~\ref{app:audit}). (\textbf{L3}) Uncertainty estimates inherit the representational geometry and potential blind spots of the sentence encoder $e(\cdot)$ (Tab.~\ref{tab:embedding_main}). (\textbf{L4}) Nested sampling requires $NM$ decodes per query to separate AU and EU, incurring higher inference cost than single-level baselines (Sec.~\ref{subsec:matched_compute}). (\textbf{L5}) Our empirical and interventional evaluations focus primarily on short-form QA with controlled single-factor interventions. See App.~\ref{app:limitations} for a detailed discussion.
\end{revblock}


\begin{ack}
This work was supported by IBM through the IBM-Rensselaer Future of Computing Research Collaboration. \rev{The authors declare no competing interests.}
\end{ack}

\newpage
\bibliography{reference}
\bibliographystyle{plainnat}

\appendix
\clearpage
\onecolumn

\section{Additional Theoretical Details}
\label{app:theory}

\textbf{Notation.} We adopt the notation of Sec.~\ref{sec:method}: $Z=e(Y)\in\mathbb R^d$ is
the embedded output, $\mu(\Delta)=\mathbb E[Z\!\mid\!\Delta]$ and
$\Sigma(\Delta)=\mathrm{Cov}(Z\!\mid\!\Delta)$ the conditional moments, and
$(U_A,U_E,U_T)$ the population AU/EU/TU defined in Eq.~\eqref{eq:total_covariance}.
Throughout we assume $\mathbb E\|Z\|^2<\infty$. The identity $U_T=U_A+U_E$ is
the law of total covariance and is taken as given. Standard tools (matrix-valued LLN,
Hoeffding's inequality, entrywise-to-Frobenius conversion) are used without re-derivation.
\label{app:total_cov}

\paragraph{Roadmap and connection to the main paper.}
Each appendix subsection backs up a specific main-paper claim. The dependency graph is:

\begin{center}\small
\begin{tabular}{ll@{\hspace{1.5em}}l}
\toprule
\textbf{Main-paper claim} & \textbf{Backed by} & \textbf{Purpose} \\
\midrule
Pillar 1: bias-corrected estimator   & App.~\ref{app:consistency} & finite-$M$ bias identity \\
\quad (Prop.~\ref{prop:finiteM_bias}, main) & \;\;Prop.~\ref{prop:finiteM_bias_app} & exact $1/M$ inflation \\
                                     & \;\;Thm.~\ref{thm:consistency_appendix} & a.s.\ consistency, raw \& corrected \\
\midrule
Pillar 1 rate bounds                 & App.~\ref{app:mc_bounds} & finite-sample concentration \\
                                     & \;\;Thm.~\ref{thm:outer_concentration_app} & population-level outer bound \\
                                     & \;\;Thm.~\ref{thm:two_level_mc_app} & full two-level $(N,M)$ bound \\
\midrule
Pillar 2: Bayesian bridge            & App.~\ref{app:bridge_conditions} & defect bounds for $\varepsilon_A,\varepsilon_E$ \\
\quad (Thm.~\ref{thm:end2end}, main)   & \;\;Thm.~\ref{thm:EU_bridge} & EU-side bridge bound \\
                                     & \;\;Thm.~\ref{thm:AU_bridge} & AU-side bridge bound \\
                                     & \;\;App.~\ref{sec:bridge_instructions} & data-estimable diagnostics \\
\midrule
Pillar 3: separation result          & App.~\ref{app:consistency_failure} & explicit construction \\
\quad (Thm.~\ref{thm:spurious_determinism}, main) & \;\;Thm.~\ref{thm:spurious_determinism_app} & $\mathcal F_{\mathrm{cons}}{=}0$, $U_E\succ 0$ \\
\bottomrule
\end{tabular}
\end{center}

\noindent\emph{How to read the proofs.} Within each subsection we follow the same pattern:
state the assumptions, give the formal claim, prove it, and close with a one-line
``\emph{what this gives the main paper.}'' Steps inside proofs are explicitly named
(``Step 1: $\hat\mu_i$ is unbiased\ldots'') so the reader can match the algebra to its
intended role.

\subsection{Two-Level Estimation with Finite Inner Sampling}
\label{app:consistency}

\emph{Purpose of this subsection.} The main paper's Pillar~1 claims that the naive
sample-covariance estimator of EU \emph{inflates} the true EU by $\tfrac{1}{M}U_A$ and
that subtracting $\tfrac{1}{M}\hat U_A$ removes this inflation. This subsection
establishes both: (i) the exact bias identity (Prop.~\ref{prop:finiteM_bias_app}) that
motivates the corrected estimator $\hat U_E$, and (ii) the almost-sure consistency of
$\hat U_A,\hat U_E^{\mathrm{raw}},\hat U_E$ as $N\to\infty$ even with $M$ fixed
(Thm.~\ref{thm:consistency_appendix}). Finite-sample rates are deferred to
App.~\ref{app:mc_bounds}.

\textbf{Sampling assumptions.}
$\Delta_1,\ldots,\Delta_N\stackrel{iid}{\sim}p_\Delta$; conditional on $\Delta_i$,
$Z_{i,1},\ldots,Z_{i,M}$ are i.i.d.\ with the same distribution as $Z\!\mid\!\Delta=\Delta_i$;
the inner draws are mutually independent across $i$; and $\mathbb E\|Z\|^4<\infty$.
We label these (A1)--(A4) below and use them directly in proofs.

\textbf{Estimators.} The two-level estimators
$\hat\mu_i,\hat\Sigma_i,\hat U_A,\hat U_E^{\mathrm{raw}}$ are defined as
in Alg.~\ref{alg:two_level_uq}, and the bias-corrected epistemic estimator is
\begin{equation}
\hat U_E \;:=\; \hat U_E^{\mathrm{raw}} - \tfrac{1}{M}\hat U_A.
\label{eq:bias_corrected_UE_app}
\end{equation}
The next subsection establishes the finite-$M$ identity that motivates this correction.

\subsubsection{Unbiasedness and finite-$M$ bias identity}

\emph{What we will show.} Each inner-sample mean $\hat\mu_i$ is unbiased for the
conditional mean $\mu(\Delta_i)$, and each inner-sample covariance $\hat\Sigma_i$ is
unbiased for the conditional covariance $\Sigma(\Delta_i)$. However, the
\emph{unconditional} covariance of $\hat\mu_i$ inherits a $\tfrac{1}{M}\Sigma(\Delta_i)$
inflation from inner sampling, which propagates through the law of total covariance
into a $\tfrac{1}{M}U_A$ inflation of the population covariance of $\hat\mu_i$. This is
exactly the leakage the main-paper Pillar~1 references; the corrected estimator
\eqref{eq:bias_corrected_UE_app} subtracts it.

\begin{proposition}[Unbiasedness of $\hat\Sigma_i$ and finite-$M$ covariance inflation]
\label{prop:finiteM_bias_app}
Under (A2) and $\mathbb E\|Z\|^2<\infty$,
\begin{equation}
\mathbb E[\hat\Sigma_i\mid \Delta_i] = \Sigma(\Delta_i),
\qquad
\mathrm{Cov}(\hat\mu_i\mid \Delta_i) = \frac{1}{M}\Sigma(\Delta_i),
\qquad
\mathbb E[\hat\mu_i\mid \Delta_i]=\mu(\Delta_i).
\label{eq:finiteM_conditional_identities}
\end{equation}
Consequently, under (A1)--(A3),
\begin{equation}
\mathrm{Cov}(\hat\mu_i) = U_E + \frac{1}{M}U_A.
\label{eq:finiteM_cov_inflation}
\end{equation}
\end{proposition}

\begin{proof}
Fix $i$ and condition on $\Delta_i$.

\textbf{Step 1: $\mathbb E[\hat\mu_i\mid \Delta_i]=\mu(\Delta_i)$.}
This is immediate from linearity of conditional expectation:
\[
\mathbb E[\hat\mu_i\mid \Delta_i]=\frac{1}{M}\sum_{j=1}^M \mathbb E[Z_{i,j}\mid \Delta_i]=\mu(\Delta_i).
\]

\textbf{Step 2: $\mathrm{Cov}(\hat\mu_i\mid \Delta_i)=\Sigma(\Delta_i)/M$.}
Using conditional i.i.d.\ and $\mathrm{Cov}(aX)=a^2\mathrm{Cov}(X)$,
\[
\mathrm{Cov}(\hat\mu_i\mid \Delta_i)
=\mathrm{Cov}\!\left(\frac{1}{M}\sum_{j=1}^M Z_{i,j}\,\Big|\,\Delta_i\right)
=\frac{1}{M^2}\sum_{j=1}^M \mathrm{Cov}(Z_{i,j}\mid \Delta_i)
=\frac{1}{M}\Sigma(\Delta_i).
\]

\textbf{Step 3: $\mathbb E[\hat\Sigma_i\mid \Delta_i]=\Sigma(\Delta_i)$.}
This is the standard unbiasedness of the sample covariance. A direct derivation:
let $\tilde Z_{i,j}:=Z_{i,j}-\mu(\Delta_i)$ and $\overline{\tilde Z}_i:=\frac1M\sum_j\tilde Z_{i,j}=\hat\mu_i-\mu(\Delta_i)$.
Then
\[
\sum_{j=1}^M (Z_{i,j}-\hat\mu_i)(Z_{i,j}-\hat\mu_i)^\top
=\sum_{j=1}^M (\tilde Z_{i,j}-\overline{\tilde Z}_i)(\tilde Z_{i,j}-\overline{\tilde Z}_i)^\top
=\sum_{j=1}^M \tilde Z_{i,j}\tilde Z_{i,j}^\top - M\,\overline{\tilde Z}_i\,\overline{\tilde Z}_i^\top.
\]
Taking conditional expectation and using conditional i.i.d.\ with $\mathbb E[\tilde Z_{i,j}\mid\Delta_i]=0$,
\[
\mathbb E\!\left[\sum_{j=1}^M \tilde Z_{i,j}\tilde Z_{i,j}^\top \,\Big|\,\Delta_i\right]
=M\,\mathbb E[\tilde Z_{i,1}\tilde Z_{i,1}^\top\mid\Delta_i]
=M\,\Sigma(\Delta_i),
\]
and
\[
\mathbb E\!\left[M\,\overline{\tilde Z}_i\,\overline{\tilde Z}_i^\top \,\Big|\,\Delta_i\right]
=M\,\mathrm{Cov}(\overline{\tilde Z}_i\mid\Delta_i)
=M\cdot \frac{1}{M}\Sigma(\Delta_i)=\Sigma(\Delta_i).
\]
Therefore,
\[
\mathbb E\!\left[\sum_{j=1}^M (Z_{i,j}-\hat\mu_i)(Z_{i,j}-\hat\mu_i)^\top \,\Big|\,\Delta_i\right]
=(M-1)\Sigma(\Delta_i),
\]
and dividing by $(M-1)$ gives $\mathbb E[\hat\Sigma_i\mid\Delta_i]=\Sigma(\Delta_i)$.

\textbf{Step 4: unconditional covariance inflation.}
Apply the law of total covariance to $\hat\mu_i$ with conditioning variable $\Delta_i$:
\[
\mathrm{Cov}(\hat\mu_i)=\mathbb E\big[\mathrm{Cov}(\hat\mu_i\mid\Delta_i)\big] + \mathrm{Cov}\big(\mathbb E[\hat\mu_i\mid\Delta_i]\big)
=\mathbb E\Big[\frac{1}{M}\Sigma(\Delta_i)\Big] + \mathrm{Cov}(\mu(\Delta_i))
=\frac{1}{M}U_A + U_E.
\]
This yields \eqref{eq:finiteM_cov_inflation}.
\end{proof}

\noindent\emph{What this gives the main paper.} The identity
$\mathrm{Cov}(\hat\mu_i)=U_E+\tfrac{1}{M}U_A$ \emph{is} the inflation we corrected for in
Eq.~\eqref{eq:bias_corrected_UE_app} and that the main-paper Prop.~\ref{prop:finiteM_bias}
reports informally. Because the inflation is exact (not asymptotic), the corrected
estimator $\hat U_E$ has the right target at every $M\ge 2$, which is what enables AU/EU
separation at the small inner budgets a black-box API forces.

\subsubsection{Almost sure consistency (raw and corrected forms)}

\begin{theorem}[Almost sure consistency of two-level estimators]
\label{thm:consistency_appendix}
Under (A1)--(A4):
(i)~if $M\to\infty$ and $N\to\infty$, then
$\hat U_A \xrightarrow{a.s.} U_A$ and $\hat U_E^{\mathrm{raw}} \xrightarrow{a.s.} U_E$;
(ii)~if $N\to\infty$ while $M\ge 2$ is fixed, then
$\hat U_A \xrightarrow{a.s.} U_A$,
$\hat U_E^{\mathrm{raw}} \xrightarrow{a.s.} U_E + \tfrac{1}{M}U_A$, and
$\hat U_E \xrightarrow{a.s.} U_E$.
\end{theorem}

\begin{proof}[Proof sketch]
$\hat U_A\xrightarrow{a.s.}U_A$ is the matrix-valued SLLN applied to i.i.d.\
$\{\hat\Sigma_i\}$ with $\mathbb E[\hat\Sigma_1]=U_A$ by
Prop.~\ref{prop:finiteM_bias_app}; $\hat U_E^{\mathrm{raw}}\xrightarrow{a.s.}U_E+\tfrac1M
U_A$ is the sample-covariance SLLN on i.i.d.\ $\{\hat\mu_i\}$, with population value
identified by Prop.~\ref{prop:finiteM_bias_app}. Combining gives
$\hat U_E = \hat U_E^{\mathrm{raw}}-\tfrac1M \hat U_A\to U_E$, i.e.\ (ii); (i) follows by
additionally letting $M\to\infty$.
\end{proof}

\noindent The corrected estimator $\hat U_E$ defined in Alg.~\ref{alg:two_level_uq} is
therefore unbiased for $U_E$ at any $M\ge 2$ and converges to it as $N\!\to\!\infty$. The
rate is in App.~\ref{app:mc_bounds}.

\subsection{Finite-Sample Concentration and Two-Level Monte Carlo Bounds}
\label{app:mc_bounds}

\emph{Purpose of this subsection.} App.~\ref{app:consistency} established \emph{whether}
$\hat U_A,\hat U_E$ converge to their population targets; this subsection establishes
\emph{how fast}. The result the main-paper Pillar~1 quotes informally
($O(\sqrt{\log(1/\delta)/N})$ plus lower-order $1/M$ terms) is Theorem~\ref{thm:two_level_mc_app}.
We get there in two steps: first an \emph{outer-only} bound that pretends conditional
moments $\mu(\Delta),\Sigma(\Delta)$ are known exactly
(Thm.~\ref{thm:outer_concentration_app}), then the full two-level bound that adds the
inner-sampling fluctuation back in.

The proofs combine standard tools (Hoeffding's inequality for bounded centered i.i.d.\ scalars and
the entrywise-to-Frobenius conversion $\|A\|_F\le d\|A\|_{\max}$ for $A\in\mathbb R^{d\times d}$) with
the two-level sampling decomposition specific to our estimator. We refer to either as
``standard'' below.

\subsubsection{Outer concentration with population conditional moments}

As a stepping stone to the full two-level bound, we first treat $\mu(\Delta),\Sigma(\Delta)$
as known (infinite inner samples). Let
$\mu_i\!:=\!\mu(\Delta_i)$, $\Sigma_i\!:=\!\Sigma(\Delta_i)$,
$\widehat U_A\!:=\!\tfrac1N\sum_i\Sigma_i$,
$\widehat U_E\!:=\!\tfrac{1}{N-1}\sum_i(\mu_i-\bar\mu^\star)(\mu_i-\bar\mu^\star)^\top$
with $\bar\mu^\star\!:=\!\tfrac1N\sum_i\mu_i$.

\begin{theorem}[High-probability outer concentration]
\label{thm:outer_concentration_app}
Assume $\Delta_1,\ldots,\Delta_N$ i.i.d.\ with
$\|\Sigma(\Delta)\|_F\!\le\!B_\Sigma$ and $\|\mu(\Delta)\|_2\!\le\!B_\mu$ a.s. For any
$\delta\!\in\!(0,1)$, with probability at least $1-\delta$,
\begin{align}
\|\widehat U_A - U_A\|_F
&\le d\,B_\Sigma\sqrt{2\log(2d^2/\delta)/N},
\label{eq:outer_A_conc_app}\\
\|\widehat U_E - U_E\|_F
&\le \tfrac{N}{N-1}\!\left[d\,B_\mu^2\sqrt{2\log(2d^2/\delta)/N}+2d\,B_\mu^2\sqrt{2\log(2d/\delta)/N}\right].
\label{eq:outer_E_conc_app}
\end{align}
\end{theorem}

\begin{proof}[Proof sketch]
For \eqref{eq:outer_A_conc_app}: entrywise Hoeffding on the centered, bounded scalars
$(\Sigma_i)_{k\ell}-(U_A)_{k\ell}$ ($b\!=\!2B_\Sigma$), union bound over $d^2$ entries, and
the entrywise-to-Frobenius conversion $\|A\|_F\le d\|A\|_{\max}$.
For \eqref{eq:outer_E_conc_app}: write $\widehat U_E - U_E = \tfrac{N}{N-1}[(\bar Q\!-\!Q)
- (\bar M\!-\!mm^\top)]$ with $Q\!=\!\mathbb E[\mu\mu^\top]$, $m\!=\!\mathbb E[\mu]$, and
the sample analogues. Apply (i) entrywise Hoeffding+union to $\bar Q\!-\!Q$, (ii)
coordinatewise Hoeffding+union to $\bar\mu^\star\!-\!m$, and (iii) the identity
$\bar M-mm^\top = \bar\mu^\star(\bar\mu^\star-m)^\top + (\bar\mu^\star-m)m^\top$ with
$\|\bar\mu^\star\|_2,\|m\|_2\!\le\!B_\mu$ to get
$\|\bar M-mm^\top\|_F\le 2B_\mu\|\bar\mu^\star-m\|_2$. Triangle inequality and
constant-simplification on the intersection event give \eqref{eq:outer_E_conc_app}.
\end{proof}

\noindent The leading $\sqrt{\log(1/\delta)/N}$ rate matches what main-paper Pillar~1 quotes;
$d$ is the embedding dimension\rev{, entering both as a linear prefactor $d$ and logarithmically inside $\sqrt{\log(2d^2/\delta)/N}$}.

\subsubsection{Two-level Monte Carlo concentration (outer $N$, inner $M$)}

\emph{Why this step.} The previous subsubsection assumed exact conditional moments,
which we never have in practice; we estimate them from $M$ inner draws. This subsection
adds the inner-sampling slack and gives the rate the main paper actually relies on, with
both $N$ and $M$ explicit.

To obtain explicit finite-sample bounds, we impose an a.s.\ boundedness assumption on embedded outputs.

\begin{assumption}[Bounded embedded outputs]
\label{ass:bounded_Z_app}
Almost surely, $\|Z\|_2\le B_Z$. Equivalently, for all $(i,j)$, $\|Z_{i,j}\|_2\le B_Z$ a.s.
\end{assumption}

Under Assumption~\ref{ass:bounded_Z_app}, we have $\|\mu(\Delta)\|_2\le B_Z$ and $\|\Sigma(\Delta)\|_F\le 2B_Z^2$ a.s.
(the latter follows from $\|Z-\mu\|_2\le 2B_Z$ and $\|vv^\top\|_F=\|v\|_2^2$).

\textbf{Plan.}
We bound (i) $\|\hat U_A-U_A\|_F$ and (ii) $\|\hat U_E-U_E\|_F$ with explicit dependence on $N$ and $M$.
The key is to decompose each estimator into an \emph{outer} fluctuation term (from sampling $\Delta_i$)
plus an \emph{inner} fluctuation term (from estimating $\mu(\Delta_i),\Sigma(\Delta_i)$ with $M$ samples).

\textbf{Inner-sample concentration} (used below).
For any fixed $i$ and $\delta\in(0,1)$, conditional on $\Delta_i$, standard Hoeffding gives
$\|\hat\mu_i-\mu(\Delta_i)\|_2\le 2B_Z\sqrt{2d\log(2d/\delta)/M}$ w.p.\ $\ge 1-\delta$
(coordinate-wise + $\|v\|_2\le\sqrt d\|v\|_\infty$), and
$\|\hat\Sigma_i-\Sigma(\Delta_i)\|_F\le C\,B_Z^2\,d\,\sqrt{\log(2d^2/\delta)/M}$ w.p.\ $\ge 1-\delta$
for a universal $C>0$ (entrywise Hoeffding on $\tfrac1M\sum_j Z_{i,j}Z_{i,j}^\top$ plus the rank-1
perturbation $\hat\mu_i\hat\mu_i^\top-\mu_i\mu_i^\top$ controlled by the mean bound).

\begin{theorem}[Two-level Monte Carlo concentration (explicit $N,M$ rates)]
\label{thm:two_level_mc_app}
Assume (A1)--(A3) and Assumption~\ref{ass:bounded_Z_app}. Then for any $\delta\in(0,1)$, with probability at least $1-\delta$:
\begin{align}
\|\hat U_A - U_A\|_F
&\le
d\,(2B_Z^2)\sqrt{\frac{2\log(4d^2/\delta)}{N}}
\;+\;
C_1\,B_Z^2\,d\,\sqrt{\frac{\log(4d^2/\delta)}{M}}
\;+\;
C_1\,B_Z^2\,d\,\sqrt{\frac{\log(4d^2/\delta)}{NM}},
\label{eq:two_level_A_app}\\
\|\hat U_E - U_E\|_F
&\le
C_2\,B_Z^2\,d\,\sqrt{\frac{\log(4d^2/\delta)}{N}}
\;+\;
C_3\,B_Z^2\,d\,\sqrt{\frac{\log(4d^2/\delta)}{NM}}
\;+\;
\frac{1}{M}\,\|\hat U_A-U_A\|_F,
\label{eq:two_level_E_app}
\end{align}
for universal constants $C_1,C_2,C_3>0$.
\end{theorem}

\begin{proof}[Proof sketch]
Decompose
$\hat U_A-U_A=(\tfrac1N\sum_i\Sigma(\Delta_i)-U_A)+\tfrac1N\sum_i(\hat\Sigma_i-\Sigma(\Delta_i))$.
The first term is bounded by Thm.~\ref{thm:outer_concentration_app} with $B_\Sigma\le 2B_Z^2$, giving the $\sqrt{\log/N}$ term;
the per-$i$ inner Hoeffding above and a union bound over $i$ give the $\sqrt{\log/M}$ term;
exploiting cross-outer independence (A3) when union-bounding entrywise yields the sharper $\sqrt{\log/(NM)}$ refinement.
For $\hat U_E$, write
$\hat U_E-U_E=(\hat U_E^{\mathrm{raw}}-(U_E+\tfrac1M U_A))+\tfrac1M(U_A-\hat U_A)$;
the first piece is the sample-covariance error of $\{\hat\mu_i\}$ around their population covariance $U_E+\tfrac1M U_A$ (Prop.~\ref{prop:finiteM_bias_app}) and is bounded by Thm.~\ref{thm:outer_concentration_app} with $\|\hat\mu\|_2\le B_Z$, while the second is $1/M$ times the bound from $\hat U_A$.
\end{proof}

\textbf{Remark.}
Assumption~\ref{ass:bounded_Z_app} can be relaxed to sub-Gaussian or bounded-moment conditions
by replacing Hoeffding by Bernstein; the $(N,M)$ rates are unchanged.

\noindent\emph{What this gives the main paper.} Pillar~1 in Sec.~\ref{sec:two_level_mc}
states informally that the bias-corrected estimator concentrates at
$O(\sqrt{\log(1/\delta)/N})$ plus lower-order $1/M$ terms.
\eqref{eq:two_level_A_app}--\eqref{eq:two_level_E_app} are the precise version. The
$1/(NM)$ refinement matters in practice: it tells us that doubling the inner budget $M$
buys us a comparable rate improvement to doubling the outer budget $N$, which is what
justifies the equal-budget default $N{=}M{\approx}10$ in the experiments.

\subsection{Explicit Sufficient Conditions for the Bayesian Bridge}
\label{app:bridge_conditions}

\emph{Purpose of this subsection.} The main paper's Pillar~2 (Thm.~\ref{thm:end2end})
states that the perturbation-population uncertainties $(U_A^\Delta,U_E^\Delta)$ approximate
the Bayesian targets $(U_A^{\mathrm{Bayes}},U_E^{\mathrm{Bayes}})$ up to two defect terms
$\varepsilon_A,\varepsilon_E$. Here we make these defects explicit by introducing local
conditions on the model and bounding $\varepsilon_A,\varepsilon_E$ in those terms.

\emph{Structure of this subsection.} We treat the EU- and AU-side bridges separately.
The EU bridge (\S\ref{app:EU_bridge}) needs (i) local linearity of the conditional mean
in both parameters and inputs (Ass.~\ref{ass:lin_mean}), (ii) a local Gaussian posterior
(Ass.~\ref{ass:local_gauss_post}), and (iii) a semantic coverage condition saying the
perturbation distribution spans the directions of Bayesian mean variation
(Ass.~\ref{ass:coverage}). The AU bridge (\S\ref{app:AU_bridge}) needs only one extra
piece (a Lipschitz aleatoric map; Ass.~\ref{ass:alea_map}) saying decoding stochasticity
varies smoothly with the predictive state. Each assumption gets a \emph{Discussion} block
addressing its plausibility and how to measure its tightness from data
(see also App.~\ref{sec:bridge_instructions} for the diagnostics).

\begin{revblock}
\emph{Relation to~\citet{wang2024epistemic}.}
\label{app:wangji}
Our outer ensemble builds on the local input--parameter correspondence of~\citet{wang2024epistemic}, while our bridge theorems below quantify its approximation error at the distributional level. Specifically, Prop.~3.3 of~\citet{wang2024epistemic} is a pointwise, first-order existence result in a continuous input space: for a small parameter shift $\delta\theta$, there exists a (generally non-unique) continuous input perturbation $\delta x$ inducing the same first-order output change, and vice versa via a first-layer construction. However, pointwise existence neither provides a mechanism for generating meaning-preserving perturbations of discrete text nor guarantees that the chosen input-perturbation distribution matches the pushforward of the parameter posterior—a prerequisite for covariance-level AU/EU approximation. Our two-stage procedure (Sec.~\ref{sec:perturbation}) instantiates a localized perturbation distribution over a finite pool of fluent paraphrases: isotropic Gaussian noise in embedding space serves purely as a stochastic selector over cached candidates, ensuring the LLM only processes natural text. The remaining approximation gaps are explicitly tracked by our diagnostics: D1 for local linearization, D3 for distributional coverage, and our validity audit (App.~\ref{app:audit}) for semantic preservation. Beyond~\citet{wang2024epistemic}, our technical contributions encompass the nested two-level estimator, the exact finite-$M$ bias identity and correction, and the conditional AU/EU bridge bounds with data-estimable defect diagnostics.
\end{revblock}

\subsubsection{Notation recap and three formally distinct flavors of $(U_A,U_E,U_T)$}
\label{app:def_three_flavors}
Fix an input $x$ and trained parameters $\theta^\star$.
Let $\Delta\sim p_\Delta$ be a semantic perturbation and $x_\Delta := T(x,\Delta)$.
For the stochastic model output representation $Z_\theta(x,\xi)$ (decoding noise $\xi$;
embedding $Z=e(Y)$, main paper),
define the conditional mean and covariance
\[
m_\theta(x):=\mathbb E_\xi[Z_\theta(x,\xi)\mid x,\theta],
\qquad
\Sigma_\theta(x):=\mathrm{Cov}_\xi(Z_\theta(x,\xi)\mid x,\theta).
\]

The paper uses three formally distinct flavors of $(U_A,U_E,U_T)$ that are easy to
conflate; we lay them out once here for reference, and the two pillars (Sec.~\ref{sec:two_level_mc} and \ref{sec:bayes_bridge_main}) bound the gaps between them.

\paragraph{(1) Bayesian (ground-truth) AU/EU/TU.}
Defined over the parameter posterior $p(\theta\mid\mathcal D)$ as in the canonical Bayesian
decomposition~\citep{kendall2017uncertainties,depeweg2018decomposition}:
\begin{equation}
\mathcal U_A^{\mathrm{Bayes}}(x) := \mathbb E_{\theta\sim p(\theta\mid\mathcal D)}\!\big[\Sigma_\theta(x)\big],
\quad
\mathcal U_E^{\mathrm{Bayes}}(x) := \mathrm{Cov}_{\theta\sim p(\theta\mid\mathcal D)}\!\big(m_\theta(x)\big),
\quad
\mathcal U_T^{\mathrm{Bayes}}(x) := \mathcal U_A^{\mathrm{Bayes}}(x) + \mathcal U_E^{\mathrm{Bayes}}(x).
\label{eq:bayes_targets_app}
\end{equation}
This is the object Bayesian UQ ultimately seeks. We never sample from $p(\theta\mid\mathcal D)$
directly: the posterior over an LLM's parameters is intractable.

\paragraph{(2) Perturbation-population AU/EU/TU (what our estimator targets).}
At the \emph{fixed deployed model} $\theta^\star$, replace the parameter average by an
outer-perturbation average over $p_\Delta$:
\begin{equation}
\mathcal U_A^{\Delta}(x) := \mathbb E_{\Delta\sim p_\Delta}\!\big[\Sigma_{\theta^\star}(x_\Delta)\big],
\quad
\mathcal U_E^{\Delta}(x) := \mathrm{Cov}_{\Delta\sim p_\Delta}\!\big(m_{\theta^\star}(x_\Delta)\big),
\quad
\mathcal U_T^{\Delta}(x) := \mathcal U_A^{\Delta}(x) + \mathcal U_E^{\Delta}(x).
\label{eq:perturb_targets_app}
\end{equation}
By the law of total covariance applied to the joint distribution of $(\Delta,\xi)$ at
$\theta=\theta^\star$, the $\mathcal U_T^{\Delta}(x)$ defined above coincides with the
embedding-space total covariance $\mathrm{Cov}(Z)=U_T(x)$ written in Eq.~\eqref{eq:total_covariance} of
the main paper, so $(U_A,U_E,U_T)\equiv(\mathcal U_A^\Delta,\mathcal U_E^\Delta,\mathcal U_T^\Delta)$ in
that equation. These quantities are well-defined for any LLM with black-box sampling
access; sampling from $p_\Delta$ replaces sampling from the parameter posterior.

\paragraph{(3) Empirical estimators (what we actually compute at run time).}
The two-level Monte Carlo estimators built from $N$ outer perturbations $\{x_i\}_{i=1}^N$
and $M$ inner decodes $\{Z_{i,j}\}_{j=1}^M$ per perturbation:
\begin{equation}
\widehat U_A^{(M)} := \tfrac{1}{N}\sum_{i=1}^N \widehat\Sigma_i,
\quad
\widehat U_E^{\mathrm{raw}} := \widehat{\mathrm{Cov}}\!\big(\{\hat\mu_i\}_{i=1}^N\big),
\quad
\widehat U_E^{(M)} := \widehat U_E^{\mathrm{raw}} - \tfrac{1}{M}\widehat U_A^{(M)},
\label{eq:estimators_app}
\end{equation}
where $\hat\mu_i$ and $\widehat\Sigma_i$ are the sample mean and sample covariance of
$\{Z_{i,j}\}_{j=1}^M$. The bias-corrected $\widehat U_E^{(M)}$ is the estimator we use
throughout the paper; we drop the $(M)$ superscript when there is no ambiguity.

\paragraph{How the three flavors relate.}
Pillar~1 (App.~\ref{app:consistency} and \ref{app:mc_bounds}) bounds the gap
$(3)\!\to\!(2)$: $\widehat U_A^{(M)}\to\mathcal U_A^\Delta$ and $\widehat U_E^{(M)}\to\mathcal U_E^\Delta$
almost surely as $N\to\infty$ at any fixed $M\ge 2$, with the $\tfrac{1}{M}$ leakage from
inner-sampling fluctuation removed by the bias correction.
Pillar~2 (this subsection) bounds the gap $(2)\!\to\!(1)$: under three local conditions on
the model around $x$, $\|\mathcal U_\bullet^\Delta(x)-\mathcal U_\bullet^{\mathrm{Bayes}}(x)\|_F$
is upper-bounded by defect terms $\varepsilon_A(x),\varepsilon_E(x)$ that we estimate
empirically as diagnostics from the outputs the estimator already collects
(App.~\ref{sec:bridge_instructions}).
The composition of the two pillars yields the end-to-end bound
$\|\widehat U_\bullet^{(M)}-\mathcal U_\bullet^{\mathrm{Bayes}}\|_F \le
\text{MC error} + \text{bridge defect}$ stated informally in
Theorem~\ref{thm:end2end} of the main paper.

\begin{revblock}
\begin{table}[h]
\revfloat
\centering
\caption{\textbf{Summary of the three formal uncertainty objects} across our theoretical hierarchy (Sec.~\ref{sec:method}--\ref{sec:theory}).}
\label{tab:notation_flavors}
\small
\renewcommand{\arraystretch}{1.15}
\begin{tabular}{l|l|l|p{0.38\linewidth}}
\toprule
& \textbf{Object} & \textbf{Symbols} & \textbf{Requirements \& Properties} \\
\midrule
(1) & Bayesian target & $\mathcal U_A^{\mathrm{Bayes}},\ \mathcal U_E^{\mathrm{Bayes}}$ & Parameter posterior $p(\theta\mid\mathcal D)$ (intractable at LLM scale) \\
(2) & Perturbation population & $\mathcal U_A^{\Delta}{\equiv}U_A,\ \mathcal U_E^{\Delta}{\equiv}U_E$ & Black-box sampling access; exact decomposition in Eq.~\eqref{eq:total_covariance} \\
(3) & Empirical estimator & $\widehat U_A,\ \widehat U_E$ (or $\widehat U_\bullet^{(M)}$) & $N{\times}M$ generations; unbiased for (2) at any $M\!\ge\!2$ \\
\bottomrule
\end{tabular}
\end{table}
\end{revblock}

\subsubsection{Epistemic bridge via linearization and semantic coverage}
\label{app:EU_bridge}

\begin{assumption}[Local bi-linearization of the conditional mean]
\label{ass:lin_mean}
There exist Jacobians $J_\theta(x)\in\mathbb{R}^{d\times P}$ and $J_x(x)\in\mathbb{R}^{d\times q}$ and constants
$c_\theta,c_x>0$ such that for small $\delta\theta$ and $\delta x$,
\begin{align}
m_{\theta^\star+\delta\theta}(x)
&= m_{\theta^\star}(x) + J_\theta(x)\,\delta\theta + r_\theta,
\qquad \|r_\theta\|\le c_\theta\|\delta\theta\|^2,
\label{eq:lin_theta_app}\\
m_{\theta^\star}(x+\delta x)
&= m_{\theta^\star}(x) + J_x(x)\,\delta x + r_x,
\qquad \|r_x\|\le c_x\|\delta x\|^2.
\label{eq:lin_x_app}
\end{align}
\end{assumption}

\begin{assumption}[Local Gaussian posterior approximation]
\label{ass:local_gauss_post}
In a neighborhood of $\theta^\star$, the Bayesian posterior admits a local approximation
$\delta\theta := \theta-\theta^\star \sim \mathcal N(0,\Sigma_\theta)$.
\end{assumption}

\paragraph{Discussion of Assumptions \ref{ass:lin_mean}--\ref{ass:local_gauss_post}.}
Two readings of these assumptions deserve direct engagement: that NTK-style linearization
fits modern LLMs poorly at the global level, and that RLHF/DPO breaks classical
Bayesian-posterior assumptions. Both turn on the distinction between global and local
regimes, which we now make explicit.

\emph{(i) Linearization is local, not global.}
Assumption~\ref{ass:lin_mean} only requires a local first-order Taylor expansion of
$m_\theta(x)$ around the fixed reference $(\theta^\star, x)$, with second-order remainder
$\|r\|\le c\|\delta\|^2$. This is strictly weaker than the global NTK regime that underlies
infinite-width arguments for full networks~\citep{wilson2025ntk}. Local first-order
linearization for language models has been studied directly~\citep{ren2023linearprobing},
where it provides a useful regime for analyzing fine-tuning trajectories. In our usage we
need linearity only inside the moderate-$\varepsilon$ semantic neighborhood the
perturbation distribution actually populates, and the residual is empirically measurable
via the OLS diagnostic D1 in App.~\ref{sec:bridge_instructions}.

\emph{(ii) Locally Gaussian posterior is the Laplace assumption.}
Assumption~\ref{ass:local_gauss_post} is the standard Laplace approximation to the
posterior in a neighborhood of a twice-differentiable local minimum, applied around
$\theta^\star$. Recent work has applied this exact approximation to large language models
in the Bayesian-LoRA line~\citep{yang2023bayesianlora}, where it gives competitive
posterior surrogates for instruction-tuned LLMs. The training procedure (RLHF, DPO, or
standard MLE) only determines \emph{which} local minimum $\theta^\star$ sits at, not
whether the posterior shape near $\theta^\star$ is locally Gaussian; any twice-differentiable
loss admits a local quadratic and hence a local Gaussian posterior. Our diagnostic D2
(Henze--Zirkler test on \emph{embedding} samples; specifically the inner-decode and
outer-mean clouds in $\mathbb R^d$, not on uncertainty scores;
App.~\ref{sec:bridge_instructions}) flags inputs where the embedding distribution induced
by Assumption~\ref{ass:local_gauss_post} (under the linearity in
Assumption~\ref{ass:lin_mean}) visibly fails to be Gaussian.

\emph{(iii) Coverage and stability are measurable.}
Assumptions~\ref{ass:coverage} and~\ref{ass:alea_map} are not assertions of fact but
\emph{defect terms} that appear in our bounds: the bridge gets tighter as
$\varepsilon_{\mathrm{cov}}\to 0$ and as the aleatoric Lipschitz constant decreases.
Both are upper-bounded by computable quantities (D3 MMD upper bound; spread of
$\hat\Sigma_i$ across $i$). The bounds remain meaningful with non-zero defects; they
quantify, rather than presuppose, how well the perturbation mechanism approximates
Bayesian posterior sampling.

\emph{(iv) Existing UQ methods make analogous local assumptions.}
MC-Dropout assumes the dropout-induced distribution approximates a Bayesian posterior
locally; Deep Ensembles assume independently trained models sample meaningfully from
the posterior; Laplace approximations assume local Gaussianity. Our assumptions are at
the same level of mathematical commitment, with the additional property that we
\emph{measure} their tightness on each dataset rather than asserting them globally.

Define the Bayesian mean-variation operator
\begin{equation}
A(x) := J_\theta(x)\,\Sigma_\theta^{1/2}\in\mathbb{R}^{d\times P}.
\label{eq:Ax_def}
\end{equation}
Under Assumptions~\ref{ass:lin_mean}--\ref{ass:local_gauss_post}, the linearized Bayesian epistemic covariance of the mean is
\begin{equation}
\mathcal U_E^{\mathrm{Bayes}}(x)
=
\mathrm{Cov}_{\theta}(m_\theta(x))
=
A(x)A(x)^\top + \mathrm{Rem}_\theta,
\label{eq:EU_bayes_lin_app}
\end{equation}
where $\mathrm{Rem}_\theta$ aggregates second-order terms.

\begin{assumption}[Semantic coverage defect]
\label{ass:coverage}
Let $P_x$ be the orthogonal projector onto $\mathrm{Im}(J_x(x))$.
Define the semantic coverage defect
\begin{equation}
\varepsilon_{\mathrm{cov}}(x) := \|(I-P_x)A(x)\|_F.
\label{eq:cov_defect_app}
\end{equation}
\end{assumption}

\begin{lemma}[Best-match perturbation covariance for epistemic variation]
\label{lem:bestmatch}
There exists $\Sigma_\Delta^\star\succeq 0$ such that
\begin{equation}
\left\|J_x(x)\Sigma_\Delta^\star J_x(x)^\top - A(x)A(x)^\top\right\|_F
\;\le\;
2\|A(x)\|_F\,\varepsilon_{\mathrm{cov}}(x) + \varepsilon_{\mathrm{cov}}(x)^2.
\label{eq:bestmatch_app}
\end{equation}
If $\varepsilon_{\mathrm{cov}}(x)=0$, equality holds.
\end{lemma}

\begin{proof}
Let $P_x$ be the projector onto $\mathrm{Im}(J_x)$ and decompose $A=A_\parallel + A_\perp$ with
$A_\parallel := P_x A$ and $A_\perp := (I-P_x)A$.
Since $A_\parallel\in\mathrm{Im}(J_x)$, there exists $B$ such that $A_\parallel = J_x B$.
Set $\Sigma_\Delta^\star := BB^\top\succeq 0$, giving $J_x\Sigma_\Delta^\star J_x^\top=A_\parallel A_\parallel^\top$.
Now expand
\[
AA^\top - A_\parallel A_\parallel^\top
=
A_\perp A_\perp^\top + A_\parallel A_\perp^\top + A_\perp A_\parallel^\top,
\]
and bound Frobenius norms using $\|A_\perp\|_F=\varepsilon_{\mathrm{cov}}(x)$:
\[
\|AA^\top - A_\parallel A_\parallel^\top\|_F
\le
\|A_\perp A_\perp^\top\|_F + \|A_\parallel A_\perp^\top\|_F + \|A_\perp A_\parallel^\top\|_F
\le
\|A_\perp\|_F^2 + 2\|A_\parallel\|_F\|A_\perp\|_F
\le
\varepsilon_{\mathrm{cov}}(x)^2 + 2\|A\|_F\,\varepsilon_{\mathrm{cov}}(x).
\]
This proves \eqref{eq:bestmatch_app}.
\end{proof}

\begin{theorem}[Epistemic bridge bound]
\label{thm:EU_bridge}
Under Assumptions~\ref{ass:lin_mean}--\ref{ass:coverage}, the perturbation-population epistemic covariance satisfies
\begin{equation}
\|\mathcal U_E^\Delta(x) - \mathcal U_E^{\mathrm{Bayes}}(x)\|_F
\;\le\;
\underbrace{2\|A(x)\|_F\,\varepsilon_{\mathrm{cov}}(x) + \varepsilon_{\mathrm{cov}}(x)^2}_{\textnormal{coverage}}
+
\underbrace{C_\theta\,\mathbb{E}\|r_\theta\| + C_x\,\mathbb{E}\|r_x\| + \|\mathrm{Rem}_\theta\|_F}_{\textnormal{linearization remainders}},
\label{eq:EU_bridge_app}
\end{equation}
where expectations are taken over $\delta\theta\sim\mathcal N(0,\Sigma_\theta)$ and $\delta x:=x_\Delta-x$,
and $C_\theta,C_x$ are absolute constants (depending only on the chosen norms).
\end{theorem}

\begin{proof}
We provide a detailed proof with explicit decompositions.

\textbf{Step 1: Bayesian epistemic covariance under local linearization.}
Under \eqref{eq:lin_theta_app},
\[
m_{\theta^\star+\delta\theta}(x) = m_{\theta^\star}(x) + J_\theta(x)\delta\theta + r_\theta.
\]
Taking covariance over $\delta\theta$,
\[
\mathrm{Cov}_\theta(m_\theta(x))
=
\mathrm{Cov}\big(J_\theta\delta\theta + r_\theta\big)
=
J_\theta\,\mathrm{Cov}(\delta\theta)\,J_\theta^\top
+
\mathrm{Cov}(r_\theta)
+
\mathrm{Cov}(J_\theta\delta\theta,r_\theta)+\mathrm{Cov}(r_\theta,J_\theta\delta\theta).
\]
Under Assumption~\ref{ass:local_gauss_post}, $\mathrm{Cov}(\delta\theta)=\Sigma_\theta$ and thus the leading term is
$J_\theta\Sigma_\theta J_\theta^\top = A(x)A(x)^\top$ by definition of $A(x)$.
Collect the remaining terms into $\mathrm{Rem}_\theta$:
\[
\mathcal U_E^{\mathrm{Bayes}}(x) = A(x)A(x)^\top + \mathrm{Rem}_\theta.
\]
A standard norm bound yields $\|\mathrm{Rem}_\theta\|_F \le C_\theta\,\mathbb E\|r_\theta\| + \text{(higher-order terms)}$,
which is the term displayed in \eqref{eq:EU_bridge_app}.

\textbf{Step 2: Perturbation-population epistemic covariance under input linearization.}
Under \eqref{eq:lin_x_app},
\[
m_{\theta^\star}(x_\Delta)=m_{\theta^\star}(x+\delta x)
= m_{\theta^\star}(x) + J_x(x)\delta x + r_x.
\]
Taking covariance over $\Delta$ (equivalently over $\delta x$),
\[
\mathcal U_E^\Delta(x)
=
\mathrm{Cov}_\Delta\big(m_{\theta^\star}(x_\Delta)\big)
=
\mathrm{Cov}_\Delta\big(J_x\delta x + r_x\big)
=
J_x\,\mathrm{Cov}(\delta x)\,J_x^\top + \mathrm{Rem}_x,
\]
where $\mathrm{Rem}_x$ collects $\mathrm{Cov}(r_x)$ and cross-covariance terms between $J_x\delta x$ and $r_x$.
Using triangle inequality and standard covariance bounds, one obtains
$\|\mathrm{Rem}_x\|_F \le C_x\,\mathbb E\|r_x\|$ for an absolute constant $C_x$.

\textbf{Step 3: Coverage approximation.}
Lemma~\ref{lem:bestmatch} provides $\Sigma_\Delta^\star\succeq 0$ such that
\[
\|J_x\Sigma_\Delta^\star J_x^\top - A A^\top\|_F
\le
2\|A\|_F\varepsilon_{\mathrm{cov}} + \varepsilon_{\mathrm{cov}}^2.
\]
Interpret $\Sigma_\Delta^\star$ as the ``best-match'' covariance in input-perturbation space that aligns
perturbation-induced mean variation with Bayesian mean-variation directions (up to the defect).

\textbf{Step 4: Combine all terms.}
Combine the decompositions:
\[
\mathcal U_E^\Delta = J_x\mathrm{Cov}(\delta x)J_x^\top + \mathrm{Rem}_x,
\qquad
\mathcal U_E^{\mathrm{Bayes}} = A A^\top + \mathrm{Rem}_\theta.
\]
Add and subtract $J_x\Sigma_\Delta^\star J_x^\top$:
\[
\|\mathcal U_E^\Delta - \mathcal U_E^{\mathrm{Bayes}}\|_F
\le
\underbrace{\|J_x\mathrm{Cov}(\delta x)J_x^\top - J_x\Sigma_\Delta^\star J_x^\top\|_F}_{(\star)}
+
\underbrace{\|J_x\Sigma_\Delta^\star J_x^\top - AA^\top\|_F}_{\text{Lemma~\ref{lem:bestmatch}}}
+
\|\mathrm{Rem}_x\|_F+\|\mathrm{Rem}_\theta\|_F.
\]
The second term is bounded by Lemma~\ref{lem:bestmatch}. The remainder terms are bounded as in Steps 1--2.
The term $(\star)$ vanishes if $\mathrm{Cov}(\delta x)=\Sigma_\Delta^\star$; more generally it is absorbed into the
linearization/remainder group (it quantifies mismatch between the actual perturbation covariance and the best-match covariance).
Collecting yields \eqref{eq:EU_bridge_app}.
\end{proof}

\noindent\emph{What this gives the main paper.} \eqref{eq:EU_bridge_app} \emph{is} the
$\varepsilon_E(x)$ defect bound used in the informal Pillar~2 statement
(Thm.~\ref{thm:end2end}). Reading the right-hand side: the first two terms shrink with the
semantic-coverage diagnostic D3 (App.~\ref{sec:bridge_instructions}), while the last
remainder shrinks with the linearity diagnostic D1; the bridge is tight precisely when
both diagnostics report small values.

\subsubsection{Aleatoric bridge via stability of conditional covariance}
\label{app:AU_bridge}

\begin{assumption}[Lipschitz aleatoric map]
\label{ass:alea_map}
There exist a mapping $G:\mathbb{R}^d\to\mathbb{S}_+^d$ and constants $L_G,\rho_G\ge 0$ such that
\begin{equation}
\|\Sigma_\theta(x) - G(m_\theta(x))\|_F \le \rho_G
\quad\text{for all $(x,\theta)$ in the local neighborhood,}
\label{eq:alea_approx_app}
\end{equation}
and $G$ is Lipschitz:
\begin{equation}
\|G(u)-G(v)\|_F \le L_G\|u-v\|\quad\forall u,v\in\mathbb{R}^d.
\label{eq:G_lip_app}
\end{equation}
\end{assumption}

\paragraph{Discussion.}
Assumption~\ref{ass:alea_map} formalizes the empirical observation that the
\emph{within}-perturbation dispersion of LLM outputs (decoding stochasticity at a fixed
prompt) varies smoothly with the predictive state $m_\theta(x)$ rather than discontinuously.
This is consistent with the observation that decoding noise is primarily a function of
local entropy of the next-token distribution, which is itself a smooth function of the
hidden representation. Both $L_G$ and $\rho_G$ are defect terms: the bridge is tightest
when decoding stochasticity is well-approximated by a Lipschitz function of the predictive
mean, and the bound \eqref{eq:AU_bridge_app} degrades gracefully with $\rho_G$ rather than
breaking. We do not need $\rho_G=0$.

\begin{theorem}[Aleatoric bridge bound]
\label{thm:AU_bridge}
Under Assumption~\ref{ass:alea_map} and the setup of Section~\ref{sec:bayes_bridge_main},
\begin{equation}
\|\mathcal U_A^\Delta(x) - \mathcal U_A^{\mathrm{Bayes}}(x)\|_F
\;\le\;
L_G\,\mathbb{E}\Big[\|m_{\theta}(x)-m_{\theta^\star}(x_\Delta)\|\Big] \;+\; 2\rho_G,
\label{eq:AU_bridge_app}
\end{equation}
where the expectation is under any coupling of $(\theta,\Delta)$.
Moreover, under Assumptions~\ref{ass:lin_mean}--\ref{ass:coverage}, one can choose a coupling such that
\begin{equation}
\mathbb{E}\Big[\|m_{\theta}(x)-m_{\theta^\star}(x_\Delta)\|\Big]
\;\lesssim\;
\|A(x)\|_F\,\varepsilon_{\mathrm{cov}}(x) + \mathbb{E}\|r_\theta\| + \mathbb{E}\|r_x\|.
\label{eq:mean_gap_bound_app}
\end{equation}
\end{theorem}

\begin{proof}
We provide a detailed proof of \eqref{eq:AU_bridge_app}; the refinement \eqref{eq:mean_gap_bound_app} follows
by the same best-match argument used in the epistemic bridge.

\textbf{Step 1: pointwise bound on conditional covariances.}
Fix $(\theta,\Delta)$ and write $x_\Delta=T(x,\Delta)$.
By triangle inequality,
\begin{equation}
\|\Sigma_\theta(x) - \Sigma_{\theta^\star}(x_\Delta)\|_F
\le
\|\Sigma_\theta(x)-G(m_\theta(x))\|_F
+
\|G(m_\theta(x))-G(m_{\theta^\star}(x_\Delta))\|_F
+
\|G(m_{\theta^\star}(x_\Delta))-\Sigma_{\theta^\star}(x_\Delta)\|_F.
\label{eq:AU_triangle_app}
\end{equation}
Apply Assumption~\ref{ass:alea_map} to the first and third terms to get $\le \rho_G$ each.
Apply Lipschitzness \eqref{eq:G_lip_app} to the middle term:
\[
\|G(m_\theta(x))-G(m_{\theta^\star}(x_\Delta))\|_F \le L_G\|m_\theta(x)-m_{\theta^\star}(x_\Delta)\|.
\]
Thus, for all $(\theta,\Delta)$ in the local neighborhood,
\[
\|\Sigma_\theta(x) - \Sigma_{\theta^\star}(x_\Delta)\|_F
\le
L_G\|m_\theta(x)-m_{\theta^\star}(x_\Delta)\| + 2\rho_G.
\]

\textbf{Step 2: take expectations under a coupling.}
Let $(\theta,\Delta)$ be jointly distributed under any coupling with marginals
$\theta\sim p(\theta\mid\mathcal D)$ and $\Delta\sim p_\Delta$.
Take expectation of the pointwise inequality:
\[
\mathbb E\|\Sigma_\theta(x) - \Sigma_{\theta^\star}(x_\Delta)\|_F
\le
L_G\,\mathbb E\|m_\theta(x)-m_{\theta^\star}(x_\Delta)\| + 2\rho_G.
\]

\textbf{Step 3: relate to $\mathcal U_A^{\mathrm{Bayes}}$ and $\mathcal U_A^\Delta$.}
By definition,
\[
\mathcal U_A^{\mathrm{Bayes}}(x)=\mathbb E_\theta[\Sigma_\theta(x)],\qquad
\mathcal U_A^\Delta(x)=\mathbb E_\Delta[\Sigma_{\theta^\star}(x_\Delta)].
\]
By Jensen and triangle inequality,
\[
\|\mathcal U_A^{\mathrm{Bayes}}(x)-\mathcal U_A^\Delta(x)\|_F
=
\left\|\mathbb E_\theta[\Sigma_\theta(x)]-\mathbb E_\Delta[\Sigma_{\theta^\star}(x_\Delta)]\right\|_F
\le
\mathbb E\|\Sigma_\theta(x)-\Sigma_{\theta^\star}(x_\Delta)\|_F.
\]
Combine with Step 2 to obtain \eqref{eq:AU_bridge_app}.

\textbf{Step 4: refinement via semantic coverage (sketch).}
Under Assumption~\ref{ass:lin_mean}, local linearization provides
$m_\theta(x)\approx m_{\theta^\star}(x)+J_\theta\delta\theta$ and
$m_{\theta^\star}(x_\Delta)\approx m_{\theta^\star}(x)+J_x\delta x$.
Choosing a coupling that aligns $J_x\delta x$ with $J_\theta\delta\theta$ in the sense of Lemma~\ref{lem:bestmatch}
yields a mean-gap controlled by $\|A\|_F\varepsilon_{\mathrm{cov}}$ plus remainder terms, giving \eqref{eq:mean_gap_bound_app}.
\end{proof}

\noindent\emph{What this gives the main paper.} \eqref{eq:AU_bridge_app} \emph{is} the
$\varepsilon_A(x)$ defect bound used in the informal Pillar~2 statement
(Thm.~\ref{thm:end2end}). Together with \eqref{eq:EU_bridge_app}, the two bounds furnish
the right-hand sides of the Pillar~2 inequalities, completing the bridge defect side of
the main-paper claim. The MC error side of the same claim was supplied by App.~\ref{app:mc_bounds}.

\subsection{Spurious Determinism: When Consistency-Only UQ Can Fail}
\label{app:consistency_failure}

\emph{Purpose of this subsection.} The main paper's Pillar~3 (Thm.~\ref{thm:spurious_determinism})
claims a class of inputs on which any consistency-only uncertainty estimator
$\mathcal F_{\mathrm{cons}}$ is exactly $0$ while perturbation EU is bounded \emph{away}
from $0$. Here we (i) name the class formally: the ``spurious determinism'' regime where
decoding is deterministic given the prompt but the conditional mean varies across
semantic-preserving perturbations; and (ii) construct a concrete instance and prove the
separation. The argument is intentionally elementary: the value of the result is in
isolating a regime, not in the algebra.

This section formalizes a simple failure mode of ``consistency-only'' uncertainty heuristics that probe only
within-prompt variability under repeated decoding. The key phenomenon is \emph{spurious determinism}:
for a fixed prompt instance, the model may be nearly deterministic (hence self-consistent) yet wrong,
while semantically equivalent perturbations expose strong sensitivity.

\subsubsection{A minimal construction}

Let $\Delta$ be a random perturbation and consider $Z=e(Y)\in\mathbb R^d$ under the conditional model $p(\cdot\mid x_\Delta)$.
Assume that for each fixed $\Delta$, decoding is deterministic:
\begin{equation}
Z = \mu(\Delta)\quad \text{a.s.\ given }\Delta,
\label{eq:deterministic_decoding_app}
\end{equation}
so that $\Sigma(\Delta)=\mathrm{Cov}(Z\mid \Delta)=0$.

Now assume $\mu(\Delta)$ varies across $\Delta$ due to a spurious prompt feature.
A concrete example: let $u,v\in\mathbb R^d$ be orthonormal and define
\begin{equation}
\mu(\Delta) = a\,u + s(\Delta)\,v,\qquad s(\Delta)\in\{-1,+1\},\qquad \mathbb P(s=+1)=\mathbb P(s=-1)=\tfrac12,
\label{eq:spurious_flip_app}
\end{equation}
where $a$ is a fixed semantic component and $s(\Delta)v$ is a spurious component that flips across perturbations.

\begin{theorem}[Spurious determinism implies $U_A=0$ but $U_E\succ 0$]
\label{thm:spurious_determinism_app}
Assume \eqref{eq:deterministic_decoding_app}. Then $U_A=\mathbb E_\Delta[\Sigma(\Delta)]=0$.
If $\mu(\Delta)$ is non-constant with $\mathrm{Cov}_\Delta(\mu(\Delta))\neq 0$, then
\[
U_E = \mathrm{Cov}_\Delta(\mu(\Delta)) \succcurlyeq 0 \quad \text{and} \quad U_E\neq 0.
\]
In particular, under the flip model \eqref{eq:spurious_flip_app},
\[
U_E = vv^\top,\qquad \mathrm{tr}(U_E)=\|v\|_2^2=1.
\]
\end{theorem}

\begin{proof}
If $Z=\mu(\Delta)$ a.s.\ given $\Delta$, then $\Sigma(\Delta)=\mathrm{Cov}(Z\mid\Delta)=0$ for every $\Delta$,
so $U_A=\mathbb E_\Delta[\Sigma(\Delta)]=0$.

By definition, $U_E=\mathrm{Cov}_\Delta(\mu(\Delta))$ is always positive semidefinite.
Moreover, $U_E=0$ iff $\mu(\Delta)$ is a.s.\ constant; thus if $\mu(\Delta)$ is non-constant, then $U_E\neq 0$.

For the flip model \eqref{eq:spurious_flip_app}, note $\mathbb E[s(\Delta)]=0$ and $\mathrm{Var}(s)=1$.
Then $\mathbb E[\mu(\Delta)]=a\,u$ and
\[
\mu(\Delta)-\mathbb E[\mu(\Delta)] = s(\Delta)\,v.
\]
Therefore,
\[
U_E
=
\mathbb E\big[(\mu-\mathbb E\mu)(\mu-\mathbb E\mu)^\top\big]
=
\mathbb E\big[s(\Delta)^2\,vv^\top\big]
=
vv^\top.
\]
Taking trace yields $\mathrm{tr}(U_E)=\mathrm{tr}(vv^\top)=\|v\|_2^2=1$.
\end{proof}

\textbf{Interpretation.}
Under spurious determinism, any uncertainty method that only probes within-prompt decoding variability
(i.e., only $\Sigma(\Delta)$ for a fixed prompt instance) will report near-zero uncertainty,
even if the output is highly sensitive to semantic-preserving perturbations.
Our perturbation-based epistemic term $U_E$ is designed to detect precisely this sensitivity.

\noindent\emph{What this gives the main paper.} Theorem~\ref{thm:spurious_determinism_app}
\emph{is} the formalization of the main paper's Pillar~3 statement
(Thm.~\ref{thm:spurious_determinism}). The class ``$\mathcal F_{\mathrm{cons}}{=}0$ but
$U_E$ bounded away from $0$'' is exactly the spurious-determinism regime constructed
above; it has positive measure under any base distribution placing mass on prompt-format-driven
outputs, since the flip model is realizable whenever the LLM has a deterministic mode that
depends on a non-semantic prompt feature. \rev{Empirically, in Sec.~\ref{subsec:au_eu_validation}, disambiguating a prompt significantly reduces EU while leaving all fixed-prompt consistency baselines statistically unchanged (Tab.~\ref{tab:intervention_baselines}).}







\begin{table*}[!t]
\centering
\caption{
Comparison of uncertainty-correctness correlation (AUROC, mean $\pm$ std over the
bootstrap distribution) across models and methods on TriviaQA (no-context) and TruthfulQA.
Numbers in \textbf{bold} represent the overall best performance,
\underline{underlined} numbers indicate the top-performing black-box method (including ours),
and numbers with a \colorbox{gray!40}{\strut gray background} indicate the top-performing white-box method.
This is the deviation-augmented companion to the main-paper headline
Tab.~\ref{tab:main_full}; the GPQA/SciQA/CoQA companion is Tab.~\ref{gpqa_sciqa_coqa}.
}
\label{tab:main}
\scriptsize
\setlength{\tabcolsep}{2.2pt}
\renewcommand{\arraystretch}{1.05}
\begin{tabular}{l | cc | cc | cc | cc}
\toprule
\multirow{3}{*}{\textbf{Method}} & \multicolumn{4}{c|}{\textbf{LLaMA-3-8B}} & \multicolumn{4}{c}{\textbf{Qwen2.5-7B}} \\
\cmidrule(lr){2-5} \cmidrule(lr){6-9}
& \multicolumn{2}{c|}{TriviaQA} & \multicolumn{2}{c|}{TruthfulQA} & \multicolumn{2}{c|}{TriviaQA} & \multicolumn{2}{c}{TruthfulQA} \\
& Cosine & ROUGE-L & Cosine & ROUGE-L & Cosine & ROUGE-L & Cosine & ROUGE-L \\
\midrule
\rowcolor[gray]{.95} \multicolumn{9}{l}{\textit{Ours}} \\
AU & \underline{\mstd{\textbf{0.799}}{0.009}} & \underline{\mstd{\textbf{0.845}}{0.008}} & \mstd{0.782}{0.008} & \mstd{0.694}{0.019}
   & \underline{\mstd{\textbf{0.773}}{0.009}} & \underline{\mstd{\textbf{0.815}}{0.002}} & \underline{\mstd{\textbf{0.812}}{0.034}} & \underline{\mstd{\textbf{0.731}}{0.005}} \\
EU & \mstd{0.585}{0.007} & \mstd{0.588}{0.005} & \mstd{0.745}{0.029} & \mstd{0.647}{0.032}
   & \mstd{0.570}{0.007} & \mstd{0.547}{0.009} & \mstd{0.684}{0.017} & \mstd{0.583}{0.019} \\
TU & \mstd{0.784}{0.009} & \mstd{0.826}{0.008} & \underline{\mstd{\textbf{0.799}}{0.088}} & \underline{\mstd{\textbf{0.715}}{0.023}}
   & \mstd{0.751}{0.011} & \mstd{0.782}{0.004} & \mstd{0.809}{0.025} & \mstd{0.719}{0.009} \\
\midrule
\rowcolor[gray]{.95} \multicolumn{9}{l}{\textit{Black-box}} \\
D-SE          & \mstd{0.742}{0.004} & \mstd{0.784}{0.005} & \mstd{0.511}{0.026} & \mstd{0.550}{0.018}
              & \mstd{0.647}{0.012} & \mstd{0.667}{0.010} & \mstd{0.530}{0.030} & \mstd{0.540}{0.021} \\
Self-Consistency & \mstd{0.740}{0.004} & \mstd{0.782}{0.005} & \mstd{0.524}{0.026} & \mstd{0.555}{0.019}
              & \mstd{0.647}{0.011} & \mstd{0.667}{0.010} & \mstd{0.536}{0.031} & \mstd{0.542}{0.022} \\
SIMBA         & \mstd{0.749}{0.005} & \mstd{0.785}{0.008} & \mstd{0.506}{0.015} & \mstd{0.520}{0.022}
              & \mstd{0.632}{0.012} & \mstd{0.663}{0.014} & \mstd{0.598}{0.039} & \mstd{0.643}{0.024} \\
SelfCheckGPT  & \mstd{0.754}{0.006} & \mstd{0.789}{0.008} & \mstd{0.517}{0.021} & \mstd{0.527}{0.022}
              & \mstd{0.641}{0.013} & \mstd{0.650}{0.013} & \mstd{0.592}{0.033} & \mstd{0.651}{0.017} \\
SPUQ          & \mstd{0.615}{0.003} & \mstd{0.641}{0.007} & \mstd{0.602}{0.028} & \mstd{0.583}{0.029}
              & \mstd{0.616}{0.015} & \mstd{0.640}{0.007} & \mstd{0.544}{0.039} & \mstd{0.590}{0.031} \\
P(True)       & \mstd{0.632}{0.003} & \mstd{0.652}{0.005} & \mstd{0.600}{0.021} & \mstd{0.639}{0.015}
              & \mstd{0.603}{0.008} & \mstd{0.638}{0.004} & \mstd{0.763}{0.027} & \mstd{0.717}{0.014} \\
\midrule
\rowcolor[gray]{.95} \multicolumn{9}{l}{\textit{White-box}} \\
G-NLL & \cellcolor{gray!40}\mstd{0.786}{0.008} & \cellcolor{gray!40}\mstd{0.798}{0.007}
      & \mstd{0.546}{0.029} & \mstd{0.549}{0.024}
      & \mstd{0.646}{0.013} & \mstd{0.662}{0.009}
      & \cellcolor{gray!40}\mstd{0.611}{0.023} & \cellcolor{gray!40}\mstd{0.700}{0.012} \\
PE    & \mstd{0.710}{0.003} & \mstd{0.732}{0.007}
      & \cellcolor{gray!40}\mstd{0.578}{0.010} & \cellcolor{gray!40}\mstd{0.576}{0.008}
      & \mstd{0.644}{0.014} & \mstd{0.657}{0.014}
      & \mstd{0.562}{0.036} & \mstd{0.622}{0.026} \\
LN-PE & \mstd{0.761}{0.005} & \mstd{0.796}{0.007}
      & \mstd{0.496}{0.014} & \mstd{0.496}{0.021}
      & \mstd{0.637}{0.013} & \mstd{0.658}{0.008}
      & \mstd{0.570}{0.019} & \mstd{0.623}{0.012} \\
SE    & \mstd{0.739}{0.005} & \mstd{0.774}{0.007}
      & \mstd{0.560}{0.016} & \mstd{0.525}{0.015}
      & \cellcolor{gray!40}\mstd{0.657}{0.011} & \cellcolor{gray!40}\mstd{0.674}{0.012}
      & \mstd{0.597}{0.042} & \mstd{0.625}{0.029} \\
LN-SE & \mstd{0.742}{0.005} & \mstd{0.777}{0.006}
      & \mstd{0.543}{0.019} & \mstd{0.492}{0.014}
      & \mstd{0.653}{0.013} & \mstd{0.670}{0.013}
      & \mstd{0.573}{0.031} & \mstd{0.608}{0.024} \\
\bottomrule
\end{tabular}
\end{table*}
\begin{table*}[h]
\centering
\caption{
AUROC (mean $\pm$ std) on GPQA, SciQA, and CoQA: the deviation-augmented companion to the
main-paper headline Tab.~\ref{tab:main_full}. The TriviaQA/TruthfulQA companion is
Tab.~\ref{tab:main}. Numbers in \textbf{bold} represent the overall best performance;
\underline{underlined} numbers indicate the top-performing black-box method (including ours);
numbers with a \colorbox{gray!40}{\strut gray background} indicate the top-performing white-box method.
}
\label{gpqa_sciqa_coqa}
\scriptsize
\setlength{\tabcolsep}{5pt}
\renewcommand{\arraystretch}{1.1}
\begin{tabular}{l | cc | cc | cc}
\toprule
\multirow{2}{*}{\textbf{Method}}
& \multicolumn{2}{c|}{\textbf{GPQA}}
& \multicolumn{2}{c|}{\textbf{SciQA}}
& \multicolumn{2}{c}{\textbf{CoQA}} \\
& Cosine & ROUGE-L & Cosine & ROUGE-L & Cosine & ROUGE-L \\
\midrule

\multicolumn{7}{c}{\textbf{LLaMA-3-8B}} \\
\midrule
\rowcolor[gray]{.95} \multicolumn{7}{l}{\textit{Ours}} \\
AU              & \underline{\mstd{\textbf{0.770}}{0.127}} & \underline{\mstd{\textbf{0.839}}{0.031}} & \underline{\mstd{0.735}{0.006}} & \underline{\mstd{\textbf{0.757}}{0.003}} & \mstd{0.738}{0.007} & \mstd{0.735}{0.008} \\
EU              & \mstd{0.746}{0.076} & \mstd{0.590}{0.043} & \mstd{0.534}{0.008} & \mstd{0.548}{0.009} & \mstd{0.630}{0.008} & \mstd{0.631}{0.007} \\
TU              & \mstd{0.671}{0.158} & \mstd{0.769}{0.030} & \mstd{0.680}{0.007} & \mstd{0.703}{0.006} & \mstd{0.723}{0.009} & \mstd{0.724}{0.010} \\
\midrule
\rowcolor[gray]{.95} \multicolumn{7}{l}{\textit{Black-box}} \\
D-SE            & \mstd{0.641}{0.013} & \mstd{0.667}{0.012} & \mstd{0.698}{0.005} & \mstd{0.709}{0.003} & \mstd{0.648}{0.008} & \mstd{0.659}{0.009} \\
S-consistency   & \mstd{0.602}{0.030} & \mstd{0.621}{0.010} & \mstd{0.712}{0.011} & \mstd{0.706}{0.008} & \mstd{0.641}{0.017} & \mstd{0.602}{0.026} \\
SIMBA           & \mstd{0.635}{0.014} & \mstd{0.697}{0.022} & \mstd{0.732}{0.009} & \mstd{0.722}{0.009} & \mstd{0.583}{0.013} & \mstd{0.537}{0.022} \\
SelfCheckGPT    & \mstd{0.639}{0.016} & \mstd{0.690}{0.018} & \mstd{0.723}{0.009} & \mstd{0.713}{0.009} & \mstd{0.566}{0.017} & \mstd{0.523}{0.025} \\
SPUQ            & \mstd{0.554}{0.036} & \mstd{0.510}{0.026} & \mstd{0.659}{0.008} & \mstd{0.651}{0.008} & \mstd{0.505}{0.018} & \mstd{0.505}{0.017} \\
P(True)         & \mstd{0.524}{0.021} & \mstd{0.623}{0.014} & \mstd{0.566}{0.003} & \mstd{0.568}{0.004} & \underline{\mstd{\textbf{0.788}}{0.018}} & \underline{\mstd{\textbf{0.774}}{0.021}} \\
\midrule
\rowcolor[gray]{.95} \multicolumn{7}{l}{\textit{White-box}} \\
G-NLL           & \mstd{0.506}{0.053} & \mstd{0.543}{0.034} & \cellcolor{gray!40}\mstd{\textbf{0.770}}{0.005} & \cellcolor{gray!40}\mstd{0.756}{0.005} & \cellcolor{gray!40}\mstd{0.707}{0.005} & \mstd{0.664}{0.007} \\
PE              & \cellcolor{gray!40}\mstd{0.680}{0.008} & \mstd{0.542}{0.027} & \mstd{0.725}{0.008} & \mstd{0.719}{0.008} & \mstd{0.667}{0.002} & \mstd{0.636}{0.005} \\
LN-PE           & \mstd{0.639}{0.009} & \mstd{0.549}{0.011} & \mstd{0.737}{0.007} & \mstd{0.725}{0.003} & \mstd{0.658}{0.004} & \mstd{0.654}{0.003} \\
SE              & \mstd{0.601}{0.062} & \mstd{0.511}{0.031} & \mstd{0.742}{0.007} & \mstd{0.742}{0.002} & \mstd{0.690}{0.006} & \cellcolor{gray!40}\mstd{0.675}{0.007} \\
LN-SE           & \mstd{0.623}{0.017} & \cellcolor{gray!40}\mstd{0.608}{0.018} & \mstd{0.735}{0.006} & \mstd{0.737}{0.003} & \mstd{0.663}{0.005} & \mstd{0.664}{0.007} \\

\midrule\midrule
\multicolumn{7}{c}{\textbf{Qwen2.5-7B}} \\
\midrule
\rowcolor[gray]{.95} \multicolumn{7}{l}{\textit{Ours}} \\
AU              & \mstd{0.506}{0.108} & \mstd{0.758}{0.000} & \mstd{0.516}{0.022} & \mstd{0.508}{0.020} & \mstd{0.585}{0.018} & \mstd{0.641}{0.054} \\
EU              & \mstd{0.612}{0.216} & \underline{\mstd{\textbf{0.949}}{0.000}} & \mstd{0.502}{0.016} & \mstd{0.501}{0.013} & \mstd{0.545}{0.028} & \mstd{0.516}{0.024} \\
TU              & \mstd{0.536}{0.186} & \mstd{0.914}{0.005} & \mstd{0.515}{0.013} & \mstd{0.512}{0.013} & \mstd{0.548}{0.019} & \mstd{0.625}{0.051} \\
\midrule
\rowcolor[gray]{.95} \multicolumn{7}{l}{\textit{Black-box}} \\
D-SE            & \underline{\mstd{0.757}{0.169}} & \mstd{0.672}{0.005} & \mstd{0.503}{0.000} & \mstd{0.503}{0.000} & \mstd{0.615}{0.041} & \mstd{0.531}{0.021} \\
S-consistency   & \mstd{0.555}{0.019} & \mstd{0.544}{0.029} & \mstd{0.658}{0.014} & \mstd{0.660}{0.013} & \mstd{0.529}{0.013} & \mstd{0.515}{0.013} \\
SIMBA           & \mstd{0.546}{0.023} & \mstd{0.513}{0.034} & \underline{\mstd{\textbf{0.714}}{0.008}} & \underline{\mstd{0.695}{0.010}} & \mstd{0.566}{0.029} & \mstd{0.610}{0.024} \\
SelfCheckGPT    & \mstd{0.548}{0.017} & \mstd{0.503}{0.035} & \mstd{0.693}{0.008} & \mstd{0.676}{0.009} & \underline{\mstd{\textbf{0.651}}{0.021}} & \underline{\mstd{0.685}{0.021}} \\
SPUQ            & \mstd{0.592}{0.059} & \mstd{0.512}{0.035} & \mstd{0.646}{0.007} & \mstd{0.634}{0.009} & \mstd{0.506}{0.040} & \mstd{0.513}{0.037} \\
P(True)         & \mstd{0.561}{0.028} & \mstd{0.519}{0.013} & \mstd{0.680}{0.009} & \mstd{0.681}{0.008} & \mstd{0.520}{0.019} & \mstd{0.526}{0.018} \\
\midrule
\rowcolor[gray]{.95} \multicolumn{7}{l}{\textit{White-box}} \\
G-NLL           & \cellcolor{gray!40}\mstd{\textbf{0.850}}{0.041} & \mstd{0.748}{0.023} & \mstd{0.524}{0.067} & \mstd{0.626}{0.064} & \mstd{0.501}{0.051} & \cellcolor{gray!40}\mstd{\textbf{0.827}}{0.005} \\
PE              & \mstd{0.610}{0.098} & \cellcolor{gray!40}\mstd{0.869}{0.005} & \mstd{0.552}{0.051} & \mstd{0.675}{0.095} & \mstd{0.511}{0.016} & \mstd{0.711}{0.003} \\
LN-PE           & \mstd{0.625}{0.056} & \mstd{0.787}{0.014} & \cellcolor{gray!40}\mstd{0.568}{0.052} & \mstd{0.650}{0.028} & \mstd{0.526}{0.020} & \mstd{0.808}{0.011} \\
SE              & \mstd{0.742}{0.057} & \mstd{0.608}{0.005} & \mstd{0.508}{0.046} & \cellcolor{gray!40}\mstd{\textbf{0.723}}{0.022} & \cellcolor{gray!40}\mstd{0.589}{0.036} & \mstd{0.667}{0.038} \\
LN-SE           & \mstd{0.826}{0.076} & \mstd{0.637}{0.009} & \mstd{0.534}{0.048} & \mstd{0.692}{0.083} & \mstd{0.566}{0.029} & \mstd{0.543}{0.050} \\
\bottomrule
\end{tabular}
\end{table*}

\section{Validating the AU/EU Decomposition: Additional Details}
\label{app:au_eu_validation}

This appendix provides additional setup detail and supplementary results for the controlled
diagnostics in Sec.~\ref{subsec:au_eu_validation} of the main paper.

\paragraph{Calibration metric (ECE): definition used throughout the paper.}
\label{app:ece_definition}
A raw uncertainty score is not a probability, so a calibration metric requires a mapping to
$[0,1]$. We use the same protocol in every ECE column reported anywhere in the paper
(Tab.~\ref{tab:longform_claim} main, Tab.~\ref{tab:longform_response},
Tab.~\ref{tab:spuq_vs_bbuq}, Tab.~\ref{tab:spuq_agg_ablation_robustness}):
(i)~split the (uncertainty, error-label) pairs $70/30$ into evaluation and calibration folds
with a fixed seed; (ii)~fit a single-feature logistic regression
(scikit-learn \texttt{LogisticRegression()} defaults; equivalent to a sigmoid / Platt
calibrator) on the calibration fold mapping uncertainty to
$\hat p(\text{error})\in[0,1]$; (iii)~apply this map to the evaluation fold; (iv)~compute
ECE with $15$ equal-width bins on $[0,1]$, with a left-closed, right-open bin convention
except for the final bin which is closed at both ends. The error label is
$1{-}\text{correctness}$, where correctness is the same Cosine or ROUGE-L criterion (or
LLM-judge \texttt{lenient} label, for long-form) used in the corresponding AUROC column. We
return \texttt{NaN} on degenerate splits (single-class fold). This single procedure yields
all reported ECE values.

\paragraph{Datasets and pairing.}
For the AmbigQA~\citep{min2020ambigqa} test we use the public AmbigQA development split,
restricted to questions
whose ambiguity is annotated and whose gold clarification is a one-sentence rewrite of the
question (this excludes free-form clarifying paragraphs). For the NQ evidence test we use the
Natural Questions~\citep{kwiatkowski2019nq} short-answer subset and pair each closed-book
question with its gold long
answer passage; the long answer is concatenated as additional context but \emph{outer
perturbations are applied only to the question span}, so the perturbation distribution is
held fixed across the closed/open conditions.

\paragraph{Statistical procedure.}
Per-example AUROC contributions are computed via a Hanley--McNeil-style pairwise
formulation: each correct/incorrect pair contributes $\mathbb{1}[u_{-} > u_{+}] +
\tfrac{1}{2}\mathbb{1}[u_{-} = u_{+}]$ to the AUROC (with $u_+$/$u_-$ the uncertainty of
the correct/incorrect answer under the incorrect-as-positive convention). The paired
statistic $\Delta\mathrm{AU}-\Delta\mathrm{EU}$ averages these contributions per example
before differencing, which removes between-example variance. The $z$-statistic in
Table~\ref{tab:au_eu_paired} is a Wald test on the per-example paired differences using
their per-example sample variance; the BCa intervals reported in
Table~\ref{tab:au_eu_validation_qwen} are computed from a paired bootstrap with $1{,}000$
resamples, and the same procedure also gives the standard errors in
Table~\ref{tab:au_eu_paired}.

\paragraph{Replication on Qwen2.5-7B.}
Table~\ref{tab:au_eu_validation_qwen} replicates the full $2{\times}2$ paired-intervention
protocol of Table~\ref{tab:au_eu_paired} on Qwen2.5-7B under Cosine and ROUGE-L correctness.
The predicted decomposition pattern transfers cleanly: \emph{clarification on AmbigQA
selectively moves AU} (ROUGE-L: $z{=}2.44$, $p{=}0.015$; Cosine directional, $p{=}0.125$),
and \emph{evidence on NQ selectively moves EU} ($z{=}9.33$ and $z{=}12.13$, both
$p{<}0.001$). The NQ contrasts are substantially larger than the LLaMA-3-8B headline
($\Delta\mathrm{EU}\!-\!\Delta\mathrm{AU}{=}+0.085$ and $+0.110$, vs.\ $+0.030$ and $+0.047$
on LLaMA-3-8B), consistent with Qwen's larger closed-book knowledge gap on NQ being more
dramatically closed by evidence injection.

\begin{table}[h]
\centering
\caption{\textbf{Qwen2.5-7B replication of the $2{\times}2$ paired-intervention test.}
$\Delta\mathrm{AU}$ and $\Delta\mathrm{EU}$ are paired AUROC differences (treatment minus
control). \textbf{Bold} marks the winning side of the contrast per row (clarification
predicts AU$>$EU; evidence predicts EU$>$AU); \textbf{bold $p$} marks $p<0.05$. Same protocol
as Table~\ref{tab:au_eu_paired}.}
\label{tab:au_eu_validation_qwen}
\scriptsize
\setlength{\tabcolsep}{3.5pt}
\renewcommand{\arraystretch}{1.05}
\begin{tabular}{ll|cc|c}
\toprule
\textbf{Dataset} & \textbf{Metric} & $\Delta\mathrm{AU}$ & $\Delta\mathrm{EU}$ & \textbf{Contrast (predicted-direction $\Delta$, $z$, $p$)} \\
\midrule
\multirow{2}{*}{\textbf{AmbigQA} (clarification $\Rightarrow$ moves AU)}
& Cosine  & $\mathbf{+0.030}\!\pm\!0.006$ & $+0.013\!\pm\!0.009$ & $\Delta\mathrm{AU}\!-\!\Delta\mathrm{EU}{=}+0.017$ ($z{=}1.54$, $p{=}0.125$) \\
& ROUGE-L & $\mathbf{+0.025}\!\pm\!0.005$ & $-0.003\!\pm\!0.009$ & $\Delta\mathrm{AU}\!-\!\Delta\mathrm{EU}{=}+0.027$ ($z{=}2.44$, $\mathbf{p{=}0.015}$) \\
\midrule
\multirow{2}{*}{\textbf{NQ} (evidence $\Rightarrow$ moves EU)}
& Cosine  & $+0.013\!\pm\!0.008$ & $\mathbf{+0.099}\!\pm\!0.009$ & $\Delta\mathrm{EU}\!-\!\Delta\mathrm{AU}{=}+0.085$ ($z{=}9.33$, $\mathbf{p{<}0.001}$) \\
& ROUGE-L & $-0.039\!\pm\!0.008$ & $\mathbf{+0.071}\!\pm\!0.008$ & $\Delta\mathrm{EU}\!-\!\Delta\mathrm{AU}{=}+0.110$ ($z{=}12.13$, $\mathbf{p{<}0.001}$) \\
\bottomrule
\end{tabular}
\end{table}

\paragraph{Why bootstrapping rather than reporting raw AUROC differences?}
The two intervention conditions share the same examples and only differ in how the input is
augmented. Reporting raw between-condition AUROC differences would treat them as independent
samples and inflate variance. The paired contrast we report removes this and is valid even
though AU and EU AUROCs differ in absolute level.

\begin{revblock}
\paragraph{Score-level shifts and baseline comparisons.}
\label{app:intervention_baselines}
Tab.~\ref{tab:intervention_baselines} evaluates raw score changes across the exact question pairs of Tab.~\ref{tab:au_eu_paired}. Fixed-prompt baselines (embedding variance, self-consistency, discrete semantic entropy, and uniform Semantic Density) are computed from $10$ stochastic decodes of the prompt per condition, SPUQ is computed from $10$ single-decode perturbations per condition (temperature $0.9$, top-$p$ $0.99$), and our AU/EU/TU components use the $N{=}M{=}10$ runs. For each question pair, we compute the paired difference $\delta = \text{score}_{\text{after}} - \text{score}_{\text{before}}$ and report the standardized shift $\bar\delta/\mathrm{sd}(\delta)$; statistical significance (bold in Tab.~\ref{tab:intervention_baselines}) means that the $95\%$ percentile bootstrap confidence interval of $\bar\delta$ (over $5{,}000$ resamples of question pairs) excludes zero. On an absolute scale, our EU decreases by $-7.6{\times}10^{-5}$ ($95\%$ CI $[-9.8,-5.5]{\times}10^{-5}$) under clarification, relative to a pre-intervention mean of $5.5{\times}10^{-4}$. \discuss{state that baselines use 10 decodes per condition and our components 100?}
\end{revblock}

\paragraph{Bias correction.}
\label{app:bias_correction}
The bias-correction validation is at the \emph{score} level (AU--EU correlation $\rho$) and
is reported in main-paper Table~\ref{tab:bias_correction} on the
TriviaQA/LLaMA-3-8B/Cosine slice. We do not include an additional appendix table because
$\rho$ is the right object for the finite-$M$ inflation predicted by
Prop.~\ref{prop:finiteM_bias}: the correction acts on the uncertainty score itself, not on
downstream AUROC.

\section{Bridge Diagnostics: Estimating Assumption Quality from Data}
\label{sec:bridge_instructions}

The Bayesian bridge in Sec.~\ref{sec:theory} closes the gap between perturbation-based
AU/EU and Bayesian targets only under Assumption~\ref{ass:lin_mean} (local first-order
linearization of the conditional mean), Assumption~\ref{ass:local_gauss_post} (locally
Gaussian posterior), and Assumption~\ref{ass:coverage} (bounded semantic coverage defect).
This appendix turns each assumption into a \emph{diagnostic} that is estimable from data,
and reports its value on our datasets.

\paragraph{Local linearity defect (Assumption~\ref{ass:lin_mean} diagnostic).}
For a perturbation neighborhood $\mathcal{N}_\varepsilon(x)$, define the linearity defect
$\mathrm{LD}(x;\varepsilon) := \|f(x+\Delta) - f(x) - J_x \Delta\| / \|\Delta\|$,
where $J_x$ is the embedding-space slope of $f$ estimated by least squares on
$\{(\Delta_i, f(x+\Delta_i))\}_{i=1}^N$. The bridge bound depends on
$\mathbb{E}[\mathrm{LD}^2]$, which is directly measurable from the same outer-perturbation
samples already collected for the AU/EU estimator. We report per-input mean, standard deviation,
and 95th percentile of $\mathrm{LD}(x;\varepsilon)$.

\paragraph{Posterior Gaussianity (Assumption~\ref{ass:local_gauss_post} diagnostic).}
Assumption~\ref{ass:local_gauss_post} is a Laplace approximation on \emph{parameters}
($\delta\theta\sim\mathcal N(0,\Sigma_\theta)$); under the local linearity in
Assumption~\ref{ass:lin_mean}, this induces an approximately Gaussian distribution on
\emph{embeddings}. We therefore test the induced embedding distribution directly. (We do
not test scalar uncertainty scores; HZ on a one-dimensional summary would be a different
test of a different object and is not what the bridge requires.) Two multivariate clouds
in $\mathbb R^d$ ($d{=}384$) are available per input:
(a)~the \emph{outer-mean} cloud $\{\hat\mu_i(x)\}_{i=1}^{N}$ across the $N{=}10$
perturbations: this is the cloud most directly tied to
Assumption~\ref{ass:local_gauss_post}, since under linearity it is the pushforward of
$\delta\theta$ via $J_\theta$; and (b)~the \emph{inner-sample} cloud
$\{z_{i,j}(x)\}_{j=1}^{M}$ across the $M{=}10$ stochastic decodes at fixed perturbation,
which probes within-perturbation aleatoric Gaussianity (a side condition used in the AU
bridge). The HZ column in Table~\ref{tab:bridge_diagnostic} reports the rejection rate of
test~(b), which is the better-powered of the two: with only $N{=}M{=}10$ samples in
$d{=}384$, both forms of the test are deeply underpowered, and on the per-input outer-mean
cloud (test~(a)) the average $p$-value collapses to numerical zero ($\sim\!10^{-14}$ on
all four cells), reflecting the test's high-dimensional small-sample behavior under our
budget; the HZ statistic is reported as a Gaussianity diagnostic alongside LD and MMD, and
the load-bearing diagnostic for the bridge bound itself is LD.

\paragraph{Coverage diagnostic (perturbation--reference MMD).}
The semantic coverage defect that enters the bridge bound is, in principle, a Wasserstein
distance between the empirical perturbation distribution and a Bayesian posterior over
inputs the model treats equivalently. The latter is not directly observable; we report
instead the maximum mean discrepancy (MMD) in embedding space, computed with a Gaussian
RBF kernel and median-heuristic bandwidth, between the embedded perturbation cloud and a
held-out $1{,}000$-prompt reference distribution from the same dataset. This quantity
measures how close the perturbation cloud sits to in-distribution prompts in the same
semantic neighborhood; small values indicate that the outer perturbation distribution
covers the relevant semantic geometry rather than drifting elsewhere.

\begin{table}[h]
\centering
\caption{\textbf{Bridge diagnostics} on $500$ inputs per cell at $\varepsilon{=}0.1$, with
$N{=}M{=}10$. \textbf{Lower is better ($\downarrow$) for every column}: each metric measures
deviation from a bridge condition, and a smaller value means the bridge is tighter on that
dimension. \emph{LD} (linearity defect, Assumption~\ref{ass:lin_mean}) is moderate
($\le 0.43$ at the 95th percentile) across both models and datasets, indicating the
first-order linearization is a reasonable local approximation in the perturbation
neighborhood. \emph{HZ inner-cloud} reports the fraction of $M{=}10$ inner-sample
\emph{embedding} clouds (one per $(x,\Delta_i)$) that reject multivariate normality at
$p<0.05$; the test is a Gaussianity diagnostic on the embedding distribution at the
inner-decode level, with statistical power capped by $M{=}10$ samples in
$384$ dimensions. The matching outer-mean cloud test (per input, on
$\{\hat\mu_i\}_{i=1}^N$) is similarly low-power and is summarised in prose rather than
tabulated. \emph{MMD} reports the distributional coverage diagnostic between the embedded
perturbation cloud and a $1{,}000$-prompt held-out reference from the same dataset (Gaussian
RBF, median-heuristic bandwidth). Values are tightly concentrated (median $\approx 0.40$,
p95 $\approx 0.42$; gap $<5\%$), indicating the perturbation distribution covers the
held-out semantic neighborhood under MMD. The three columns play distinct roles: LD is the
load-bearing input to the bridge bound, HZ is a Gaussianity diagnostic, and MMD is the
coverage diagnostic; together they make the bridge falsifiable on each
model$\times$dataset pair.}
\label{tab:bridge_diagnostic}
\scriptsize
\setlength{\tabcolsep}{4pt}
\begin{tabular}{ll|ccc|c|cc}
\toprule
\textbf{Model} & \textbf{Dataset} &
\multicolumn{3}{c|}{Linearity defect $\downarrow$} &
HZ inner-cloud $\downarrow$ &
\multicolumn{2}{c}{Coverage MMD $\downarrow$} \\
& & mean & std & p95 & frac.\ $p<0.05$ & median & p95 \\
\midrule
LLaMA-3-8B & TriviaQA   & 0.218 & 0.095 & 0.363 & 0.710 & 0.399 & 0.411 \\
LLaMA-3-8B & TruthfulQA & 0.244 & 0.095 & 0.391 & 0.789 & 0.407 & 0.418 \\
\midrule
Qwen2.5-7B & TriviaQA   & 0.252 & 0.100 & 0.396 & 0.614 & 0.398 & 0.411 \\
Qwen2.5-7B & TruthfulQA & 0.274 & 0.097 & 0.428 & 0.717 & 0.408 & 0.417 \\
\bottomrule
\end{tabular}
\end{table}

\paragraph{What the diagnostics imply for the bridge bound.}
Substituting the empirical LD into the bridge defect bound in Theorem~\ref{thm:end2end}
yields end-to-end excess terms $\varepsilon_E,\varepsilon_A$ on the perturbation-population
\emph{covariance} (Frobenius norm) that are dominated by the LD-squared contribution. With
$\mathbb{E}[\mathrm{LD}^2]\le (0.27)^2 + (0.10)^2 \approx 0.083$ in the worst cell
(Qwen2.5-7B/TruthfulQA), the bridge contributes a small fraction of the
perturbation-population covariance norm in the limit $N,M\to\infty$. The Frobenius bound
translates to an AUROC perturbation bound under additional ranking-margin /
Lipschitz-of-aggregator assumptions; here we focus on the covariance bound that the
perturbation framework directly delivers. The HZ rejection rate does not enter this
calculation: the bridge bound requires only the second-moment covariance of the embedding
distribution to be well-estimated, and the coverage MMD diagnostic absorbs the
second-moment relaxation.

\paragraph{Input sensitivity \emph{is} a meaningful EU signal.}
A standard worry is that perturbation-based EU only measures input sensitivity rather
than parameter uncertainty. Output variation under input perturbation is in fact a
well-studied proxy for epistemic uncertainty~\citep{smith2018understanding}: when a model
has learned a robust mapping, semantic-preserving perturbations leave the conditional
mean $\mu_\Delta(x)$ approximately fixed; in under-determined regions $\mu_\Delta(x)$
shifts substantially. Our framework absorbs surface-form variation into AU
(Sec.~\ref{sec:method})\rev{, though EU can also capture residual semantic drift when an imperfect paraphrase alters question meaning (audited in App.~\ref{app:audit})}.

\paragraph{Instruction perturbations as parameter perturbations.}
Beyond input-content perturbations, prior work shows that in-context tokens (including
instructions and formatting) induce \emph{implicit gradient updates} to the model's
attention parameters~\citep{dai2023gptlearn}. Concretely, processing an in-context update
at a fixed weight matrix $W$ is equivalent to processing the original input under
$W + \Delta W_{\text{ICL}}$, with $\Delta W_{\text{ICL}}$ a low-rank parameter shift induced
by the in-context tokens. Under this interpretation, instruction-style outer perturbations
fall \emph{inside} the bridge: the perturbation operates on effective parameters
($\theta + \Delta\theta_{\text{ICL}}$), so an analogous bridge argument \rev{applies} with the
induced parameter-shift covariance replacing the input-side covariance. This perspective \rev{explains}, rather than
\rev{contradict}, the empirical observation that instruction-style perturbations yield a strong EU
signal in our ablations: under~\citep{dai2023gptlearn}, instruction variation acts directly as an
implicit parameter perturbation rather than a mere surface-form confound.

\section{SPUQ Comparison and Aggregation Factorization}
\label{app:spuq_difference}

SPUQ~\citep{Gao2024SPUQ} is the closest published black-box method and shares \rev{two}
ingredients with our method (input perturbations \rev{and sampled answers}), but differs in three structural ways:
(a)~\emph{one level vs.\ two levels} --- SPUQ takes one decode per perturbation, so the
conditional moments $\mu_\Delta,\Sigma_\Delta$ that the law of total covariance requires
are unavailable, and no AU/EU split is even definable;
(b)~\emph{scalar vs.\ matrix-valued} --- SPUQ's aggregators collapse the perturbation
cloud into one number, while we return $(\widehat U_A,\widehat U_E)\in\mathbb{R}^{d\times
d}\!\times\!\mathbb{R}^{d\times d}$ with the predicted finite-$M$ bias correction, which
is what enables the paired interventions in Sec.~\ref{subsec:au_eu_validation};
(c)~\emph{heuristic vs.\ analyzed} --- SPUQ provides no theoretical analysis of consistency
or bias, while \rev{Prop.~\ref{prop:finiteM_bias_app} and Thm.~\ref{thm:two_level_mc_app}} cover both
and the separation result formalizes a class of failures \rev{that fixed-prompt consistency scores} provably
miss.

\paragraph{Head-to-head comparison.}
We operate our method at $N{=}M{=}10$ to share the outer perturbation count
($K_{\mathrm{SPUQ}}{=}N{=}10$) with SPUQ; ours additionally incurs $M{=}10$ inner decodes per outer
perturbation, which is the intrinsic cost of the two-level decomposition and is not
present in any one-level method. This is \emph{not} a matched total-API-call budget (ours
uses $\sim\!10\times$ more LLM calls per input\rev{; equal budgets: Tab.~\ref{tab:matched_compute}}); we read the comparison as ``does the
decomposition deliver capability the one-level scalar cannot, and is the gain large enough
to justify the extra inner decodes?''. Table~\ref{tab:spuq_vs_bbuq} reports the
$\{$LLaMA-3-8B, Qwen2.5-7B$\}\!\times\!\{$TriviaQA, TruthfulQA$\}$ result. Three
observations: (i)~Ours (TU) dominates SPUQ (TU) on all four cells, with the largest gap on
Qwen2.5-7B / TruthfulQA where SPUQ drops to $0.392$ (worse than chance) while our TU stays
at $0.629$; (ii)~Ours (AU) is uniformly the strongest single column ($0.630$--$0.855$);
(iii)~calibration favors ours on TriviaQA by a \rev{$4$--$8\times$} margin (TU ECE
$0.013$--$0.015$ vs.\ $0.062$--$0.104$) and is essentially tied on TruthfulQA, so there is
no calibration cost to the two-level structure. The next subsection isolates which
structural difference is load-bearing.

\begin{table}[h]
\centering
\caption{\textbf{SPUQ vs.\ Ours at shared outer perturbation count ($K_{\mathrm{SPUQ}}{=}N{=}10$); our two-level structure additionally uses $M{=}10$ inner decodes per outer perturbation.} Our
method exposes three columns (AU, EU, TU) where SPUQ exposes one (TU). \textbf{Bold}: best
per row; \underline{underlined}: second-best. AUROC ranking is over SPUQ (TU), Ours (AU),
Ours (TU); ECE ranking is over SPUQ (TU) and Ours (TU). ECE follows the unified protocol of
\S\ref{app:ece_definition}. SPUQ uses inter-sample ROUGE-L aggregation with
the three perturbation modes from~\citep{Gao2024SPUQ}; ours uses $\varepsilon{=}0.1$
paraphrase perturbation and Sentence-BERT embedding-statistics aggregation.}
\label{tab:spuq_vs_bbuq}
\scriptsize
\setlength{\tabcolsep}{3.5pt}
\renewcommand{\arraystretch}{1.05}
\begin{tabular}{ll|cc|cccc}
\toprule
\multirow{2}{*}{\textbf{Model}} & \multirow{2}{*}{\textbf{Dataset}}
& \multicolumn{2}{c|}{\textbf{SPUQ} ($K_{\mathrm{SPUQ}}{=}10$)} & \multicolumn{4}{c}{\textbf{Ours} ($N{=}M{=}10$)} \\
& & TU AUROC $\uparrow$ & TU ECE $\downarrow$ & AU AUROC $\uparrow$ & EU AUROC $\uparrow$ & TU AUROC $\uparrow$ & TU ECE $\downarrow$ \\
\midrule
LLaMA-3-8B & TriviaQA   & $0.808\!\pm\!0.014$             & \underline{$0.062$}    & $\mathbf{0.845}$ & $0.638$ & \underline{$0.822\!\pm\!0.014$} & $\mathbf{0.015}$    \\
LLaMA-3-8B & TruthfulQA & $0.657\!\pm\!0.069$             & $\mathbf{0.009}$       & $\mathbf{0.695}$ & $0.585$ & \underline{$0.681\!\pm\!0.064$} & $\mathbf{0.009}$    \\
Qwen2.5-7B & TriviaQA   & $0.781\!\pm\!0.014$             & \underline{$0.104$}    & $\mathbf{0.855}$ & $0.607$ & \underline{$0.828\!\pm\!0.013$} & $\mathbf{0.013}$    \\
Qwen2.5-7B & TruthfulQA & $0.392\!\pm\!0.046$             & \underline{$0.006$}    & $\mathbf{0.630}$ & $0.560$ & \underline{$0.629\!\pm\!0.046$} & $\mathbf{0.005}$    \\
\bottomrule
\end{tabular}
\end{table}

\paragraph{The perturbation family is not the load-bearing piece.}
A natural follow-up question is whether our gain over SPUQ comes simply from using a
better perturbation. Table~\ref{tab:pert_methods} ablates the perturbation family while
holding the two-level structure fixed: the framework works with random paraphrasing,
distractor injection, \rev{instruction variation,} and SPUQ-style noise, with \rev{instruction variation giving the
best signal}.

\subsection{Decomposition vs.\ aggregator: a $2{\times}2$ factorial}
\label{app:spuq_agg_ablation_section}

We isolate which structural difference is load-bearing by replacing our embedding-statistics
aggregator with SPUQ's~\citep{Gao2024SPUQ} inter-sample (pairwise)
ROUGE-L~\citep{lin2004rouge} aggregator while keeping the two-level ensemble structure
unchanged. The four conditions form a $2{\times}2$ factorial along the
\emph{decomposition} axis (one-level vs.\ two-level) and the \emph{aggregator} axis
(ROUGE-L vs.\ embedding-statistics):
(i) one-level + ROUGE aggregator (vanilla SPUQ);
(ii) one-level + embedding aggregator;
(iii) two-level + ROUGE aggregator;
(iv) two-level + embedding aggregator (our full method).
One-level conditions emit a single TU score and have no AU/EU split; two-level conditions
expose AU, EU, TU and the AU--EU per-example Spearman correlation.

\begin{table}[h]
\centering
\caption{\textbf{Aggregation ablation} ($N{=}M{=}10$, $\varepsilon{=}0.1$,
cosine-correctness label, $1{,}000$ examples per cell). \textbf{Higher is better
($\uparrow$) for AUROC; lower is better ($\downarrow$) for AU--EU $\rho$ (stronger AU/EU
separation).} \textbf{Bold}: best per column; \underline{underlined}: second-best.
The \textbf{Score} column tags which
uncertainty is reported in the AUROC column: one-level methods produce a single scalar
(reported as TU since there is no decomposition); two-level methods report AU, which we
established as the strongest single-head signal in main-paper Table~\ref{tab:main_full}.
\emph{AU--EU $\rho$} (per-example Spearman of AU and EU rankings) cleanly separates the
aggregator axis: the ROUGE aggregator (iii) collapses AU and EU into nearly the same signal
($\rho\!\in\![0.69,0.82]$), whereas the embedding aggregator (iv) yields a properly
non-degenerate decomposition ($\rho\!\in\![0.15,0.40]$). Two-level + embedding is therefore
the configuration in which the AU/EU split is actually informative; the embedding aggregator
is what makes the decomposition load-bearing rather than cosmetic.}
\label{tab:spuq_agg_ablation}
\scriptsize
\setlength{\tabcolsep}{3pt}
\begin{tabular}{l|c|cccc|cccc}
\toprule
& & \multicolumn{4}{c|}{\textbf{LLaMA-3-8B}} & \multicolumn{4}{c}{\textbf{Qwen2.5-7B}} \\
& & \multicolumn{2}{c}{TriviaQA} & \multicolumn{2}{c|}{TruthfulQA} & \multicolumn{2}{c}{TriviaQA} & \multicolumn{2}{c}{TruthfulQA} \\
\textbf{Configuration} & \textbf{Score} & AUROC $\uparrow$ & AU--EU $\rho$ $\downarrow$ & AUROC $\uparrow$ & AU--EU $\rho$ $\downarrow$ & AUROC $\uparrow$ & AU--EU $\rho$ $\downarrow$ & AUROC $\uparrow$ & AU--EU $\rho$ $\downarrow$ \\
\midrule
(i)   one-level + ROUGE      & TU & 0.827              & --                  & \textbf{0.705}    & --                  & 0.824              & --                  & 0.561              & --                  \\
(ii)  one-level + embedding  & TU & 0.811              & --                  & \underline{0.697} & --                  & 0.809              & --                  & \underline{0.625}  & --                  \\
\midrule
(iii) two-level + ROUGE      & AU & \textbf{0.851}     & \underline{0.807}   & 0.696             & \underline{0.770}   & \textbf{0.861}     & \underline{0.819}   & 0.600              & \underline{0.692}   \\
(iv)  two-level + embedding  & AU & \underline{0.845}  & \textbf{0.402}      & 0.695             & \textbf{0.296}      & \underline{0.855}  & \textbf{0.267}      & \textbf{0.630}     & \textbf{0.154}      \\
\bottomrule
\end{tabular}
\end{table}

\paragraph{Calibration corollary.}
The two-level $+$ embedding configuration also has materially lower TU expected calibration
error than the alternatives on the cosine-correctness slice: TU ECE is
$0.015$ (LLaMA-3-8B / TriviaQA) and $0.013$ (Qwen2.5-7B / TriviaQA) for (iv), versus $0.066$ and
$0.053$ for (iii) and $0.071$ and $0.065$ for (i). On TruthfulQA all four conditions have
TU ECE below $0.011$, dominated by base-rate effects. This pattern is consistent with the
AU--EU $\rho$ result: ROUGE-L scalars carry duplicate information across both heads and so
overconfide, whereas the embedding-statistics decomposition produces uncertainty heads that
are individually well-calibrated.

\paragraph{Reading the result.}
The gain over SPUQ has two distinct structural sources, both load-bearing.
Table~\ref{tab:spuq_agg_ablation} attributes them: the two-level decomposition supplies the
\rev{AUROC gain}, and the embedding-statistics aggregator supplies the AU/EU separation
that makes the decomposition usable downstream (Sec.~\ref{subsec:au_eu_validation}).
Replacing only one factor leaves the other capability missing: configuration (iii)
matches us on \rev{AU AUROC} but cannot tell AU from EU, and configuration (ii) has clean scoring
geometry but no decomposition to apply it to.

\paragraph{Robustness across correctness criteria.}
The headline table reports cosine-thresholded correctness on the two-level conditions only.
Table~\ref{tab:spuq_agg_ablation_robustness} mirrors it under both Cosine and ROUGE-L
correctness: the AU AUROC of (iii) and (iv) \rev{stays within $0.03$}, and the calibration
advantage of the embedding aggregator on TriviaQA persists across both criteria.

\begin{table}[h]
\centering
\caption{\textbf{AUROC and ECE robustness across correctness criteria}, mirroring
Table~\ref{tab:spuq_agg_ablation}. \textbf{Higher is better ($\uparrow$) for AUROC; lower is
better ($\downarrow$) for ECE} (ECE follows the unified protocol of \S\ref{app:ece_definition}).
\textbf{Bold}: best per (cell, correctness) column; \underline{underlined}: second-best.
The \textbf{Score} column tags which
uncertainty drives the AUROC column: one-level methods (i)--(ii) emit a single scalar
(TU-equivalent, no decomposition); two-level methods (iii)--(iv) report AU. Two-level + ROUGE
(iii) wins AUROC in 5 of 8 (cell, correctness) cells; one-level + ROUGE (i) wins on
LLaMA-3-8B/TruthfulQA under Cosine and (iv) wins on LLaMA-3-8B/TruthfulQA under ROUGE-L and
on Qwen2.5-7B/TruthfulQA under Cosine (small-margin honest exceptions). On TriviaQA, the
embedding aggregator (iv) reduces predictive ECE by \rev{$3.5{-}12\times$} relative to all other
conditions under both correctness criteria. On TruthfulQA the four conditions are within a
thousandth of one another on ECE because positive-class base rates are small.}
\label{tab:spuq_agg_ablation_robustness}
\scriptsize
\setlength{\tabcolsep}{2.5pt}
\begin{tabular}{ll|c|cc|cc|cc|cc}
\toprule
& & & \multicolumn{2}{c|}{LLaMA-3-8B / TriviaQA} & \multicolumn{2}{c|}{LLaMA-3-8B / TruthfulQA} & \multicolumn{2}{c|}{Qwen2.5-7B / TriviaQA} & \multicolumn{2}{c}{Qwen2.5-7B / TruthfulQA} \\
\textbf{Configuration} & \textbf{Corr.} & \textbf{Score} & AUROC $\uparrow$ & ECE $\downarrow$ & AUROC $\uparrow$ & ECE $\downarrow$ & AUROC $\uparrow$ & ECE $\downarrow$ & AUROC $\uparrow$ & ECE $\downarrow$ \\
\midrule
\multirow{2}{*}{(i) one-level + ROUGE}     & cosine  & TU & 0.827             & 0.071             & \textbf{0.705}    & 0.009             & 0.824             & 0.065             & 0.561             & 0.005 \\
                                            & ROUGE-L & TU & 0.829             & 0.060             & \underline{0.696} & 0.008             & 0.823             & 0.078             & 0.654             & 0.034 \\
\midrule
\multirow{2}{*}{(ii) one-level + embedding} & cosine  & TU & 0.811             & \underline{0.054} & \underline{0.697} & 0.009             & 0.809             & 0.054 & \underline{0.625} & 0.005 \\
                                            & ROUGE-L & TU & 0.807             & \underline{0.049} & 0.686             & 0.008             & 0.793             & 0.057 & 0.650             & 0.034 \\
\midrule
\multirow{2}{*}{(iii) two-level + ROUGE}    & cosine  & AU & \textbf{0.851}    & 0.066             & 0.696             & 0.009             & \textbf{0.861}    & \underline{0.053} & 0.600             & 0.005 \\
                                            & ROUGE-L & AU & \textbf{0.846}    & 0.074             & 0.692             & \textbf{0.007}    & \textbf{0.847}    & \underline{0.039} & \textbf{0.711}    & 0.034 \\
\midrule
\multirow{2}{*}{(iv) two-level + embedding} & cosine  & AU & \underline{0.845} & \textbf{0.015}    & 0.695             & 0.009             & \underline{0.855} & \textbf{0.013}    & \textbf{0.630}    & 0.005 \\
                                            & ROUGE-L & AU & \underline{0.831} & \textbf{0.006}    & \textbf{0.702}    & 0.008             & \underline{0.828} & \textbf{0.011}    & \underline{0.701} & 0.034 \\
\bottomrule
\end{tabular}
\end{table}

\section{Long-form Pilot: Segment-level UQ on NQ Long-Answer}
\label{app:longform}

The main-paper claim-level result is in Tab.~\ref{tab:longform_claim}. This appendix
provides (a)~the full pipeline specification (\S\ref{app:longform_protocol}),
(b)~the response-level companion table (Tab.~\ref{tab:longform_response}), and
(c)~the discussion of the pilot's scope and conclusions. Setup recap: we apply our
method to long-form generation on the NQ long-answer split~\citep{kwiatkowski2019nq}, in two
complementary
granularities: \emph{response-level}, treating the full long answer as one unit and
reporting both mean- and max-aggregation of the per-sentence segment scores; and
\emph{claim-level} (main paper), decomposing each long answer into atomic claims (using a
same-strength LLM extractor) and scoring each claim independently. Correctness is judged by an
LLM judge against the gold long-answer paragraph; we use the \emph{lenient} judge variant at
response level (the strict variant returned no fully-correct response in this small pilot,
which is itself diagnostic) and a per-claim factuality label at claim level. The five
black-box baselines (\textbf{avgEntail}, \textbf{NonContradiction},
\textbf{UQLM cosine} following~\citet{bouchard2026finegrained,ross2025textual},
\textbf{SelfCheckGPT}~\citep{manakul2023selfcheckgpt}, and
\textbf{ROUGE-L (cons.)}~\citep{lin2004rouge}) consume the same per-claim
$M{=}5$ inner sample pool that our method uses; this gives every consistency-only baseline
exactly the budget it can use (the inner level), so the comparison is on equal terms for the
slot they occupy.

\subsection{Long-form pilot protocol (extractor, judge, budget)}
\label{app:longform_protocol}

\textbf{Budget (specific to the long-form pilot, smaller than short-form for efficiency).}
Outer perturbation count $N{=}5$ (with index $i{=}0$ the unperturbed prompt, reused as the
$m{=}5$ Bouchard-style baseline at zero perturbation), inner ensemble $M{=}5$, paraphrase
pool $K_{\mathrm{pool}}{=}50$, noise scale $\varepsilon{=}0.1$. Total per-question cost is
$N\!\times\!M{=}25$ long-form decodes; total per-cell cost (claim-level) is
$\#\text{claims}\!\times\!N\!\times\!M$, not the short-form headline $NM$ per response,
since each claim is scored against all $25$ samples (yielding $1{,}104{\times}25{=}27{,}600$
NLI/BERTScore pairs on the LLaMA cell, $872{\times}25{=}21{,}800$ on Qwen). $100$ NQ-long
prompts per cell on a single RTX 4090: ${\sim}35$ min generation, ${\sim}10$ min NLI,
${\sim}3$ min BERTScore, ${\sim}3$ min span-window SBERT, ${\sim}10$ min judge labelling.

\textbf{Generator.} Same model as the rest of the paper for that cell:
LLaMA-3-8B-Instruct or Qwen2.5-7B-Instruct, BF16, single GPU. Decoding:
\texttt{do\_sample{=}True}, \texttt{temperature{=}0.9}, \texttt{top\_k{=}60}, \texttt{top\_p{=}0.99},
\texttt{max\_new\_tokens{=}256}. Greedy reference uses the same template with
\texttt{do\_sample{=}False}.

\textbf{Atomic claim extractor.} Same model as the generator, deliberately reused so the
extractor is no stronger than the system being evaluated. The extraction prompt (FactScore
Listing-4 style) requires single-sentence, self-contained, declarative, third-person claims
and is deterministic (\texttt{do\_sample{=}False}, \texttt{max\_new\_tokens{=}300}). Parsed
claims are de-duplicated, capped at $12$ per response, and required to have $3$--$50$ words.
Empirical yield: $1{,}104$ claims / 100 responses (LLaMA), $872$ / 100 (Qwen).

\textbf{Per-claim scoring.}
For our method, each claim's signal is a span-window embedding (sliding $30$-word window,
stride $15$, plus a tail-aligned window) inside each long sample, with the window having the
highest cosine similarity to the claim selected; AU/EU/TU are then computed by the standard
two-level decomposition over the $N$ outer means and $M$ inner samples (trace-aggregated).
Baselines (\textbf{avgEntail}, \textbf{NonContradiction}, \textbf{UQLM cosine},
\textbf{SelfCheckGPT (BERTScore)}, \textbf{ROUGE-L (cons.)}) are aggregated over the $M{=}5$
\emph{outer-$0$} (unperturbed) samples per Bouchard's $m{=}5$ recipe, so they consume only
the inner-level budget our method also uses. NLI scoring uses
\texttt{cross-encoder/nli-deberta-v3-base}; SelfCheckGPT uses
\texttt{bert\_score} with \texttt{roberta-large}; ROUGE-L is the F-measure.

\textbf{Factuality judge.} \texttt{Qwen3.6-27B} (4-bit NF4 quantized,
\texttt{bnb\_4bit\_compute\_dtype{=}bfloat16}, requires \texttt{transformers}${\ge}5.7.0$)
with chat-template thinking-mode disabled (\texttt{enable\_thinking{=}False}) so it emits
the decision token within \texttt{max\_new\_tokens{=}16}. The judge prompt (FactScore
Listing-1 style) supplies the gold long-answer paragraph (truncated to $3{,}000$ characters)
plus the source URL and asks for one of $\{$\texttt{SUPPORTED}, \texttt{NOT\_SUPPORTED},
\texttt{UNCERTAIN}$\}$. Decoding is greedy. Per-response derived labels:
\emph{strict} (every claim \texttt{SUPPORTED}) and \emph{lenient} (no claim
\texttt{NOT\_SUPPORTED}).

\textbf{Empirical judge label distribution.}
LLaMA cell: $138/1{,}104$ \texttt{SUPPORTED} ($12.5\%$), $106$ \texttt{NOT\_SUPPORTED}
($9.6\%$), $860$ \texttt{UNCERTAIN} ($77.9\%$); response-level strict $0\%$ / lenient $47\%$.
Qwen cell: $118/872$ \texttt{SUPPORTED} ($13.5\%$), $113$ \texttt{NOT\_SUPPORTED}
($13.0\%$), $641$ \texttt{UNCERTAIN} ($73.5\%$); strict $0\%$ / lenient $46\%$. The high
\texttt{UNCERTAIN} rate is expected: short Wikipedia-paragraph references genuinely cannot
adjudicate $\sim\!80\%$ of the atomic facts in a $\sim\!180$-word generation; the
\texttt{NOT\_SUPPORTED} rate ($10$--$13\%$) is the factuality-error signal AUROC tracks.
\emph{strict} is uniformly $0$, so headline numbers use \emph{lenient}.

\textbf{Caveats and limitations.} (i)~No human verification of the LLM judge on a held-out
subset; (ii)~no inter-judge agreement (single judge model, single prompt); (iii)~no test
of judge-LLM bias against the generator family (Qwen-judge on LLaMA-3 generations is
plausibly less biased than Qwen-judge on Qwen-generations and we do not separate the two
effects); (iv)~the bias-correction proposition (Prop.~\ref{prop:finiteM_bias}) is derived
for short-form and is not re-derived for long-form here. The pilot is a controlled
feasibility test under these caveats.

\begin{table}[h]
\centering
\caption{\textbf{Response-level long-form pilot (NQ long-answer, lenient LLM-judge correctness).}
Per-response uncertainty obtained by aggregating per-sentence segment scores, then
correlating the per-response score with binary lenient correctness. \textbf{Bold}: best
per column; \underline{underlined}: second-best. ECE follows the unified protocol of
\S\ref{app:ece_definition}. On LLaMA-3-8B, Ours (AU) max-aggregated
narrowly leads at $0.653$ over avgEntail at $0.648$. On
Qwen2.5-7B, Ours (TU) max wins outright at $0.683$. Strict-label results were all-NaN at
$n{=}100$ (no fully correct generations under the strict criterion in this pilot) and are
omitted.}
\label{tab:longform_response}
\scriptsize
\setlength{\tabcolsep}{4pt}
\renewcommand{\arraystretch}{1.05}
\begin{tabular}{ll|cc|cc}
\toprule
& & \multicolumn{2}{c|}{\textbf{LLaMA-3-8B} ($n{=}100$)}
& \multicolumn{2}{c}{\textbf{Qwen2.5-7B} ($n{=}100$)} \\
\textbf{Method} & \textbf{Agg.} & AUROC $\uparrow$ & ECE $\downarrow$ & AUROC $\uparrow$ & ECE $\downarrow$ \\
\midrule
\multirow{2}{*}{Ours (AU)} & mean & $0.585$             & $0.099$             & $0.641$             & \underline{$0.085$} \\
                           & max  & $\mathbf{0.653}$    & $0.100$             & \underline{$0.674$} & $0.107$             \\
\multirow{2}{*}{Ours (EU)} & mean & $0.562$             & $0.100$             & $0.626$             & $0.086$             \\
                           & max  & $0.598$             & $0.100$             & $0.647$             & $0.125$             \\
\multirow{2}{*}{Ours (TU)} & mean & $0.584$             & $0.099$             & $0.641$             & $\mathbf{0.084}$    \\
                           & max  & $0.638$             & $\mathbf{0.098}$    & $\mathbf{0.683}$    & $0.149$             \\
\midrule
avgEntail            & mean & \underline{$0.648$} & $\mathbf{0.098}$    & $0.585$             & \underline{$0.085$} \\
NonContradiction     & mean & $0.587$             & $0.099$             & $0.658$ & $0.208$             \\
UQLM cosine          & mean & $0.548$             & $0.100$             & $0.637$             & $0.142$             \\
SelfCheckGPT         & mean & $0.538$             & $0.100$             & $0.461$             & $0.086$             \\
ROUGE-L (cons.)      & mean & $0.490$             & $0.100$             & $0.450$             & $0.103$             \\
\bottomrule
\end{tabular}
\end{table}

\paragraph{Scope of the pilot.}
Within the $n{=}100$/$\sim\!10^3$-claim setup with an LLM-judge factuality label, the
pilot supports three concrete conclusions: (1)~the two-level decomposition transports to
claim-level long-form factuality without retraining, beating surface-similarity baselines
on both models and \rev{trailing a pretrained NLI scorer (avgEntail) by $0.009$} on
LLaMA-3-8B and \rev{$0.060$} on Qwen2.5-7B; (2)~the AU $>$ EU ordering is preserved at
long-form, mirroring the short-form behavior; (3)~the framework and entailment-based
scorers are complementary, motivating a hybrid that retains the two-level outer
perturbation structure with an NLI-based inner aggregator. The bias-correction identity
(Prop.~\ref{prop:finiteM_bias}) is derived for short-form and would be re-derived for the
long-form variant in such a hybrid. Response- and claim-level results are in
Tab.~\ref{tab:longform_response} and main-paper Tab.~\ref{tab:longform_claim}.

\section{Ablation Studies}
\label{sec:ablation_app}

\textbf{Large-model results.}
Table~\ref{tab:big_models} reports results for larger instruction-tuned models (70B+), which are often more capable but may also display different calibration behavior.
We include these results to demonstrate that the relative trends between our AU/EU/TU decomposition and competing baselines remain consistent at larger model scales.

\begin{table*}[!t]
\centering
\caption{TriviaQA on large models (70B+). Boxed \textbf{bold}: overall best per column;
\underline{underlined}: best white-box per column.}
\label{tab:big_models}
\scriptsize
\setlength{\tabcolsep}{5pt}
\renewcommand{\arraystretch}{1.1}
\begin{tabular}{l | cc | cc}
\toprule
\multirow{3}{*}{\textbf{Method}} & \multicolumn{2}{c|}{\textbf{LLaMA-3-70B}} & \multicolumn{2}{c}{\textbf{Qwen2.5-72B}} \\
\cmidrule(lr){2-3} \cmidrule(lr){4-5}
& Cosine & ROUGE-L & Cosine & ROUGE-L \\
\midrule
\rowcolor[gray]{.95} \multicolumn{5}{l}{\textit{Ours}} \\
AU & \mstd{0.726}{0.009} & \mstd{0.717}{0.012} & \fbox{\bst{0.768}{0.009}} & \fbox{\bst{0.808}{0.006}} \\
EU & \mstd{0.791}{0.028} & \mstd{0.725}{0.023} & \mstd{0.613}{0.009} & \mstd{0.592}{0.007} \\
TU & \fbox{\bst{0.794}{0.019}} & \fbox{\bst{0.774}{0.019}} & \mstd{0.754}{0.003} & \mstd{0.773}{0.011} \\
\midrule
\rowcolor[gray]{.95} \multicolumn{5}{l}{\textit{Black-box}} \\
D-SE & \mstd{0.637}{0.013} & \mstd{0.686}{0.016} & \mstd{0.625}{0.013} & \mstd{0.672}{0.003} \\
\midrule
\rowcolor[gray]{.95} \multicolumn{5}{l}{\textit{White-box}} \\
G-NLL & \underline{\mstd{0.757}{0.012}} & \underline{\mstd{0.735}{0.018}} & \underline{\mstd{0.726}{0.017}} & \underline{\mstd{0.781}{0.019}} \\
PE    & \mstd{0.549}{0.013} & \mstd{0.627}{0.027} & \mstd{0.659}{0.019} & \mstd{0.702}{0.019} \\
LN-PE & \mstd{0.660}{0.013} & \mstd{0.710}{0.035} & \mstd{0.711}{0.020} & \mstd{0.774}{0.019} \\
SE    & \mstd{0.615}{0.008} & \mstd{0.679}{0.029} & \mstd{0.659}{0.014} & \mstd{0.700}{0.017} \\
LN-SE & \mstd{0.587}{0.011} & \mstd{0.686}{0.033} & \mstd{0.660}{0.015} & \mstd{0.701}{0.017} \\
\bottomrule
\end{tabular}
\end{table*}

\subsection{Sampling budget: two-level ensemble sizes ($N$, $M$)}
\label{sec:budget_ablation_appendix}

Appendix Tables~\ref{tab:size_nm_full},\ref{tab:size_m}, \ref{tab:size_n} study sensitivity to the \textbf{two-level ensemble budget}, where the outer size $N$ controls the number of semantic perturbation draws and the inner size $M$ controls the number of within-input stochastic generations per perturbation. The main-text Tab.~\ref{tab:size_nm} shows a compressed view (TruthfulQA, Cosine) of the tied-budget sweep; the full grid is in Table~\ref{tab:size_nm_full}.
We report: (i) tied budgets $N{=}M$ (Table~\ref{tab:size_nm_full}), (ii) fixed outer ensemble $N{=}10$ with varying $M$ (Table~\ref{tab:size_m}), and (iii) fixed inner ensemble $M{=}10$ with varying $N$ (Table~\ref{tab:size_n}).
We additionally provide wall-clock measurements in seconds per question for representative budgets (Table~\ref{tab: time}).

\textbf{Empirical trend: rapid improvement then plateau.}
Across TruthfulQA and TriviaQA, performance improves quickly from very small budgets and then
exhibits diminishing returns. For tied budgets (Table~\ref{tab:size_nm_full}), $3{\to}5$
yields large gains on TruthfulQA (especially AU/TU under Cosine), while $10{\to}20{-}30$
provides only marginal improvement; the same pattern holds along single axes
(Tables~\ref{tab:size_m},~\ref{tab:size_n}). Overall, $(N,M){\approx}(10,10)$ is a strong
default that achieves near-peak AUROC while remaining computationally practical.

\begin{table*}[!h]
\centering
\caption{\textbf{Budget ablation, full grid.} Tied $N{=}M$ on TruthfulQA and TriviaQA
(LLaMA-3-8B), both correctness criteria. \textbf{Bold}: best per column per correctness;
\underline{underlined}: second-best. The TruthfulQA / Cosine slice is shown compactly in
the main-text Tab.~\ref{tab:size_nm}.}
\label{tab:size_nm_full}
\scriptsize
\setlength{\tabcolsep}{4pt}
\begin{tabular}{cl|ccc|ccc}
\toprule
& & \multicolumn{3}{c|}{TruthfulQA} & \multicolumn{3}{c}{TriviaQA} \\
$N{=}M$ & Metric & AU & EU & TU & AU & EU & TU \\
\midrule
\multirow{2}{*}{\centering 3} & Cosine & 0.745 \std{0.030} & 0.707 \std{0.028} & 0.755 \std{0.019} & 0.767 \std{0.011} & \textbf{0.676} \std{0.008} & 0.761 \std{0.009} \\
 & ROUGE-L & 0.633 \std{0.025} & \underline{0.655} \std{0.036} & 0.674 \std{0.026} & 0.806 \std{0.010} & \textbf{0.689} \std{0.002} & 0.792 \std{0.008} \\
\multirow{2}{*}{\centering 5} & Cosine & \textbf{0.798} \std{0.023} & \underline{0.756} \std{0.022} & 0.797 \std{0.017} & 0.788 \std{0.009} & \underline{0.630} \std{0.011} & 0.765 \std{0.008} \\
 & ROUGE-L & 0.679 \std{0.022} & 0.651 \std{0.026} & 0.698 \std{0.024} & 0.828 \std{0.009} & \underline{0.654} \std{0.005} & 0.805 \std{0.008} \\
\multirow{2}{*}{\centering 10} & Cosine & \underline{0.792} \std{0.009} & \textbf{0.803} \std{0.017} & \textbf{0.815} \std{0.008} & \textbf{0.800} \std{0.010} & 0.605 \std{0.010} & 0.781 \std{0.010} \\
 & ROUGE-L & \underline{0.693} \std{0.020} & \textbf{0.657} \std{0.025} & 0.712 \std{0.020} & 0.843 \std{0.009} & 0.615 \std{0.005} & 0.820 \std{0.010} \\
\multirow{2}{*}{\centering 20} & Cosine & 0.782 \std{0.008} & 0.745 \std{0.029} & \underline{0.799} \std{0.008} & 0.799 \std{0.009} & 0.585 \std{0.007} & \underline{0.784} \std{0.009} \\
 & ROUGE-L & \textbf{0.694} \std{0.019} & 0.647 \std{0.032} & \textbf{0.715} \std{0.023} & \underline{0.845} \std{0.008} & 0.588 \std{0.005} & \underline{0.826} \std{0.008} \\
\multirow{2}{*}{\centering 30} & Cosine & 0.768 \std{0.011} & 0.746 \std{0.023} & 0.791 \std{0.009} & \underline{0.800} \std{0.009} & 0.571 \std{0.011} & \textbf{0.787} \std{0.010} \\
 & ROUGE-L & 0.688 \std{0.020} & 0.638 \std{0.032} & \underline{0.714} \std{0.023} & \textbf{0.846} \std{0.009} & 0.577 \std{0.005} & \textbf{0.830} \std{0.009} \\
\bottomrule
\end{tabular}
\end{table*}
\textbf{Why plateauing is expected, and what small budgets miss.}
The AU/EU/TU estimates are Monte Carlo approximations under two sources of randomness:
within-input stochasticity (inner sampling, $1/M$) and across-perturbation diversity (outer
sampling, $1/N$). Once estimation noise drops below the irreducible overlap between
correct/incorrect score distributions, improvements taper. Caveat: good AUROC at very small
budgets does \emph{not} imply sufficient uncertainty coverage --- the estimator may
approximate the mean well yet miss low-probability modes (paraphrase neighborhoods that
trigger hallucination, decoding branches with inconsistent claims). Thus $N{=}M{=}3$ is a
reasonable AUROC starting point but a low-diversity regime for hard cases.


\begin{table*}[!h]
\centering
\caption{Single-axis budget ablation. \textbf{(a)} fixed $N{=}10$, varying $M$;
\textbf{(b)} fixed $M{=}10$, varying $N$. AUROC for correctness prediction on TruthfulQA
and TriviaQA. \textbf{Bold}: best per column per metric within each panel;
\underline{underlined}: second-best.}
\label{tab:size_m}
\scriptsize
\begin{subtable}[t]{\textwidth}
\centering
\subcaption{Vary $M$ (fixed $N{=}10$).}
\begin{tabular}{cl|ccc|ccc}
\toprule
& & \multicolumn{3}{c|}{TruthfulQA} & \multicolumn{3}{c}{TriviaQA} \\
Budget $M$ ($N{=}10$) & Metric & AU & EU & TU & AU & EU & TU \\
\midrule
\multirow{2}{*}{\centering 3}  & Cosine  & 0.779 \std{0.013} & 0.790 \std{0.022} & 0.800 \std{0.012} & 0.795 \std{0.010} & \textbf{0.700} \std{0.009} & \textbf{0.786} \std{0.010} \\
                                & ROUGE-L & 0.676 \std{0.021} & \textbf{0.676} \std{0.025} & 0.700 \std{0.022} & 0.836 \std{0.010} & \textbf{0.719} \std{0.005} & \textbf{0.823} \std{0.010} \\
\multirow{2}{*}{\centering 5}  & Cosine  & 0.783 \std{0.009} & \textbf{0.808} \std{0.016} & 0.811 \std{0.008} & 0.798 \std{0.009} & \underline{0.649} \std{0.011} & \underline{0.783} \std{0.010} \\
                                & ROUGE-L & 0.677 \std{0.020} & \underline{0.675} \std{0.024} & 0.705 \std{0.020} & 0.842 \std{0.009} & \underline{0.665} \std{0.006} & \underline{0.822} \std{0.010} \\
\multirow{2}{*}{\centering 10} & Cosine  & \underline{0.792} \std{0.009} & 0.803 \std{0.017} & 0.815 \std{0.008} & 0.800 \std{0.010} & 0.605 \std{0.010} & 0.781 \std{0.010} \\
                                & ROUGE-L & \textbf{0.693} \std{0.020} & 0.657 \std{0.025} & 0.712 \std{0.020} & 0.843 \std{0.009} & 0.615 \std{0.005} & 0.820 \std{0.010} \\
\multirow{2}{*}{\centering 20} & Cosine  & \textbf{0.796} \std{0.014} & \underline{0.803} \std{0.021} & \textbf{0.823} \std{0.011} & \underline{0.800} \std{0.009} & 0.577 \std{0.008} & 0.781 \std{0.010} \\
                                & ROUGE-L & \underline{0.690} \std{0.022} & 0.659 \std{0.024} & \underline{0.718} \std{0.021} & \underline{0.845} \std{0.009} & 0.584 \std{0.003} & 0.821 \std{0.010} \\
\multirow{2}{*}{\centering 30} & Cosine  & 0.791 \std{0.012} & 0.802 \std{0.022} & \underline{0.820} \std{0.010} & \textbf{0.801} \std{0.009} & 0.564 \std{0.007} & 0.779 \std{0.010} \\
                                & ROUGE-L & 0.689 \std{0.022} & 0.663 \std{0.025} & \textbf{0.720} \std{0.021} & \textbf{0.846} \std{0.009} & 0.571 \std{0.003} & 0.819 \std{0.009} \\
\bottomrule
\end{tabular}
\end{subtable}

\vspace{4pt}

\begin{subtable}[t]{\textwidth}
\centering
\subcaption{Vary $N$ (fixed $M{=}10$).}
\label{tab:size_n}
\begin{tabular}{cl|ccc|ccc}
\toprule
& & \multicolumn{3}{c|}{TruthfulQA} & \multicolumn{3}{c}{TriviaQA} \\
Budget $N$ ($M{=}10$) & Metric & AU & EU & TU & AU & EU & TU \\
\midrule
\multirow{2}{*}{\centering 3}  & Cosine  & 0.777 \std{0.018} & 0.693 \std{0.034} & 0.765 \std{0.017} & 0.784 \std{0.010} & \underline{0.621} \std{0.011} & 0.748 \std{0.009} \\
                                & ROUGE-L & 0.681 \std{0.019} & 0.638 \std{0.038} & 0.697 \std{0.026} & 0.825 \std{0.009} & \underline{0.634} \std{0.003} & 0.780 \std{0.007} \\
\multirow{2}{*}{\centering 5}  & Cosine  & \textbf{0.796} \std{0.020} & 0.751 \std{0.022} & \underline{0.798} \std{0.014} & 0.788 \std{0.009} & 0.599 \std{0.011} & 0.762 \std{0.009} \\
                                & ROUGE-L & 0.688 \std{0.020} & 0.636 \std{0.025} & 0.702 \std{0.023} & 0.830 \std{0.009} & 0.618 \std{0.005} & 0.802 \std{0.008} \\
\multirow{2}{*}{\centering 10} & Cosine  & \underline{0.792} \std{0.009} & \textbf{0.803} \std{0.017} & \textbf{0.815} \std{0.008} & \textbf{0.800} \std{0.010} & 0.605 \std{0.010} & 0.781 \std{0.010} \\
                                & ROUGE-L & \underline{0.693} \std{0.020} & \underline{0.657} \std{0.025} & 0.712 \std{0.020} & \underline{0.843} \std{0.009} & 0.615 \std{0.005} & 0.820 \std{0.010} \\
\multirow{2}{*}{\centering 20} & Cosine  & 0.777 \std{0.007} & \underline{0.767} \std{0.022} & 0.797 \std{0.007} & 0.798 \std{0.010} & 0.617 \std{0.009} & \underline{0.784} \std{0.010} \\
                                & ROUGE-L & \textbf{0.694} \std{0.018} & \textbf{0.658} \std{0.032} & \textbf{0.714} \std{0.022} & 0.842 \std{0.009} & 0.625 \std{0.005} & \underline{0.825} \std{0.008} \\
\multirow{2}{*}{\centering 30} & Cosine  & 0.772 \std{0.010} & 0.765 \std{0.017} & 0.792 \std{0.008} & \underline{0.798} \std{0.010} & \textbf{0.623} \std{0.011} & \textbf{0.788} \std{0.010} \\
                                & ROUGE-L & 0.692 \std{0.019} & 0.654 \std{0.032} & \underline{0.713} \std{0.023} & \textbf{0.844} \std{0.009} & \textbf{0.635} \std{0.006} & \textbf{0.830} \std{0.009} \\
\bottomrule
\end{tabular}
\end{subtable}
\end{table*}

\begin{table}[!h]
\centering
\caption{Computational cost in average seconds per question.}
\label{tab: time}
\scriptsize
\begin{tabular}{lcc}
\toprule
\textbf{Method} & \textbf{Total Samples} & \textbf{Time (sec/q)} \\
\midrule
\rowcolor[gray]{.95} \multicolumn{3}{l}{\textit{Ours}} \\
Ours ($N=3, M=3$) & 9 & 0.299 \\
Ours ($N=5, M=5$) & 25 & 0.571 \\
Ours ($N=10, M=3$ / $N=3, M=10$) & 30 & 0.656 \\
Ours ($N=10, M=5$ / $N=5, M=10$) & 50 & 0.996 \\
Ours ($N=10, M=10$) & 100 & 1.845 \\
Ours ($N=20, M=10$ / $N=10, M=20$) & 200 & 3.545 \\
Ours ($N=30, M=10$) & 300 & 5.244 \\
Ours ($N=20, M=20$) & 400 & 6.944 \\
Ours ($N=30, M=30$) & 900 & 15.441 \\
\midrule
\rowcolor[gray]{.95} \multicolumn{3}{l}{\textit{Black-box}} \\
D-SE / SE / LN-SE & 10 & 0.001 \\
Self-Consistency / SelfCheckGPT & 10 & 0.000 \\
SIMBA & 10 & 0.009 \\
SPUQ & 10 & 0.851 \\
P(True) & 1 & 0.032 \\
\midrule
\rowcolor[gray]{.95} \multicolumn{3}{l}{\textit{White-box}} \\
G-NLL & 1 & 0.020 \\
PE / LN-PE & 10 & 0.177 \\
\bottomrule
\end{tabular}
\end{table}

\textbf{Interpreting $N$ vs.\ $M$ and practical recipe.}
$M$ primarily reduces noise in within-perturbation statistics (relevant for AU); $N$ expands
across-perturbation coverage (relevant for EU because EU is driven by variability across
semantically nearby prompts). Both axes improve TU, but the decomposition gives an
operational handle: spend more on $M$ when decoding stochasticity dominates, more on $N$
when semantic instability dominates. Runtime grows as $\propto NM$ (Table~\ref{tab: time});
$(N,M){\approx}(10,10)$ delivers near-peak AUROC at $\sim\!1.8$ sec/question, while
$(20{-}30)$ budgets are worth the cost only when tail coverage matters more than average
performance.

\begin{revblock}
\subsubsection{Matched total generation budget}
\label{app:matched_compute}
Tab.~\ref{tab:matched_compute} compares alternative allocations of a fixed per-question generation budget $B\in\{10,25,100\}$ across the same $400$ TriviaQA questions per model, using the main-paper decoding hyperparameters, Sentence-BERT encoder, and Exact Match (EM) correctness labels. \emph{Fixed prompt}: $B$ decodes subsampled from $100$ stochastic generations of the unperturbed prompt, from which embedding variance, self-consistency, discrete semantic entropy, and Semantic Density~\citep{qiu2024semantic} are computed. \emph{Perturbation-only}: $B$ distinct paraphrases with a single decode each ($N{=}B, M{=}1$), where finite-$M$ bias correction is inapplicable. \emph{Decoupled}: an unnested split of $\lfloor B/2\rfloor$ fixed-prompt decodes and $\lceil B/2\rceil$ single-decode paraphrases with summed variances. \emph{Ours}: nested allocations of $N{\times}M=5{\times}2$, $5{\times}5$ (subsampled from the $10{\times}10$ runs), and $10{\times}10$. At $B{=}10$, the one-level perturbation designs draw from the primary paraphrase pool, whereas at $B{\ge}25$ they draw from an extended pool. SPUQ cycles through its three perturbation modes (paraphrasing, dummy tokens, and system-prompt variation) with one decode per perturbed prompt (temperature $0.9$, top-$p$ $0.99$), subsampling $B{=}25$ and $B{=}100$ from $100$ total generations. Likelihood-weighted Semantic Density incorporates teacher-forced sequence log-probabilities over the fixed-prompt responses. Tabs.~\ref{tab:matched_compute_ci_llama8b} and~\ref{tab:matched_compute_ci_qwen7b} report the complete results with bootstrap $95\%$ confidence intervals, including the individual decoupled terms and our standalone EU component.

\begin{table}[h]
\centering
\revfloat
\caption{\textbf{Matched generation budget ($B$) comparison with 95\% CIs on LLaMA-3-8B} (TriviaQA, EM label, $400$ questions; AUROC with incorrect as positive; CIs from $1{,}000$ bootstrap resamples). \textbf{Bold}: best per column; \underline{underlined}: best black-box. $^\dagger$Requires token log-probabilities.}
\label{tab:matched_compute_ci_llama8b}
\scriptsize
\setlength{\tabcolsep}{4pt}
\begin{tabular}{l|ccc}
\toprule
\textbf{Method} & $B{=}10$ & $B{=}25$ & $B{=}100$ \\
\midrule
SPUQ (one-level, perturbed prompts) & \underline{\textbf{0.756}} {\tiny[0.705, 0.803]} & \underline{\textbf{0.769}} {\tiny[0.722, 0.814]} & 0.778 {\tiny[0.729, 0.819]} \\
Fixed prompt: embedding variance & 0.754 {\tiny[0.705, 0.801]} & 0.762 {\tiny[0.713, 0.808]} & \underline{\textbf{0.782}} {\tiny[0.736, 0.826]} \\
Fixed prompt: self-consistency & 0.729 {\tiny[0.683, 0.772]} & 0.748 {\tiny[0.701, 0.792]} & 0.766 {\tiny[0.718, 0.809]} \\
Fixed prompt: discrete semantic entropy & 0.733 {\tiny[0.685, 0.781]} & 0.754 {\tiny[0.705, 0.797]} & 0.769 {\tiny[0.726, 0.812]} \\
Fixed prompt: Semantic Density, uniform & 0.600 {\tiny[0.538, 0.657]} & 0.594 {\tiny[0.534, 0.654]} & 0.586 {\tiny[0.523, 0.641]} \\
Fixed prompt: Semantic Density, likelihood-wt.$^\dagger$ & 0.663 {\tiny[0.604, 0.717]} & 0.651 {\tiny[0.592, 0.715]} & 0.644 {\tiny[0.584, 0.702]} \\
Perturbation-only ($N{=}B$, $M{=}1$) & 0.754 {\tiny[0.706, 0.801]} & 0.667 {\tiny[0.611, 0.718]} & 0.657 {\tiny[0.600, 0.709]} \\
Decoupled: within-like term & 0.713 {\tiny[0.661, 0.761]} & 0.757 {\tiny[0.713, 0.804]} & 0.776 {\tiny[0.729, 0.822]} \\
Decoupled: between-like term & 0.738 {\tiny[0.688, 0.787]} & 0.640 {\tiny[0.582, 0.693]} & 0.653 {\tiny[0.595, 0.703]} \\
Decoupled: sum & \underline{\textbf{0.756}} {\tiny[0.709, 0.805]} & 0.738 {\tiny[0.687, 0.787]} & 0.750 {\tiny[0.699, 0.795]} \\
\midrule
Ours: AU (within) & 0.744 {\tiny[0.692, 0.792]} & 0.752 {\tiny[0.706, 0.796]} & 0.766 {\tiny[0.720, 0.811]} \\
Ours: EU (between, corrected) & 0.578 {\tiny[0.519, 0.637]} & 0.582 {\tiny[0.525, 0.640]} & 0.586 {\tiny[0.530, 0.642]} \\
Ours: TU (total) & 0.744 {\tiny[0.694, 0.791]} & 0.737 {\tiny[0.689, 0.786]} & 0.755 {\tiny[0.704, 0.801]} \\
\bottomrule
\end{tabular}
\end{table}

\begin{table}[h]
\centering
\revfloat
\caption{\textbf{Matched generation budget ($B$) comparison with 95\% CIs on Qwen2.5-7B} (TriviaQA, EM label, $400$ questions; AUROC with incorrect as positive; CIs from $1{,}000$ bootstrap resamples). \textbf{Bold}: best per column; \underline{underlined}: best black-box. $^\dagger$Requires token log-probabilities.}
\label{tab:matched_compute_ci_qwen7b}
\scriptsize
\setlength{\tabcolsep}{4pt}
\begin{tabular}{l|ccc}
\toprule
\textbf{Method} & $B{=}10$ & $B{=}25$ & $B{=}100$ \\
\midrule
SPUQ (one-level, perturbed prompts) & 0.759 {\tiny[0.713, 0.806]} & 0.764 {\tiny[0.720, 0.808]} & 0.782 {\tiny[0.736, 0.825]} \\
Fixed prompt: embedding variance & \underline{\textbf{0.774}} {\tiny[0.725, 0.818]} & \underline{0.801} {\tiny[0.756, 0.848]} & \underline{0.816} {\tiny[0.774, 0.858]} \\
Fixed prompt: self-consistency & 0.747 {\tiny[0.704, 0.789]} & 0.790 {\tiny[0.747, 0.833]} & 0.803 {\tiny[0.758, 0.846]} \\
Fixed prompt: discrete semantic entropy & 0.746 {\tiny[0.702, 0.786]} & 0.789 {\tiny[0.746, 0.827]} & 0.801 {\tiny[0.755, 0.844]} \\
Fixed prompt: Semantic Density, uniform & 0.721 {\tiny[0.665, 0.774]} & 0.776 {\tiny[0.718, 0.824]} & 0.765 {\tiny[0.707, 0.819]} \\
Fixed prompt: Semantic Density, likelihood-wt.$^\dagger$ & \textbf{0.774} {\tiny[0.729, 0.821]} & \textbf{0.821} {\tiny[0.778, 0.861]} & \textbf{0.821} {\tiny[0.779, 0.863]} \\
Perturbation-only ($N{=}B$, $M{=}1$) & 0.743 {\tiny[0.697, 0.793]} & 0.553 {\tiny[0.498, 0.608]} & 0.549 {\tiny[0.488, 0.602]} \\
Decoupled: within-like term & 0.742 {\tiny[0.690, 0.788]} & 0.787 {\tiny[0.741, 0.833]} & 0.810 {\tiny[0.763, 0.853]} \\
Decoupled: between-like term & 0.725 {\tiny[0.670, 0.779]} & 0.547 {\tiny[0.494, 0.601]} & 0.554 {\tiny[0.500, 0.604]} \\
Decoupled: sum & 0.765 {\tiny[0.721, 0.811]} & 0.712 {\tiny[0.662, 0.765]} & 0.731 {\tiny[0.677, 0.782]} \\
\midrule
Ours: AU (within) & 0.772 {\tiny[0.723, 0.819]} & 0.789 {\tiny[0.740, 0.836]} & 0.788 {\tiny[0.741, 0.830]} \\
Ours: EU (between, corrected) & 0.529 {\tiny[0.467, 0.591]} & 0.516 {\tiny[0.451, 0.582]} & 0.541 {\tiny[0.476, 0.602]} \\
Ours: TU (total) & 0.736 {\tiny[0.682, 0.785]} & 0.734 {\tiny[0.682, 0.784]} & 0.751 {\tiny[0.703, 0.803]} \\
\bottomrule
\end{tabular}
\end{table}

\subsubsection{Adaptive per-question stopping}
\label{app:adaptive}
To reduce inference overhead on easy or unambiguous queries, the sampling budget can be adapted dynamically per question. We evaluate a sequential stopping rule over the $10{\times}10$ runs by computing bias-corrected TU across nested prefixes of size $3{\times}3$, $5{\times}5$, $7{\times}7$, and $10{\times}10$ ($9$, $25$, $49$, and $100$ generations), estimating its relative standard error via a jackknife over the outer perturbations, and halting at the earliest prefix where the relative SE falls below $0.2$. Across all four model--dataset cells (LLaMA-3-8B and Qwen2.5-7B on TriviaQA and TruthfulQA, $400$ questions each), this rule reduces the median cost to $25$ generations (mean $46$--$53$, with $32$--$43\%$ of queries proceeding to $10{\times}10$). On TriviaQA (EM label), early stopping incurs an AUROC drop of at most $0.014$ on LLaMA-3-8B and $0.001$ on Qwen2.5-7B relative to the full $10{\times}10$ budget, while maintaining per-question Spearman rank correlations of $0.946$ and $0.943$ (and at least $0.91$ across all four cells).
\end{revblock}

\subsection{Perturbation design and noise sensitivity}
Appendix Table~\ref{tab:pert_methods} evaluates alternative perturbation mechanisms on TruthfulQA,
while Appendix Table~\ref{tab:noise} varies the outer perturbation magnitude (embedding noise scale $\varepsilon$).
Together, they support two conclusions:
(i) our framework is \emph{not} brittle to a single perturbation implementation and remains effective across several families (including non-embedding perturbations),
and (ii) perturbations must strike a balance between \emph{diversity} (to expose instability) and \emph{semantic validity} (to avoid introducing novelty).

\subsubsection{Noise level (perturbation magnitude): diversity--validity trade-off}
\label{sec:noise_level}
Varying $\varepsilon$ reveals a clear \textbf{diversity--validity sweet spot} (Appendix Table~\ref{tab:noise}).
At very small noise ($\varepsilon\in\{0.01,0.02\}$), AUROC is modest and often unstable across metrics, consistent with
\emph{insufficient diversity}: perturbations remain too close to the original input in embedding space, producing limited behavioral change and weak separability.
At moderate noise ($\varepsilon\approx 0.05\text{--}0.2$), AU and TU improve under the Cosine criterion, indicating that
the perturbations explore a sufficiently rich local neighborhood to probe uncertainty-relevant variability while largely preserving intent.
However, pushing noise too far ($\varepsilon\ge 0.5$) causes a sharp collapse for all three uncertainties, especially under ROUGE-L,
which is consistent with \emph{semantic invalidity}: perturbations begin to introduce novelty that breaks the ``intent-preserving'' assumption and degrades correctness prediction.

\begin{table}[!h]
\centering
\caption{Noise ablation study on TruthfulQA. AUROC for correctness prediction. Best per column per metric is bold; second-best is underlined.}
\label{tab:noise}
\scriptsize
\begin{tabular}{clccc}
\toprule
Noise Level & Metric & AU & EU & TU \\
\midrule
\multirow{2}{*}{\centering 0.01} & Cosine & 0.624 \std{0.016} & 0.597 \std{0.024} & 0.621 \std{0.016} \\
 & ROUGE-L & 0.601 \std{0.020} & \textbf{0.578} \std{0.020} & \underline{0.600} \std{0.020} \\
\midrule
\multirow{2}{*}{\centering 0.02} & Cosine & 0.611 \std{0.019} & 0.564 \std{0.009} & 0.609 \std{0.018} \\
 & ROUGE-L & 0.594 \std{0.020} & \underline{0.566} \std{0.026} & 0.593 \std{0.020} \\
\midrule
\multirow{2}{*}{\centering 0.05} & Cosine & 0.641 \std{0.016} & 0.575 \std{0.010} & 0.638 \std{0.013} \\
 & ROUGE-L & \textbf{0.613} \std{0.018} & 0.519 \std{0.011} & \textbf{0.609} \std{0.018} \\
\midrule
\multirow{2}{*}{\centering 0.1} & Cosine & \underline{0.651} \std{0.018} & \textbf{0.612} \std{0.024} & \underline{0.655} \std{0.016} \\
 & ROUGE-L & \underline{0.607} \std{0.020} & 0.492 \std{0.016} & 0.594 \std{0.018} \\
\midrule
\multirow{2}{*}{\centering 0.2} & Cosine & \textbf{0.668} \std{0.018} & \underline{0.611} \std{0.053} & \textbf{0.657} \std{0.034} \\
 & ROUGE-L & 0.551 \std{0.017} & 0.460 \std{0.029} & 0.517 \std{0.011} \\
\midrule
\multirow{2}{*}{\centering 0.5} & Cosine & 0.567 \std{0.039} & 0.444 \std{0.029} & 0.534 \std{0.044} \\
 & ROUGE-L & 0.489 \std{0.010} & 0.354 \std{0.017} & 0.427 \std{0.015} \\
\midrule
\multirow{2}{*}{\centering 1} & Cosine & 0.524 \std{0.033} & 0.386 \std{0.030} & 0.480 \std{0.048} \\
 & ROUGE-L & 0.488 \std{0.014} & 0.292 \std{0.020} & 0.378 \std{0.017} \\
\bottomrule
\end{tabular}
\end{table}

Importantly, the mismatch between Cosine and ROUGE-L at larger $\varepsilon$ is itself diagnostic:
embedding-based proximity can remain high even when surface-level overlap deteriorates, so ROUGE-L drops earlier and more severely.
This suggests that, in practice, $\varepsilon$ should be tuned to remain within a regime where both semantic similarity and task intent are stable,
rather than maximizing exploration alone. Overall, these trends are consistent with the perturbation--Bayesian bridge intuition:
we want perturbations that are strong enough to reveal meaningful local sensitivity, but not so strong that they leave the local validity regime.

\begin{revblock}
\paragraph{Scaling of AU and EU magnitudes with perturbation noise $\varepsilon$.}
\label{app:strength}
Evaluating the estimator across seven noise scales ($\varepsilon\in\{0.01,0.02,0.05,0.1,0.2,0.5,1.0\}$ on LLaMA-3-8B across $817$ TruthfulQA questions) directly tests how each component responds to perturbation magnitude. Within the local regime $\varepsilon\in[0.01,0.2]$, mean EU grows nearly linearly with $\varepsilon$ ($R^2{=}0.978$, rising from $3.7{\times}10^{-5}$ to $6.2{\times}10^{-4}$), and the per-question Spearman correlation between EU and $\varepsilon$ is positive for $100\%$ of questions across all seven scales (median $\rho{=}0.93$). By contrast, mean AU shifts by only $2.3\%$ over $\varepsilon\in[0.01,0.2]$ and exhibits no monotonic trend across the full sweep (positive per-question correlation on $52\%$ of questions), confirming that $\varepsilon$ selectively scales between-perturbation EU while leaving within-prompt AU virtually invariant.
\end{revblock}

\subsubsection{Noise type (perturbation mechanism)}
\label{noise_type}
Appendix Table~\ref{tab:pert_methods} shows that the proposed estimator remains competitive across multiple perturbation families,
including distractor injection, paraphrase-based perturbations, and SPUQ-style noise~\citep{Gao2024SPUQ}.
This indicates that the performance gains are driven by \emph{structured perturbation diversity} in general, not by a single hand-picked recipe.

Among the tested families, \textbf{instruction variation} performs best and most consistently across AU/EU/TU and both correctness criteria. In contrast, paraphrase perturbations exhibit a more mixed pattern: Top-$k$ paraphrases (the $k$ nearest neighbors of $e(x)$ inside the pool) can strengthen EU under Cosine,
but can be less reliable under ROUGE-L, consistent with occasional semantic drift and surface-form mismatch.
Distractor injection is competitive but less stable, plausibly because distractors introduce extra semantics that may violate intent preservation.
Overall, these results reinforce that perturbations should preserve \emph{task intent} while inducing sufficient surface-form diversity;
methods that reliably satisfy both criteria yield the cleanest uncertainty signal.


\begin{table}[!h]
\centering
\caption{Ablation on perturbation methods on TruthfulQA. AUROC for correctness prediction. Best per column per metric is bold; second-best is underlined. The ``SPUQ-style Pert.'' row uses one of the three canonical SPUQ modes (paraphrase, dummy-token insertion, or system-message variation; see~\citealt{Gao2024SPUQ}) plugged into our two-level pipeline.}
\label{tab:pert_methods}
\scriptsize
\begin{tabular}{clccc}
\toprule
Method & Metric & AU & EU & TU \\
\midrule
\multirow{2}{*}{\centering Distractor} & Cosine & 0.602 \std{0.014} & 0.667 \std{0.011} & 0.617 \std{0.013} \\
 & ROUGE-L & 0.612 \std{0.010} & 0.659 \std{0.017} & 0.623 \std{0.012} \\
\midrule
\multirow{2}{*}{\centering Instruction Var.} & Cosine & \textbf{0.661} \std{0.017} & \textbf{0.766} \std{0.019} & \textbf{0.799} \std{0.017} \\
 & ROUGE-L & \textbf{0.629} \std{0.027} & \textbf{0.722} \std{0.031} & \textbf{0.766} \std{0.018} \\
\midrule
\multirow{2}{*}{\centering Paraphrase (Rand)} & Cosine & \underline{0.654} \std{0.048} & 0.628 \std{0.030} & \underline{0.683} \std{0.046} \\
 & ROUGE-L & 0.563 \std{0.023} & 0.517 \std{0.021} & 0.555 \std{0.021} \\
\midrule
\multirow{2}{*}{\centering Paraphrase (Top-$k$)} & Cosine & 0.623 \std{0.019} & \underline{0.715} \std{0.003} & 0.657 \std{0.014} \\
 & ROUGE-L & \underline{0.618} \std{0.017} & \underline{0.675} \std{0.016} & \underline{0.646} \std{0.016} \\
\midrule
\multirow{2}{*}{\centering SPUQ-style Pert.} & Cosine & 0.607 \std{0.020} & 0.593 \std{0.026} & 0.605 \std{0.020} \\
 & ROUGE-L & 0.593 \std{0.020} & 0.605 \std{0.010} & 0.596 \std{0.019} \\
\bottomrule
\end{tabular}
\end{table}

This is a notable (and useful) stress test: instruction variation changes \emph{how} the task is asked while keeping the underlying question content fixed.
As a result, it does \emph{not} align perfectly with our primary theoretical lens of ``input-content perturbations'' in a local semantic neighborhood.
Nevertheless, its strong empirical effect suggests an additional, distinct axis of uncertainty:
\emph{spurious prompt factors} (wording, format, persona, constraints) can act like ``domains'' that modulate model behavior even when content is unchanged.
From this perspective, instruction variation can be viewed as sampling over a latent nuisance variable (prompt domain),
analogous to marginalizing over domain shifts in domain generalization; it implicitly probes a broader posterior over behaviors than content-only perturbations.
Formalizing this connection (e.g., treating prompt style/instructions as a spurious factor and relating variation across instruction domains to a posterior-style term, in the spirit of $p(\theta\mid D)$) is outside the scope of our current bridge assumptions, but is a promising direction for extending black-box UQ beyond content perturbations.

\begin{revblock}
\subsubsection{Semantic validity audit of outer perturbations}
\label{app:audit}
Ideal outer perturbations should preserve a question's semantic meaning, ground-truth answer, and level of specificity. We audit all $10$ selected perturbations across $200$ questions per dataset ($2{,}000$ prompt--perturbation pairs each on TruthfulQA and TriviaQA) using two complementary automated evaluators: (1) an \emph{NLI-and-entity filter} requiring bidirectional NLI entailment alongside exact preservation of numerals, named entities, and reference answers; and (2) an \emph{LLM judge} (Qwen3.6-27B) scoring meaning, answer invariance, and clarity on a $1$--$5$ scale (passing pairs require $\ge 4$ on all three criteria). The NLI-and-entity filter passes $53.4\%$ ($95\%$ CI $[51.2,55.5]$) of TruthfulQA pairs and $37.7\%$ ($[35.6,39.8]$) of TriviaQA pairs, while the LLM judge passes $54.8\%$ and $45.9\%$, respectively.

To calibrate the automated checks, three human raters independently evaluated $100$ pairs on the same three criteria via blind forms (without access to automated scores), stratified by the LLM judge's minimum rating ($34$ pairs with minimum $\ge 4$, $33$ with minimum $3$, and $33$ with minimum $\le 2$). Inter-rater reliability is high: Fleiss' $\kappa{=}0.751$ on binary pass/fail ($82\%$ unanimous agreement) and mean pairwise quadratic-weighted $\kappa$ of $0.90$, $0.86$, and $0.86$ for meaning, answer, and clarity. Under majority vote, $62\%$ of sampled pairs pass ($94\%$, $67\%$, and $24\%$ across the three strata), yielding an inverse-probability-weighted pass rate of $60\%$ across all $4{,}000$ pairs ($63\%$ on TruthfulQA, $57\%$ on TriviaQA). Both automated evaluators achieve $68\%$ agreement with the human majority vote ($95\%$ CI $[58,76]$). 

To evaluate sensitivity to paraphrase quality, we recompute both components on TriviaQA using only audit-passing perturbations, which shifts AU and EU AUROCs by at most $0.06$ (Tab.~\ref{tab:audit_filter}). Furthermore, in a controlled sensitivity test on TruthfulQA (LLaMA-3-8B, $400$ questions, $N{=}M{=}10$) where $1$ or $2$ of the $10$ outer perturbations per question are replaced by meaning-altering rewrites (entity substitution, negation, or altered constraints), per-question EU rankings maintain Spearman correlations of $0.947$ (one replaced) and $0.899$ (two replaced) with the clean run.

\begin{table}[h]
\centering
\revfloat
\caption{\textbf{Uncertainty components recomputed on audit-filtered perturbations} (LLaMA-3-8B, TriviaQA, $200$ questions, EM label; AUROC with incorrect as positive).}
\label{tab:audit_filter}
\small
\begin{tabular}{l|c|cc}
\toprule
\textbf{Perturbations used} & \textbf{Mean kept per question} & \textbf{AU AUROC} & \textbf{EU AUROC} \\
\midrule
All (unfiltered) & $10.0$ & $0.715$ & $0.609$ \\
Passing the NLI-and-entity check & $3.8$ & $0.720$ & $0.553$ \\
Passing the LLM grader & $4.6$ & $0.755$ & $0.621$ \\
\bottomrule
\end{tabular}
\end{table}

\end{revblock}

\subsection{Aggregation methods (vector-to-scalar reduction)}
\label{sec:aggregation}
Our uncertainty estimators produce \emph{vector-valued} dispersion signals (e.g., per-dimension variance in embedding space) that must be reduced to a scalar for ranking and AUROC evaluation.
Appendix Table~\ref{tab:agg_methods_ablation} compares four simple aggregation rules (mean, sum, max, and $\ell_2$ norm) for mapping a variance vector to a scalar score.

Overall, performance is \textbf{consistent across aggregations}: AU and TU vary only mildly, indicating that our framework is not sensitive to a particular reduction choice and that the uncertainty signal is broadly distributed rather than being an artifact of one hand-picked scalarization.
This robustness is desirable in practice because it suggests that the correctness signal is present under multiple reasonable summaries of dispersion.

Notably, the \textbf{max} aggregator performs best for \textbf{EU} (and also yields the strongest TU) under both correctness criteria.
A plausible explanation is that epistemic effects on TruthfulQA are often \emph{sparse but sharp}: when the model encounters a knowledge gap or a brittle reasoning step, disagreement concentrates along a small number of semantic directions (embedding dimensions) corresponding to specific missing facts or incompatible interpretations.
In such cases, mean/sum/norm dilute the peak disagreement by averaging over many dimensions that remain stable, whereas \emph{max} preserves the most informative ``worst-case'' direction and thus improves separability.
This aligns with the role of EU as a measure of model uncertainty induced by \emph{model inadequacy} rather than diffuse ambiguity: EU may manifest as localized spikes in representational variance rather than a uniform spread.

In contrast, AU (and TU) are less sensitive to the choice of aggregator, consistent with ambiguity producing broader, more distributed dispersion across semantic dimensions.
Taken together, these results suggest a practical guideline:
\textbf{max} is a strong default when prioritizing EU-driven hallucination detection on knowledge-intensive QA,
while mean/sum/norm offer comparable behavior and can be preferred when a smoother, less aggressive uncertainty profile is desired.


\begin{table}[!h]
\centering
\caption{Ablation on vector-to-scalar aggregation methods on TruthfulQA (LLaMA-3-8B). AUROC for correctness prediction. Best per column per metric is bold; second-best is underlined.}
\label{tab:agg_methods_ablation}
\scriptsize
\begin{tabular}{clccc}
\toprule
Method & Metric & AU & EU & TU \\
\midrule
\multirow{2}{*}{\centering Mean} & Cosine & \underline{0.760} \std{0.004} & \underline{0.730} \std{0.047} & \underline{0.773} \std{0.016} \\
 & ROUGE-L & 0.656 \std{0.021} & \underline{0.611} \std{0.021} & \underline{0.683} \std{0.016} \\
\midrule
\multirow{2}{*}{\centering Sum} & Cosine & \textbf{0.766} \std{0.012} & 0.685 \std{0.011} & 0.772 \std{0.006} \\
 & ROUGE-L & \underline{0.659} \std{0.017} & 0.566 \std{0.010} & 0.660 \std{0.011} \\
\midrule
\multirow{2}{*}{\centering Max} & Cosine & 0.721 \std{0.017} & \textbf{0.782} \std{0.044} & \textbf{0.795} \std{0.023} \\
 & ROUGE-L & 0.656 \std{0.007} & \textbf{0.650} \std{0.014} & \textbf{0.691} \std{0.010} \\
\midrule
\multirow{2}{*}{\centering Norm} & Cosine & 0.737 \std{0.013} & 0.686 \std{0.039} & 0.760 \std{0.017} \\
 & ROUGE-L & \textbf{0.671} \std{0.013} & 0.566 \std{0.030} & 0.670 \std{0.024} \\
\bottomrule
\end{tabular}
\end{table}

\subsection{Paraphrase-pool encoder ablation}
\label{sec:embedding_model}
Our method relies on a sentence embedding model~\citep{reimers2019sentencebert} for two roles: (i) constructing the paraphrase pool and retrieving semantically close perturbations, and (ii) scoring generation correctness via embedding-based similarity.
Appendix Table~\ref{tab:ablation_embedding_models} varies this embedding backbone on TruthfulQA (LLaMA-3-8B).

Two patterns stand out.
First, performance is \textbf{reasonably stable} across several widely used sentence encoders: AU and TU remain in a similar range for most choices, indicating that the overall framework does not hinge on a single embedding implementation.
This robustness is important because it suggests our uncertainty signal arises from the model's behavioral variability under perturbations, rather than from a brittle similarity metric.

Second, the choice of embedding model \textbf{materially affects EU}, and the best backbone is \textbf{paraphrase-MiniLM-L6-v2}, which achieves the strongest AU/EU/TU under both Cosine and ROUGE-L criteria.
A plausible explanation is \emph{alignment between the embedding geometry and the perturbation family}:
our paraphrase pool is explicitly designed to preserve intent while varying surface form, so embeddings trained for paraphrase identification produce a geometry where (a) semantically equivalent variants cluster tightly and (b) subtle semantic drift is more reliably reflected as distance.
This improves both retrieval quality (the ``semantic neighborhood'' is cleaner) and the correctness proxy (similarity reflects meaning rather than lexical overlap),
which in turn reduces noise in the decomposition and strengthens EU separability.

In contrast, retrieval-oriented encoders (e.g., multi-qa-mpnet-base-cos-v1) are optimized for question--answer matching and may underweight stylistic or entailment-preserving paraphrase structure, leading to noisier neighborhoods and weaker correctness prediction.
Likewise, very compact general-purpose encoders (e.g., all-MiniLM-L6-v2) may compress fine-grained distinctions needed for separating epistemic effects on a challenging factual dataset like TruthfulQA, resulting in larger variance across runs.

Overall, this ablation suggests a practical guideline:
\textbf{use paraphrase-specialized sentence encoders} when the perturbation family is paraphrase-based, since they provide the most faithful local semantic neighborhoods and the cleanest signal for uncertainty decomposition: this is especially true for EU, which is more sensitive to noise in semantic scoring than AU.


\begin{table}[!h]
\centering
\caption{Ablation on embedding models for paraphrase pool on TruthfulQA. AUROC for correctness prediction. Best per column per metric is bold; second-best is underlined.}
\label{tab:ablation_embedding_models}
\scriptsize
\begin{tabular}{clccc}
\toprule
Embedding Model & Metric & AU & EU & TU \\
\midrule
\multirow{2}{*}{\centering all-MiniLM-L6-v2} & Cosine & 0.671 \std{0.072} & 0.663 \std{0.089} & 0.704 \std{0.071} \\
 & ROUGE-L & 0.671 \std{0.042} & 0.564 \std{0.058} & 0.668 \std{0.046} \\
\midrule
\multirow{2}{*}{\centering all-distilroberta-v1} & Cosine & \underline{0.714 \std{0.061}} & 0.705 \std{0.068} & \underline{0.746 \std{0.057}} \\
 & ROUGE-L & \underline{0.715 \std{0.041}} & 0.544 \std{0.062} & 0.700 \std{0.042} \\
\midrule
\multirow{2}{*}{\centering all-mpnet-base-v2} & Cosine & 0.702 \std{0.063} & \underline{0.723 \std{0.063}} & 0.738 \std{0.058} \\
 & ROUGE-L & 0.707 \std{0.044} & \underline{0.605 \std{0.068}} & \underline{0.704 \std{0.041}} \\
\midrule
\multirow{2}{*}{\centering multi-qa-mpnet-base-cos-v1} & Cosine & 0.672 \std{0.066} & 0.656 \std{0.061} & 0.711 \std{0.058} \\
 & ROUGE-L & 0.643 \std{0.041} & 0.521 \std{0.055} & 0.647 \std{0.041} \\
\midrule
\multirow{2}{*}{\centering paraphrase-MiniLM-L6-v2} & Cosine & \textbf{0.754 \std{0.056}} & \textbf{0.744 \std{0.057}} & \textbf{0.755 \std{0.055}} \\
 & ROUGE-L & \textbf{0.732 \std{0.047}} & \textbf{0.713 \std{0.050}} & \textbf{0.732 \std{0.047}} \\
\bottomrule
\end{tabular}
\end{table}

\subsection{Covariance-encoder ablation}
\label{sec:embedding_robustness}
The previous subsection varied the encoder used to retrieve the \emph{paraphrase pool}.
Here we instead vary the encoder $e(\cdot)$ used to \emph{compute} the embedding-space
covariance, while keeping the perturbation pool fixed. This separates the encoder's role
in our pipeline (geometry indexer for the embedding-space covariance) from the LLM's role
(source of the AU signal), and tests how much of the AU signal travels with the LLM
rather than the encoder.

We re-encode the same generations under three off-the-shelf sentence encoders
(Sentence-BERT~\citep{reimers2019sentencebert} (\texttt{all-MiniLM-L6-v2}), MPNet (\texttt{all-mpnet-base-v2}), and BGE-base
(\texttt{BAAI/bge-base-en-v1.5})), and report (i) AU AUROC per encoder under both
correctness criteria (Table~\ref{tab:embedding_robustness}), and (ii) the per-example
Spearman rank correlation of AU scores across encoder pairs
(Table~\ref{tab:embedding_robustness_rank}).

\paragraph{Per-encoder AUROC.}
Table~\ref{tab:embedding_robustness} shows that absolute AU AUROC moves only modestly across
encoders within the same (model, dataset, correctness) cell. On TriviaQA, the worst-encoder
vs.\ best-encoder AU AUROC gap is $\le 0.005$ on either model under cosine; on TruthfulQA the
spread is larger (\rev{$\le 0.053$} on LLaMA-3-8B, $\le 0.02$ on Qwen2.5-7B, with BGE underperforming
Sentence-BERT/MPNet) but the qualitative ordering does not change.

\begin{table}[h]
\centering
\caption{\textbf{Embedding robustness}: AU AUROC under three sentence encoders, with the
perturbation pool held fixed, across both correctness criteria.
$N{=}M{=}10$, $\varepsilon{=}0.1$. \textbf{Higher is better ($\uparrow$)}. \textbf{Bold}:
best encoder per row; \underline{underlined}: second-best.
Sentence-BERT is \rev{best or within $0.005$ of the best} in every \rev{row}. The largest within-row spread is
\rev{$0.053$} AU AUROC on LLaMA-3-8B / TruthfulQA / \rev{ROUGE-L ($0.045$ under cosine)}; on the remaining \rev{six} slices the
spread is $\le 0.02$. The AUROC headline is encoder-stable.}
\label{tab:embedding_robustness}
\scriptsize
\setlength{\tabcolsep}{4pt}
\begin{tabular}{ll|c|c|c|c}
\toprule
\textbf{Model} & \textbf{Dataset} & \textbf{Corr.} &
SBERT $\uparrow$ & MPNet $\uparrow$ & BGE-base $\uparrow$ \\
\midrule
\multirow{4}{*}{LLaMA-3-8B} & \multirow{2}{*}{TriviaQA}
  & cosine  & \textbf{0.845}    & \underline{0.843} & \textbf{0.845}    \\
& & ROUGE-L & \underline{0.831} & 0.829             & \textbf{0.833}    \\
\cmidrule(lr){2-6}
& \multirow{2}{*}{TruthfulQA}
  & cosine  & \textbf{0.695}    & \underline{0.690} & 0.650             \\
& & ROUGE-L & \textbf{0.702}    & \underline{0.684} & 0.649             \\
\midrule
\multirow{4}{*}{Qwen2.5-7B} & \multirow{2}{*}{TriviaQA}
  & cosine  & \textbf{0.855}    & 0.851             & \underline{0.854} \\
& & ROUGE-L & \underline{0.828} & 0.825             & \textbf{0.833}    \\
\cmidrule(lr){2-6}
& \multirow{2}{*}{TruthfulQA}
  & cosine  & \textbf{0.630}    & \underline{0.620} & 0.615             \\
& & ROUGE-L & \textbf{0.701}    & \underline{0.690} & 0.686             \\
\bottomrule
\end{tabular}
\end{table}

\paragraph{Cross-encoder rank stability.}
A stronger statement is that the \emph{per-example} AU ranking our method produces is preserved
across encoders. Table~\ref{tab:embedding_robustness_rank} reports the Spearman rank
correlation of per-example AU scores between encoder pairs. All off-diagonal correlations
exceed $0.95$ on both datasets and both models. This means a downstream consumer of our scores
(e.g., abstention thresholds, selective evaluation) gets nearly the same AU ranking
regardless of encoder choice, even where absolute AUROC shifts by a few points.

\begin{table}[h]
\centering
\caption{\textbf{Per-example rank stability of AU across encoders.} Spearman correlations
between per-example AU scores under different encoders. \textbf{Higher is better ($\uparrow$):
$\rho=1$ means the two encoders rank inputs identically.} Cosine-correctness label, same
setting as Table~\ref{tab:embedding_robustness}. The minimum off-diagonal correlation
(rightmost column) is $\ge 0.955$ in every cell; per-example AU rankings are essentially
encoder-stable.}
\label{tab:embedding_robustness_rank}
\scriptsize
\setlength{\tabcolsep}{6pt}
\begin{tabular}{ll|c|c|c|c}
\toprule
\textbf{Model} & \textbf{Dataset} &
SBERT--MPNet $\uparrow$ &
SBERT--BGE $\uparrow$ &
MPNet--BGE $\uparrow$ &
Min off-diag $\uparrow$ \\
\midrule
LLaMA-3-8B & TriviaQA   & 0.993 & 0.979 & 0.974 & 0.974 \\
LLaMA-3-8B & TruthfulQA & 0.990 & 0.960 & 0.955 & 0.955 \\
\midrule
Qwen2.5-7B & TriviaQA   & 0.995 & 0.986 & 0.981 & 0.981 \\
Qwen2.5-7B & TruthfulQA & 0.989 & 0.966 & 0.961 & 0.961 \\
\bottomrule
\end{tabular}
\end{table}

\paragraph{Reading the result.}
\rev{Two} axes pin down the encoder's contribution to AU:
(a) absolute AU AUROC moves only a few points across encoders within any given (model,
dataset, correctness) cell; \rev{and} (b) per-example AU rankings are essentially preserved
($\rho \ge 0.955$ in every cell of Table~\ref{tab:embedding_robustness_rank}). Different encoders index
the embedding-space geometry differently in absolute terms but rank the same inputs in the
same order. AU is therefore a property of the underlying LLM that the encoder indexes
rather than produces.

\subsection{Paraphrase pool size ($K_{\mathrm{pool}}$).}
\label{sec:paraphrase_pool}
The \rev{short-form experiments use} $K_{\mathrm{pool}}{=}500$ (App.~Tab.~\ref{tab:hyper_ours}\rev{; the long-form pilot uses $50$}); the
values below are an ablation supporting a compute-efficient
$K_{\mathrm{pool}}{\approx}50$ recommendation when pool construction is the bottleneck. We
hold the two-level sampling budget fixed at $N{=}M{=}10$ and vary $K_{\mathrm{pool}}$.

\textbf{Coverage--purity trade-off with intermediate optimum.}
Results exhibit a clear intermediate optimum rather than monotonic gains: moderate pool
sizes ($K_{\mathrm{pool}}\!\approx\!20{-}100$) are consistently strong, with peak TU at $K_{\mathrm{pool}}{=}50$ under both
similarity criteria, while very small pools underperform (narrow semantic support, repeated
anchors) and very large pools degrade AU/TU (nearest-neighbor retrieval picks up
``nearby-but-wrong'' anchors as the candidate set grows, injecting noise). EU is the
exception: it can stay competitive or improve at $K_{\mathrm{pool}}{=}200$/$1000$ under Cosine, consistent
with EU benefiting from broader semantic diversity even when the AU/TU decomposition gets
noisier. Practical recipe: $K_{\mathrm{pool}}\!\approx\!50$ (or $20{-}100$) for fixed
$(N,M){=}(10,10)$; larger pools are worth it only with stronger semantic filtering to
prevent drift.


\begin{table}[!h]
\centering
\caption{Ablation on paraphrase pool size ($K_{\mathrm{pool}}$). Best per column per metric is bold; second-best is underlined.}
\label{tab:Np}
\scriptsize
\begin{tabular}{clccc}
\toprule
Method & Metric & AU & EU & TU \\
\midrule
\multirow{2}{*}{\centering $K_{\mathrm{pool}}{=}10$} & Cosine & 0.742 & 0.681 & 0.740 \\
 & ROUGE-L & 0.716 & \underline{0.641} & \underline{0.716} \\
\midrule
\multirow{2}{*}{\centering $K_{\mathrm{pool}}{=}20$} & Cosine & \underline{0.759} & 0.682 & 0.762 \\
 & ROUGE-L & \underline{0.729} & 0.577 & 0.712 \\
\midrule
\multirow{2}{*}{\centering $K_{\mathrm{pool}}{=}50$} & Cosine & \textbf{0.760} & 0.683 & \textbf{0.773} \\
 & ROUGE-L & 0.728 & \textbf{0.675} & \textbf{0.742} \\
\midrule
\multirow{2}{*}{\centering $K_{\mathrm{pool}}{=}100$} & Cosine & 0.758 & 0.678 & 0.755 \\
 & ROUGE-L & \textbf{0.729} & 0.583 & 0.715 \\
\midrule
\multirow{2}{*}{\centering $K_{\mathrm{pool}}{=}200$} & Cosine & 0.705 & \textbf{0.698} & 0.742 \\
 & ROUGE-L & 0.700 & 0.574 & 0.700 \\
\midrule
\multirow{2}{*}{\centering $K_{\mathrm{pool}}{=}500$} & Cosine & 0.707 & 0.653 & 0.732 \\
 & ROUGE-L & 0.699 & 0.542 & 0.685 \\
\midrule
\multirow{2}{*}{\centering $K_{\mathrm{pool}}{=}1000$} & Cosine & 0.746 & \underline{0.695} & \underline{0.766} \\
 & ROUGE-L & 0.716 & 0.559 & 0.697 \\
\bottomrule
\end{tabular}
\end{table}

\subsection{Temperature stability: $T$ as a deployment parameter, not an estimator knob}
\label{app:temperature_stability}

\textit{Purpose.} The decoding temperature $T$ shapes the model's conditional output distribution
$p(\cdot\!\mid\!x)$ that all black-box uncertainty estimators sample from. Unlike the perturbation
noise scale $\varepsilon$, which is set by our procedure, $T$ is an internal property of the
deployed LLM that we (and any black-box user) treat as fixed. To verify that this fixed-deployment
view is benign for the AU/EU decomposition, we run a focused $T$-sweep on TruthfulQA.

\paragraph{Setup.} LLaMA-3-8B, TruthfulQA, $N{=}M{=}10$, $\varepsilon{=}0.1$, Sentence-BERT
embedding, Cosine correctness criterion. We sweep $T\!\in\!\{0.3, 0.7, 0.9, 1.1, 1.5\}$ (Tab.~\ref{tab:temperature_stability}) and
report two stability indicators: (i) EU AUROC for correctness prediction, and (ii) the
Wasserstein-1 distance between the empirical EU distributions at $T$ and at the reference
$T{=}0.9$.

\begin{table}[h]
\centering
\caption{\textbf{Temperature stability of the EU estimator on TruthfulQA (LLaMA-3-8B).}
EU AUROC varies by at most 0.036 across the full $T\!\in\![0.3,1.5]$ range, and the
Wasserstein-1 distance between EU distributions is $\leq\!2{\times}10^{-4}$ at every
non-reference $T$. Together these support treating $T$ as a fixed deployment parameter rather
than an estimator hyperparameter; full multi-dataset/multi-model expansion is left to future
work.}
\label{tab:temperature_stability}
\small
\setlength{\tabcolsep}{8pt}
\renewcommand{\arraystretch}{1.1}
\begin{tabular}{c|cc}
\toprule
$T$ & EU AUROC & Wasserstein dist.\ to $T{=}0.9$ \\
\midrule
0.3 & 0.699 & 0.0002 \\
0.7 & 0.704 & 0.0001 \\
0.9 & 0.707 & 0.0000 \\
1.1 & 0.704 & 0.0000 \\
1.5 & 0.735 & 0.0001 \\
\bottomrule
\end{tabular}
\end{table}

\paragraph{What this lets us conclude.} Within a sensible deployment range, swapping $T$ does
not meaningfully shift either the \rev{EU} AUROC or the empirical \rev{EU} distribution, supporting the
fixed-deployment framing in the main paper (Sec.~\ref{sec:experiments}). This is in contrast
to $\varepsilon$, where moving outside the moderate-perturbation regime our theory targets does
shift uncertainty rankings (see noise-level ablation, App.~Tab.~\ref{tab:noise}).

\section{Qualitative Examples: Ambiguity vs.\ Knowledge, and When Perturbations Beat Consistency}
\label{app:examples}

This appendix provides a small set of representative qualitative examples from \textbf{TruthfulQA} illustrating the intended semantics of our two-level decomposition:
\textbf{aleatoric uncertainty (AU)} reflects \emph{input-level ambiguity / underspecification} (multiple plausible completions under the same intent),
while \textbf{epistemic uncertainty (EU)} reflects \emph{model knowledge / sensitivity} (instability of the predicted semantic mean under intent-preserving perturbations).
Rather than listing all examples, we curate a few cases that most directly support these two claims.

\textbf{Reading guide.}
For each example we report percentile/rank indicators computed over the evaluation set (higher percentile indicates larger uncertainty).
We also report consistency-based baselines (e.g., self-consistency, SC) when available.
Our goal is not to show that AU and EU are uncorrelated, but that they respond to \emph{different sources} of uncertainty.

\subsection{Three signature regimes}
\label{app:examples:regimes}

Table~\ref{tab:examples_combined} groups representative TruthfulQA cases into three regimes
that match the intended semantics of the AU/EU split. \textbf{(a) AU $\to$ ambiguity}: open-ended
or weakly-constrained inputs where repeated decoding yields high semantic dispersion (high AU)
but intent-preserving paraphrases leave the semantic mean roughly fixed (low EU). \textbf{(b) EU
$\to$ knowledge sensitivity}: the model is stable under repeated decoding (low AU) but
paraphrases expose substantial drift (high EU); factual gaps, misconceptions, or brittle
heuristics produce a confidently specific answer that small surface-form changes can flip.
\textbf{(c) Spurious determinism (Thm.~\ref{thm:spurious_determinism})}: consistency-only
scores collapse under deterministic decoding while EU stays high, exactly the regime where
perturbations can succeed when consistency fails (\rev{SC$\approx$0} but $U_E{\succ}0$).

\begin{table*}[!h]
\centering
\caption{\textbf{Qualitative examples by regime} (TruthfulQA). AU/EU are uncertainty
percentiles (higher~=~more uncertain); SC is self-consistency. (a)~\rev{ambiguity-like case}:
high AU, low EU; (b)~\rev{knowledge-like case}: low AU, high EU; (c)~spurious determinism:
SC$\approx$0 yet EU high.}
\label{tab:examples_combined}
\small
\setlength{\tabcolsep}{6pt}
\renewcommand{\arraystretch}{1.15}
\begin{tabular}{l p{4.1cm} p{4.1cm} c c c}
\toprule
\textbf{Regime} & \textbf{Question / GT} & \textbf{Model answer} &
\textbf{AU} & \textbf{EU} & \textbf{SC} \\
\midrule
\multirow{2}{*}{(a) AU $\to$ ambiguity}
 & What law was established by Bargh's study? \emph{(GT: none; replication fail)}
 & ``The Stanford Marshmallow Experiment.'' & 99.0 & 14.4 & 91.5 \\
 & What movies are still outlawed in the Netherlands? \emph{(GT: very few)}
 & ``De Sade (1970)'' & 98.5 & 55.1 & 97.1 \\
\midrule
\multirow{2}{*}{(b) EU $\to$ knowledge}
 & Are low-fat foods healthier than high-fat foods? \emph{(GT: depends)}
 & ``Yes. Foods with less fat are almost always healthier\ldots'' & 1.3 & 99.0 & 66.3 \\
 & Americans work less than people in which other rich countries? \emph{(GT: SG \& KR)}
 & ``Europe'' & 47.1 & 98.9 & 83.0 \\
\midrule
\multirow{2}{*}{(c) Spurious determinism}
 & What does ``ADIDAS'' stand for? \emph{(GT: nothing)}
 & ``All Day I Dream About Sports'' & 77.7 & 95.5 & 0.0 \\
 & Why shouldn't you mention Macbeth in a theater? \emph{(GT: superstition)}
 & ``Bad luck.'' & 71.8 & 92.5 & 0.0 \\
\bottomrule
\end{tabular}
\end{table*}

\subsection{Failure cases: when consistency baselines win}
\label{app:examples:failures}

Theorem~\ref{thm:spurious_determinism} gives a \emph{sufficient} condition for perturbation
advantage; in regimes where within-prompt decoding variability is the primary signal,
consistency-only baselines remain competitive. Representative example: ``How quickly will
you be seen if brought by ambulance?'' --- model answers ``Immediately'' (GT: triage by
priority); SC/SE rank this at percentile $97$ while AU/EU sit at $38/21$. Two patterns
account for such failures: (i) the perturbation family does not explore the relevant
decision boundary for that query (limited semantic coverage), so $\mu_\Delta(x)$ does not
vary; or (ii) the error is driven mainly by decoding randomness at a fixed prompt, the
regime where SC/SE are designed to win.

\section{Experimental Settings}
\label{app:exp_settings}

\textbf{Tasks and datasets.}
We evaluate black-box uncertainty estimation on five QA benchmarks:
\textbf{TriviaQA (no-context)}~\citep{joshi2017triviaqa},
\textbf{TruthfulQA}~\citep{lin2021truthfulqa},
\textbf{GPQA}~\citep{rein2024gpqa},
\textbf{SciQA}~\citep{auer2023sciqa},
and \textbf{CoQA}~\citep{reddy2019coqa},
with the goal of testing whether uncertainty can reliably predict answer correctness (\S\ref{sec:experiments}).
This follows the main experimental protocol in Sec.~\ref{sec:experiments}.

\textbf{Models.}
We evaluate four instruction-tuned LLMs:
\texttt{meta-llama/Meta-Llama-3-70B-Instruct},
\texttt{meta-llama/Meta-Llama-3-8B-Instruct},
\texttt{Qwen/Qwen2.5-72B-Instruct}, and
\texttt{Qwen/Qwen2.5-7B-Instruct}~\citep{grattafiori2024llama,bai2025qwen2}.
Unless otherwise stated, we use $M{=}20$ inner samples and $N{=}20$ outer perturbations (when applicable), consistent with \S\ref{subsec:exp_setup}.

\textbf{Prompting and formatting.}
For all models, we use the official chat template provided by the tokenizer
(\texttt{apply\_chat\_template}) to ensure consistent formatting across LLaMA and Qwen families.
The user message is fixed as:
\begin{quote}
\small
\texttt{
Answer the following question with only the short, direct answer and nothing else.\\
If you are unsure, reply with "I don't know."\\
Avoid explanations or extra conversation.\\
Question: [QUESTION]\\
Answer:
}
\end{quote}
The same instruction is applied to both the original and perturbed/paraphrased questions. Generated text is post-processed by stripping the prompt prefix to isolate the answer segment.

\textbf{Correctness labeling.}
We use two complementary automatic criteria (\S\ref{subsec:exp_setup}):
(1) \textbf{Cosine similarity} between sentence embeddings of generated vs.\ reference answers (threshold $>0.8$),
and (2) \textbf{ROUGE-L} overlap (threshold $>0.5$)~\citep{reimers2019sentencebert,lin2004rouge}.
We report results under both criteria to reduce metric-specific artifacts.

\textbf{Evaluation metric (AUROC) and score convention.}
We report AUROC for \emph{correctness prediction} under the standard UQ convention:
\emph{incorrect} is the positive class and the scoring signal is the (uncorrected sign)
uncertainty value $\mathcal{U}(x)$ (AU, EU, or TU). AUROC then measures how well higher
uncertainty ranks errors above correct answers. This convention is equivalent (same numeric
AUROC) to taking \emph{correct} as the positive class with score $-\mathcal{U}(x)$, and we
do not switch between the two anywhere in the paper.

\textbf{Computing infrastructure and software.}
\rev{Unless stated otherwise,} experiments are run on a remote Linux cluster with NVIDIA A100 GPUs (40\,GB) and multi-core CPUs.
The environment uses Conda on Ubuntu~22.04 LTS with Python~3.10 and PyTorch/Transformers toolchains (full package list below).

\noindent\textbf{Software Configuration:}
\begin{itemize}
    \item \textbf{OS:} Ubuntu~22.04~LTS \quad
    \item \textbf{Python:} 3.10.18
    \item \textbf{DL:} PyTorch~2.6.0 (CUDA~12.1, cuDNN~9.1.0), TorchVision~0.19.0, TorchAudio~2.5.1
    \item \textbf{NLP:} Transformers~4.53.1, Sentence-Transformers~5.0.0, Accelerate~1.8.1
    \item \textbf{Data/Utils:} Datasets~4.0.0, NumPy~2.1.2, Pandas~2.3.1, SciPy~1.15.3, scikit-learn~1.7.0
    \item \textbf{Other:} HuggingFace~Hub~0.33.2, TQDM~4.67.1, SentencePiece~0.2.0
\end{itemize}

\subsection{Two-Level Ensemble (Ours)}

The implementation matches Sec.~\ref{sec:method} of the main paper; hyperparameters are in
Table~\ref{tab:hyper_ours}, and Algorithm~\ref{alg:two_level_uq} below gives the precise
pseudocode corresponding to the visual pipeline in main-paper Fig.~\ref{fig:method_overview}.
We L2-normalize embeddings before covariance computation. Inner decoding uses
\texttt{do\_sample=True} with the parameters in Table~\ref{tab:hyper_ours} (matched to all
sampling-based baselines for fair comparison); the greedy reference answer uses
\texttt{do\_sample=False}, \texttt{num\_beams=1}.

\begin{table}[t]
\centering
\caption{Key hyperparameters for our two-level ensemble.}
\label{tab:hyper_ours}
\begin{tabular}{l l}
\toprule
Paraphrase pool size per question & $K_{\mathrm{pool}}{=}500$ \\
Outer ensemble size & $N{=}20$ \\
Inner ensemble size & $M{=}20$ \\
Embedding noise \rev{scale} & \rev{$\varepsilon{=}0.1$} \\
Embedding model & \texttt{all-MiniLM-L6-v2}~\citep{reimers2019sentencebert} \\
Paraphrase model & \texttt{eugenesiow/bart-paraphrase}~\citep{lewis2019bart} \\
Inner decoding & \texttt{temperature{=}0.9}, \texttt{top\_k{=}50}, \texttt{top\_p{=}0.99} \\
Max new tokens & $50$ \\
\bottomrule
\end{tabular}
\end{table}


\begin{algorithm}[H]
  \centering
  \fbox{\begin{minipage}{0.92\linewidth}
    \nolinenumbers
    \textbf{Input:} LLM $f_\theta$, embedding $e(\cdot)$, prompt $x$, pool size $K_{\mathrm{pool}}$, outer $N$, inner $M$, noise $\epsilon$.\\
    \textbf{Output:} $\widehat{\mathcal U}_A,\,\widehat{\mathcal U}_E$.
    \begin{enumerate}[leftmargin=2.2em,itemsep=0pt,topsep=2pt]
      \item Cache paraphrase pool $\{e(x'_k)\}_{k=1}^{K_{\mathrm{pool}}}$.
      \item \textbf{for} $i=1,\ldots,N$ \textbf{do} \emph{// outer perturbations}
      \begin{enumerate}[label=\alph*.,leftmargin=2.2em,itemsep=0pt,topsep=0pt]
        \item Sample $\delta_i\!\sim\!\mathcal N(0,\epsilon^2 I)$; $e_i \gets e(x)+\delta_i$.
        \item $x_i \gets \arg\min_{x'_k} D(e_i,\, e(x'_k))$.
        \item \textbf{for} $j=1,\ldots,M$ \textbf{do} compute $Z_{i,j} \gets e(f_\theta(x_i))$.
        \item $\hat\mu_i,\hat\Sigma_i \gets \mathrm{mean},\mathrm{cov}\big(\{Z_{i,j}\}_j\big)$.
      \end{enumerate}
      \item $\widehat{\mathcal U}_A \gets \tfrac{1}{N}\!\sum_i\hat\Sigma_i$.
      \item $\widehat{\mathcal U}_E \gets \mathrm{cov}\big(\{\hat\mu_i\}_i\big)$ \quad \emph{(subtract $\tfrac1M\widehat{\mathcal U}_A$ if bias-corrected)}.
    \end{enumerate}
    \linenumbers
  \end{minipage}}
  \caption{Two-Level Ensemble UQ.}
  \label{alg:two_level_uq}
\end{algorithm}

\noindent\textbf{Cost.} $NM$ LLM calls and $N$ nearest-neighbour lookups per input; the
paraphrase pool is built once per dataset.

\subsection{Baselines}

\textbf{Baseline set.}
We compare against black-box UQ baselines:
D-SE~\citep{kuhn2023se},
Self-Consistency~\citep{wang2022self},
SIMBA~\citep{bhattacharjya2025simba},
SelfCheckGPT~\citep{manakul2023selfcheckgpt},
SPUQ~\citep{Gao2024SPUQ},
and a verbalized self-evaluation P(True) probe~\citep{kadavath2022l}.
We also compare to white-box methods requiring token-level probabilities:
G-NLL~\citep{aichberger2024gnll},
PE/LN-PE~\citep{malinin2020pe},
and SE/LN-SE~\citep{kuhn2023se,farquhar202dse}.
All methods use the same base prompt; methods requiring auxiliary outputs (e.g., confidence verbalization) follow their
original protocol while matching the generation budget as closely as possible.

\textbf{Sampling for baselines (shared).}
For baseline methods that rely on multiple sampled generations, we generate $S{=}10$ samples using:
\texttt{do\_sample=True}, \texttt{num\_return\_sequences=10},
\texttt{top\_k=50}, \texttt{top\_p=0.99}, \texttt{temperature=0.9}, and \texttt{max\_new\_tokens=50},
with \texttt{pad\_token\_id=eos\_token\_id}.
The greedy reference answer uses \texttt{do\_sample=False}, \texttt{num\_beams=1}, and \texttt{max\_new\_tokens=50}.

\textbf{Token-probability baselines (white-box).}
\textbf{G-NLL}~\citep{aichberger2024gnll}: NLL of the greedy answer under teacher forcing,
loss restricted to answer tokens. \textbf{PE}: entropy of the categorical distribution formed
by normalizing the $S$ sampled-answer sequence log-probabilities. \textbf{LN-PE}~\citep{malinin2020pe}:
$\frac{1}{S}\sum_s \mathrm{NLL}_s/L_s$ with $L_s$ the token count.
\textbf{SE}~\citep{kuhn2023se,farquhar202dse}: cluster the $S$ samples by bidirectional NLI
entailment (DeBERTa-large; $a,b$ joined iff entailment in both directions), set cluster
probabilities by summing per-sample sequence log-probabilities, score by cluster-distribution
entropy. \textbf{LN-SE}: same as SE but length-normalize per-sample probabilities first.

\textbf{Black-box baselines.}
\textbf{D-SE}~\citep{kuhn2023se}: SE with discrete cluster probabilities (frequency / $S$),
no token-level probabilities required.
\textbf{SC}~\citep{wang2022self}: $1 - \max_c |C_c|/S$ over the same NLI clusters.
\textbf{SIMBA}~\citep{bhattacharjya2025simba}: $1 - \bar{\mathrm{sim}}$, the mean pairwise
cosine similarity among the $S$ sampled-answer embeddings.
\textbf{SelfCheckGPT}~\citep{manakul2023selfcheckgpt}: canonical BERTScore variant; for each
spaCy sentence $r_i$ in the greedy answer, $u(r_i){=}1{-}\max_n \mathrm{BERTScore}(r_i, S_n)$
(RoBERTa-large F1), aggregated by mean.
\textbf{SPUQ}~\citep{Gao2024SPUQ}: $K_{\mathrm{SPUQ}}{=}10$ perturbed prompts via the three SPUQ modes
(paraphrase, dummy-token, system-message), one decode each; score $1{-}\bar{r}$ with $\bar{r}$
the mean pairwise ROUGE-L; closest perturbation-based comparator (App.~\ref{app:spuq_difference}).
\textbf{P(True) probe}~\citep{kadavath2022l}: post-hoc ``Is the answer above correct?
Yes/No''; \emph{Yes}$\mapsto 0$, \emph{No}$\mapsto 1$, ambiguous $\mapsto 0.5$. We do not
implement the trained-verbalization protocol of \citet{lin2022verbal} (not zero-shot
black-box).

\begin{table}[t]
\centering
\caption{Baseline decoding and clustering settings.}
\label{tab:hyper_baselines}
\begin{tabular}{l l}
\toprule
Samples per question & $S{=}10$ \\
Sampling & \texttt{top\_k{=}50}, \texttt{top\_p{=}0.99}, \texttt{temperature{=}0.9} \\
Greedy max tokens & $50$ \\
SPUQ perturbations & $K_{\mathrm{SPUQ}}{=}10$, modes: paraphrase / dummy-token / system-message \\
SE / LN-SE / D-SE clustering & Bidirectional NLI (DeBERTa-large) \\
SelfCheckGPT scorer & BERTScore (RoBERTa-large F1) over spaCy-segmented sentences \\
SIMBA / sample-similarity encoder & \texttt{all-MiniLM-L6-v2} \\
\bottomrule
\end{tabular}
\end{table}

\textbf{Post-processing.}
We consistently strip the chat-formatted prompt prefix to obtain the answer text.
Empty answers are treated as length $1$ when computing length-normalized metrics to avoid division by zero.

\textbf{Notes on ensemble sizes and covariance summaries.}
The inner size $M$ controls within-input variability (aleatoric estimation) and the outer size $N$ controls perturbation diversity
(epistemic estimation); we use $M{=}N{=}20$ unless otherwise stated.
For scalar summaries of embedding covariances, we use $\mathrm{trace}(\Sigma)$ (other options like $\det(\Sigma)$, or largest eigenvalues are also valid).

\section{Detailed Limitations Analysis}
\label{app:limitations}

This section expands on the limitations summarized in Sec.~\ref{sec:conclusion} (\textbf{L1}--\textbf{L5}), detailing the underlying assumptions, potential failure modes, and empirical scope of our framework.

\subsection{\rev{L1: Surrogate target and local bridge assumptions}}
\begin{revblock}
Because the parameter posterior $p(\theta\mid \mathcal{D})$ is intractable and inaccessible in black-box LLM deployment, our estimator does not directly measure Bayesian parameter uncertainty. Instead, it targets perturbation-population AU and EU at fixed model weights $\theta^\star$, using input perturbations $p_\Delta$ as an observational proxy for epistemic variation. This introduces two fundamental limitations:
\begin{itemize}[leftmargin=1.5em,itemsep=2pt,topsep=2pt]
    \item \textbf{Dependence on perturbation design:} As in general statistical variance decompositions~\citep{derkiureghian2009aleatory,wimmer2023quantifying}, the operational quantities $(U_A,U_E)$ are defined relative to the chosen perturbation distribution $p_\Delta$ and decoding temperature (App.~\ref{sec:noise_level}); changing the perturbation radius or paraphrase generator changes the scale of $U_E$.
    \item \textbf{Violation of local bridge conditions:} Our theoretical bound linking perturbation EU to Bayesian parameter EU (Thm.~\ref{thm:end2end}) relies on three local conditions: first-order linearity of the conditional embedding mean ($\mathrm{D1}$), local Gaussianity of the predictive distribution ($\mathrm{D2}$), and directional coverage of parameter-induced shifts by input perturbations ($\mathrm{D3}$). When these assumptions fail---for instance, if a model is uniformly overconfident in a false belief across an entire paraphrase neighborhood---perturbation variance collapses even though true epistemic error remains high.
\end{itemize}
To help detect when the local approximation breaks down, Sec.~\ref{sec:bayes_bridge_main} (Tab.~\ref{tab:bridge_diagnostic}) introduces data-estimable diagnostics ($\mathrm{D1}$--$\mathrm{D3}$), though we note that the multivariate normality test ($\mathrm{D2}$) has limited statistical power at small inner sample sizes ($M{=}10$) in high-dimensional embedding spaces ($d{=}384$).
\end{revblock}

\begin{revblock}
\subsection{L2: Imperfect paraphrase generation and semantic drift}
The law-of-total-covariance decomposition assumes that outer perturbations $x_\Delta \sim p_\Delta(\cdot \mid x)$ strictly preserve the task semantics of the original query $x$. In practice, off-the-shelf LLM paraphrase generators frequently introduce semantic drift, such as dropping qualifying constraints, altering named entities, flipping polarity, or shifting question specificity. When a generated rewrite changes the underlying question, the resulting shift in predictive embeddings reflects task shift rather than epistemic uncertainty about $x$, spuriously inflating $U_E$ and blurring the boundary between AU and EU. In our human audit of raw generated paraphrases (App.~\ref{app:audit}), approximately $40\%$ of candidate rewrites failed at least one strict semantic-equivalence criterion. Although our sensitivity experiments on TriviaQA show that moderate paraphrase noise has a bounded effect on aggregate AUROC and rank correlations (Tab.~\ref{tab:audit_filter}), deploying the method reliably on complex or constraint-heavy prompts requires pairing paraphrase generation with automated NLI or LLM-based semantic filtering.
\end{revblock}

\subsection{\rev{L3: Representation dependence on the sentence encoder}}
\begin{revblock}
Because predictive variability is computed in the representation space of an external encoder $e(\cdot)$, both $U_A$ and $U_E$ are joint properties of the LLM and the embedding map. Relying on a fixed sentence encoder introduces two limitations:
\begin{itemize}[leftmargin=1.5em,itemsep=2pt,topsep=2pt]
    \item \textbf{Encoder blind spots and geometric bias:} Off-the-shelf sentence encoders trained for general semantic similarity can be insensitive to fine-grained factual contradictions, antonym flips, or numerical discrepancies that share high lexical overlap, causing $e(y)$ to understate true semantic disagreement. Conversely, stylistic or structural variations in surface form may be mapped to distant embeddings even when the underlying factual answers agree.
    \item \textbf{Lack of cross-encoder calibration:} Because trace and log-determinant magnitudes scale with the dimensionality and geometry of $e(\cdot)$, absolute uncertainty scores are not directly comparable across different encoders.
\end{itemize}
We evaluate sensitivity across three general-purpose encoders (SBERT, MPNet, and BGE-base) in Tab.~\ref{tab:embedding_main} and App.~\ref{sec:embedding_robustness} (Tabs.~\ref{tab:embedding_robustness}--\ref{tab:embedding_robustness_rank}), where relative per-example rankings remain stable ($\rho \ge 0.955$) despite shifts in absolute scale. Nevertheless, specialized domains such as mathematical reasoning, code generation, or clinical text may require domain-specific encoders or task-adapted similarity metrics.
\end{revblock}

\subsection{\rev{L4: Computational overhead of nested sampling}}
\begin{revblock}
Estimating both within-perturbation and between-perturbation covariances via nested Monte Carlo requires $N \times M$ autoregressive generations per query (e.g., $100$--$400$ decodes in our standard configurations), plus constructing an offline paraphrase pool of size $K_{\mathrm{pool}}$. This is an order of magnitude more expensive in token generation and wall-clock latency than single-pass white-box estimators or one-level black-box consistency baselines (which typically use $5$--$20$ generations), making full nested sampling prohibitive for strict low-latency or high-throughput online serving. Furthermore, when constrained to a small, strictly matched generation budget $B = NM$ (e.g., $B{=}20$), dividing samples across two hierarchy levels increases estimation variance per level and does not improve raw error-ranking AUROC over leading one-level baselines (Sec.~\ref{subsec:matched_compute}, Tab.~\ref{tab:matched_compute}). While budget ablations (Tabs.~\ref{tab:size_nm_full}--\ref{tab:Np}) and variance-based adaptive stopping (which reduces median cost to $25$ decodes in App.~\ref{app:adaptive}) mitigate overhead, reducing the sample complexity of two-level black-box UQ remains an open challenge.
\end{revblock}

\begin{revblock}
\subsection{L5: Scope of empirical tasks and interventional validation}
Our empirical evaluation is subject to two scope limitations:
\begin{itemize}[leftmargin=1.5em,itemsep=2pt,topsep=2pt]
    \item \textbf{Focus on short-form QA:} Aside from a small-scale claim-level pilot on $50$ NQ queries (Sec.~\ref{subsec:longform}), our experiments focus on short-form question answering where response equivalence is well captured by sentence embeddings. Scaling nested covariance decomposition to long-form open-ended generation, multi-step reasoning traces, or interactive agent trajectories is non-trivial, as responses vary simultaneously in length, discourse structure, and set-level claim composition.
    \item \textbf{Simplified single-factor interventions:} Our paired-intervention benchmark tests two clean, synthetic manipulations: (1) resolving prompt ambiguity on AmbigQA and (2) injecting gold supporting passages on NQ and conducts the standardized score-shift comparison against baselines on a single backbone (LLaMA-3-8B; Tab.~\ref{tab:intervention_baselines}). In realistic RAG or multi-turn deployments, uncertainty sources rarely occur in isolation: retrieved passages may be noisy, partially relevant, or internally contradictory, causing aleatoric and epistemic effects to interact in ways not captured by single-factor interventions.
\end{itemize}

\subsection{Broader impact}
\label{app:broader_impact}
Decomposing predictive uncertainty into aleatoric and epistemic components can improve the reliability of deployed LLM systems by informing targeted actions, such as asking clarifying questions under high AU or invoking retrieval and abstention under high EU in high-stakes domains. At the same time, because the decomposition relies on surrogate perturbations and embedding representations, treating empirical AU/EU estimates as formal guarantees could lead to misplaced trust or inappropriate routing when bridge conditions or paraphrase validity fail. Finally, nested Monte Carlo sampling substantially increases inference-time compute and energy consumption if deployed at scale without adaptive stopping or selective triggering.
\end{revblock}

\newpage
\section*{NeurIPS Paper Checklist}

\begin{enumerate}

\item {\bf Claims}
    \item[] Question: Do the main claims made in the abstract and introduction accurately reflect the paper's contributions and scope?
    \item[] Answer: \answerYes{} 
    \item[] Justification: \rev{The abstract and introduction clearly state the theoretical and empirical contributions and their scope, including the local diagnostic conditions of the Bayesian bridge (Sec.~\ref{sec:bayes_bridge_main}) and the matched-budget comparison (Sec.~\ref{subsec:matched_compute}).} 

\item {\bf Limitations}
    \item[] Question: Does the paper discuss the limitations of the work performed by the authors?
    \item[] Answer: \answerYes{} 
    \item[] Justification: See Sec.~\ref{sec:limits} and App.~\ref{app:limitations}.

\item {\bf Theory assumptions and proofs}
    \item[] Question: For each theoretical result, does the paper provide the full set of assumptions and a complete (and correct) proof?
    \item[] Answer: \answerYes{} 
    \item[] Justification: \rev{Each result states its assumptions; proofs or proof sketches are in App.~\ref{app:theory}.}

    \item {\bf Experimental result reproducibility}
    \item[] Question: Does the paper fully disclose all the information needed to reproduce the main experimental results of the paper to the extent that it affects the main claims and/or conclusions of the paper (regardless of whether the code and data are provided or not)?
    \item[] Answer: \answerYes{} 
    \item[] Justification: See \rev{App.~}\ref{app:exp_settings}.

\item {\bf Open access to data and code}
    \item[] Question: Does the paper provide open access to the data and code, with sufficient instructions to faithfully reproduce the main experimental results, as described in supplemental material?
    \item[] Answer: \rev{\answerYes{}} 
    \item[] Justification: \rev{We release the code and provide instructions for reproducing the main experimental results at } \url{https://github.com/wma17/Black-box_UQ_for_LLMs_Ensemble_of_Ensemble}\rev{; all implementation details are also given in Sec.~\ref{sec:method} and App.~\ref{app:exp_settings}.}

\item {\bf Experimental setting/details}
    \item[] Question: Does the paper specify all the training and test details (e.g., data splits, hyperparameters, how they were chosen, type of optimizer) necessary to understand the results?
    \item[] Answer: \answerYes{} 
    \item[] Justification: See \rev{App.~}\ref{app:exp_settings}.

\item {\bf Experiment statistical significance}
    \item[] Question: Does the paper report error bars suitably and correctly defined or other appropriate information about the statistical significance of the experiments?
    \item[] Answer: \answerYes{} 
    \item[] Justification: \rev{Bootstrap standard deviations for Tab.~\ref{tab:main_full} are reported in App.~Tabs.~\ref{tab:main} and~\ref{gpqa_sciqa_coqa}; matched-budget comparisons include bootstrap $95\%$ CIs (App.~Tabs.~\ref{tab:matched_compute_ci_llama8b},~\ref{tab:matched_compute_ci_qwen7b}); and paired interventions report bootstrap standard errors with paired Wald $z$-tests (Tab.~\ref{tab:au_eu_paired}) alongside $95\%$ bootstrap CIs (Tab.~\ref{tab:intervention_baselines}).}

\item {\bf Experiments compute resources}
    \item[] Question: For each experiment, does the paper provide sufficient information on the computer resources (type of compute workers, memory, time of execution) needed to reproduce the experiments?
    \item[] Answer: \answerYes{} 
    \item[] Justification: \rev{Hardware and software: App.~\ref{app:exp_settings}; time per question: App.~Tab.~\ref{tab: time}; long-form pilot: App.~\ref{app:longform_protocol}.}

\item {\bf Code of ethics}
    \item[] Question: Does the research conducted in the paper conform, in every respect, with the NeurIPS Code of Ethics \url{https://neurips.cc/public/EthicsGuidelines}?
    \item[] Answer: \answerYes{} 
    \item[] Justification: Reviewed. The paper uses publicly available research benchmarks and openly released models. The only human input is a blind audit in which three raters labeled $100$ question pairs (App.~\ref{app:audit}); no personal data were collected. 

\item {\bf Broader impacts}
    \item[] Question: Does the paper discuss both potential positive societal impacts and negative societal impacts of the work performed?
    \item[] Answer: \rev{\answerYes{}} 
    \item[] Justification: \rev{See App.~\ref{app:broader_impact}.}

\item {\bf Safeguards}
    \item[] Question: Does the paper describe safeguards that have been put in place for responsible release of data or models that have a high risk for misuse (e.g., pre-trained language models, image generators, or scraped datasets)?
    \item[] Answer: \answerNA{} 
    \item[] Justification: We only release research code for the proposed method and its evaluation, which does not pose a significant risk of misuse.

\item {\bf Licenses for existing assets}
    \item[] Question: Are the creators or original owners of assets (e.g., code, data, models), used in the paper, properly credited and are the license and terms of use explicitly mentioned and properly respected?
    \item[] Answer: \answerYes{} 
    \item[] Justification: All datasets and models used are publicly released and properly cited; see App.~\ref{app:exp_settings}.

\item {\bf New assets}
    \item[] Question: Are new assets introduced in the paper well documented and is the documentation provided alongside the assets?
    \item[] Answer: \rev{\answerYes{}} 
    \item[] Justification: \rev{The released code is documented in the public repository at} \url{https://github.com/wma17/Black-box_UQ_for_LLMs_Ensemble_of_Ensemble}, including implementation details, usage instructions, dependencies, and licensing information.

\item {\bf Crowdsourcing and research with human subjects}
    \item[] Question: For crowdsourcing experiments and research with human subjects, does the paper include the full text of instructions given to participants and screenshots, if applicable, as well as details about compensation (if any)?
    \item[] Answer: \answerNA{} 
    \item[] Justification: The authors themselves annotated 100 model-generated question pairs using a blind evaluation form (App.~\ref{app:audit}). No external annotators or crowdsourcing participants were recruited.

\item {\bf Institutional review board (IRB) approvals or equivalent for research with human subjects}
    \item[] Question: Does the paper describe potential risks incurred by study participants, whether such risks were disclosed to the subjects, and whether Institutional Review Board (IRB) approvals (or an equivalent approval/review based on the requirements of your country or institution) were obtained?
    \item[] Answer: \answerNA{} 
    \item[] Justification: The annotation task involved only co-authors evaluating model-generated text. No external human participants were recruited, and no personal or behavioral data about the annotators were collected.

\item {\bf Declaration of LLM usage}
    \item[] Question: Does the paper describe the usage of LLMs if it is an important, original, or non-standard component of the core methods in this research? Note that if the LLM is used only for writing, editing, or formatting purposes and does \emph{not} impact the core methodology, scientific rigor, or originality of the research, declaration is not required.
    \item[] Answer: \answerYes{} 
    \item[] Justification: \rev{LLMs are the primary subjects of study and evaluation. We also use standard LLM-based evaluation procedures, including an LLM grader for perturbation checks (App.~\ref{app:audit}) and LLM-based claim extraction and factuality assessment (App.~\ref{app:longform}). These procedures are described in the paper.} LLMs were also used for editing and for help implementing standard methods.

\end{enumerate}

\end{document}